# NUMERICAL APPROXIMATION IN CFD PROBLEMS USING PHYSICS INFORMED MACHINE LEARNING

*A Thesis*

*Submitted by*

## SIDDHARTH ROUT

*for the award of the degree*

*of*

## BACHELOR OF TECHNOLOGY

## &

## MASTER OF TECHNOLOGY

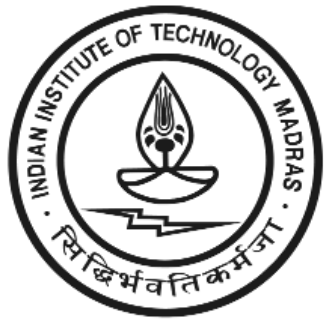

## DEPARTMENT OF MECHANICAL ENGINEERING
## INDIAN INSTITUTE OF TECHNOLOGY MADRAS
## CHENNAI-600036

## MAY 2019

i

*Jai Jagannath*

# THESIS CERTIFICATE

This is to certify that the thesis entitled **"Numerical Approximation in CFD Problems Using Physics Informed Machine Learning"** submitted by **Siddharth Rout** to the **Indian Institute of Technology, Madras** for the award of **B. Tech – M. Tech Dual Degree** is a bona fide record of research work carried out by him under my supervision. The contents of this thesis, in full or in parts, have not been submitted to any other Institute or University for the award of any degree or diploma.

**Prof. Balaji Srinivasan**
Associate Professor
Department of Mechanical Engineering
Indian Institute of Technology Madras
Chennai – 600 036.

Place: Chennai

Date: May 2019

# ACKNOWLEDGEMENTS

My father often says, "Every one of us needs a good mentor and back support to achieve something in life. A good mentor is a privilege and privileges should always be respected." It gives me a sense of satisfaction to express my sincere gratitude to my supervisor Dr. Balaji Srinivasan for being a mentor and guide to me in true sense. He is the most sensible person I have ever met. He has given a direction to my career. I would like to thank him for introducing me to the exciting field of applied machine learning in fluid mechanics. I would like to thank God for putting me at the right place. It is a privilege to be part of his research team.

I would like to thank my lab mate Vikas Dwivedi for being a constant support throughout the last year. We have spent hours and days discussing various problems or test results which we encountered during the course of the project. No matter if we got the right or wrong solution, that helped us to stay motivated.

# ABSTRACT

*Keywords*: Numerical approximation; Advection-Diffusion Equation; Advection Dominant Flows; Neural Networks; Physics Informed Learning; Extreme Learning Machine.

The thesis focuses on various techniques to find an alternate approximation method that could be universally used for a wide range of CFD problems but with low computational cost and low runtime. Various techniques have been explored within the field of machine learning to gauge the utility in fulfilling the core ambition. Steady advection diffusion problem has been used as the test case to understand the level of complexity up to which a method can provide solution. Ultimately, the focus stays over physics informed machine learning techniques where solving differential equations is possible without any training with computed data. The prevalent methods by I.E. Lagaris et.al. and M. Raissi et.al are explored thoroughly.

The prevalent methods cannot solve advection dominant problems. A physics informed method, called as Distributed Physics Informed Neural Network (DPINN), is proposed to solve advection dominant problems. It increases the flexibility and capability of older methods by splitting the domain and introducing other physics-based constraints as mean squared loss terms. Various experiments are done to explore the end to end possibilities with the method. Parametric study is also done to understand the behavior of the method to different tunable parameters. The method is tested over steady advection-diffusion problems and unsteady square pulse problems. Very accurate results are recorded.

Extreme learning machine (ELM) is a very fast neural network algorithm at the cost of tunable parameters. The ELM based variant of the proposed model is tested over the advection-diffusion problem. ELM makes the complex optimization simpler and since the method is non-iterative, the solution is recorded in a single shot. The ELM based variant seems to work better than the simple DPINN method. Simultaneously scope for various development in future are hinted throughout the thesis.

# TABLE OF CONTENTS

# LIST OF FIGURES

# LIST OF TABLES

# CHAPTER 1
# INTRODUCTION

## 1.1. INTRODUCTION

Computational fluid dynamics (CFD) is the science of simulation, prediction and analysis of fluid flows, heat transfer, mass transfer, chemical reactions and related phenomena by solving the physics based governing equations for large set of finite points or finite volume cells. The governing equations are mostly ordinary differential equations (ODEs) and partial differential equations (PDEs). Exact solution for most of the ODEs and PDEs is not known. To solve a large set of differential equations various numerical approximation methods are used. The differential equations are discretized to get the approximate equation called difference equation. Solving the sets of equations for CFD problems is computationally expensive and time consuming. A person needs to have high performance computing facility and cooling facility before he thinks to solve a high-fidelity problem.

Again, the equations being solved are not exact, so the solution is also not exact. To reduce the error in solution various higher order methods are used. There can be three types of error in solution:

a) Instability: Unphysical oscillations
b) Accuracy
c) Shift

The methods to solve each type of error is different. There is no universal method to address all the types of error. Again, typically the methods are problem specific. Application of these methods extensively increases the accompanied computational cost.

The concept of machine learning can help us to tackle all the above issues. In fact, machine learning models could be used to do a whole lot of other tasks like generation high resolution solution from a coarse grid solution or corrupt solution, prediction of correction, prediction of unpredictable parameters, data driven analysis of flow-field etc.

In order to reduce the computation expense, can we think of any alternate approximation method? The thesis will revolve around this broad question.

## 1.2. OBJECTIVE OF THE WORK

The main concern in this thesis has been devised from the motivation that an alternate approximation method could reduce the computation expense. It may not reflect in the scope of this project but the thesis would justify the claim.

Our prime objective is to develop an alternate approximation method which could be universally applied. The main targets of the project are to generate a more accurate solution and/or speed up the computation. We believe the concept of machine learning could help us achieving our ambition.

Primarily, the focus is to develop a neural network based differential equation solver which could generate a function that could approximate the solution of the differential equation. This proposition is based on the results of the initial research by George Cybenko (1989), where he proves that neural networks are universal approximators (Universal Approximation Theorem). If we could approximate a differential equation over the domain or over parts of the domain, we would be essentially devising an appropriate shape function which could be used on larger grids to produce equivalent approximation. There would be a tradeoff between the complexity of shape function and the grid size. Vaguely stating, a less complex shape function approximated over a way larger grid would make the computation less expensive.

The proposition has supporting reasons to justify the claim. The main challenge is to conceptualize and device a method.

## 1.3. ARTIFICIAL NEURAL NETWORKS

Artificial neural networks (ANN) are a class of algorithms loosely modelled on the typical connections between biological neurons in the brain. It also adopts the fact that neurons loosely process the information before let it out. The robustness of information processing inspires computational researchers to adopt the base concept. However, the engine that drives an ANN is completely different from the biological neural network. There is no match in the way ANN functions and the way biological neurons function.

Though the hype in the usages of ANNs is very new to the world, but the concepts are many generations old. The recent hype is due to the availability of computational power and surge in the utility in the field of image processing. The first artificial neuron was produced by neurophysiologist Warren McCulloch and logician Walter Pits in 1943. The concept of neural networks was first proposed by Alan Turing in 1948.

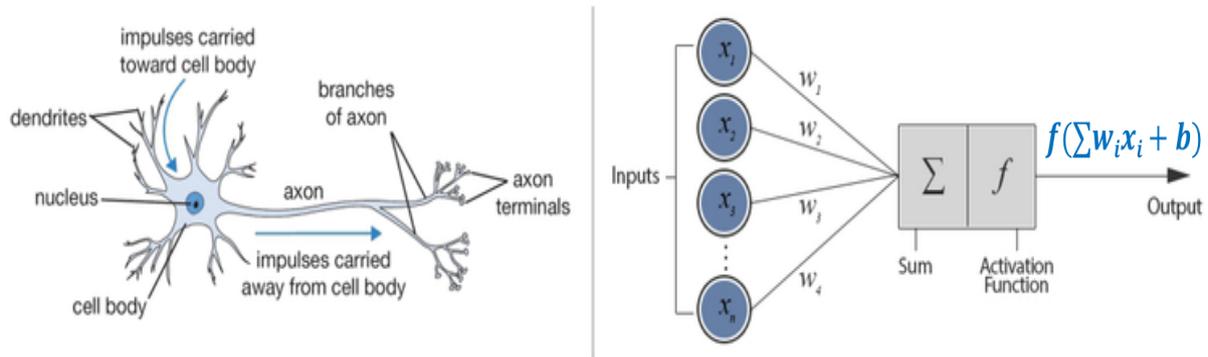

*Figure 1.1: Analogy between biological and artificial neuron*

A neural network takes single or multiple feed(s) as input. It can produce more than one output as well. The process that happens in between the input and output is a combination of weighed linear summation and non-linear activation. Each block that does these processes are called a neuron (or node). The first task for a neuron is to assign weights to each input and produce a weighed linear combination of inputs. This generates a single value that represents the total impact of all the inputs. Most often a bias numerical value is added to it to allow flexibility. The second task for the neuron is to introduce a non-linearity. The neuron feeds the value obtained in the last step into a function (called activation function) to get the final output. The commonly used activation functions are tanh(x), sigmoid(x), ReLU(x), SeLU(x), Heaviside(x) etc. These non-linear functions allow non-linear scaling of the output. Neural network is a set of interconnected neurons. A set of parallel neurons is called a hidden layer. If the number of hidden neurons is more than one, then the network is called deep neural network.

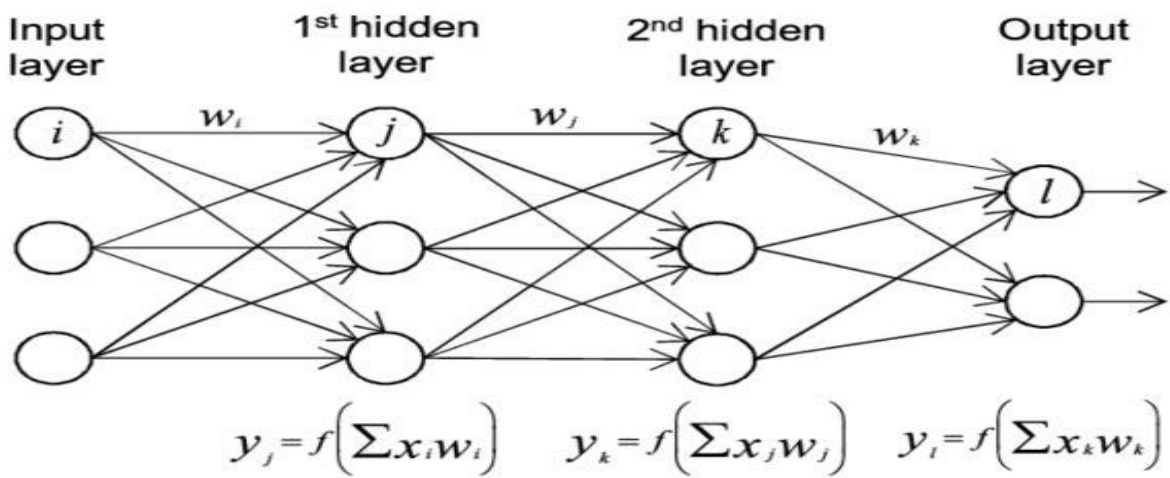

*Figure 1.2: Deep neural network architecture.*

The final prediction can be represented in matrix and vector form:

$$Y = W_k f_j \big(W_j f_i (W_i X + B_i) + B_j\big) + B_k$$

X  = Input Vector        W = Weight Matrix

Y  = Output Vector       B  = Bias Vector

The complexity of the architecture allows to adjust a greater number of weights and biases. This makes the network more tuneable since it has more controls. The weights and biases cannot be set manually, so we train the network to find the values which minimises the error in output. It can be considered as determining the weights and biases for fitting the function to our data (X, Y). Typically, this regression is done using various optimization algorithms like Gradient Descent and variants, Levenberg Marquardt, BFGS etc.

Applications of neural networks can be broadly classified into:

- Regression (function approximation)
- Classification
- Data processing

## 1.4. NEURAL NETWORKS AS FUNCTION APPROXIMATORS

Neural networks have various applications but the class of problems discussed in the thesis lies under regression or function approximation category. Universal approximation theorem states that a feed-forward network with a single hidden layer containing a finite number of neurons can approximate continuous functions on compact subsets of $R^n$, under mild assumptions on the activation function. George Cybenko gave the first proposal in 1989 for sigmoid activation functions. In fact, in 1991 Kurt Hornik proved that it is not the specific choice of the activation function, but rather the multilayer feedforward architecture which gives the potential of being universal approximators.

Ordinary differential equations (ODEs) and partial differential equations (PDEs) cannot always be solved exactly. In this thesis, the focus is to approximate differential equations. This is not so straight forward. If the equations are time variant or non-linear then underlying issues like computational load and optimisation difficulty come into picture. Anyway, the neural network function would be an approximation of the exact function or an approximate shape function. A good algorithm can take advantage of this capability addressing the underlying issues.

Three benchmark problems that could targeted are:

1. Steady Advection Diffusion

$$C\frac{du}{dx} = \epsilon \frac{d^2u}{dx^2}$$

2. Unsteady Linear Advection

$$\frac{du}{dt} + C\frac{du}{dx} = 0$$

3. Unsteady Non-Linear Burgers

$$\frac{du}{dt} + u\frac{du}{dx} = \epsilon \frac{d^2u}{dx^2}$$

## 1.5. THE PROBLEM WITH ADVECTION-DIFFUSION

Solving one dimensional steady advection diffusion equation is a base problem to start because it is time invariant and linear. It is still precious because it gives us the scope to figure out and solve the problems in extreme cases.

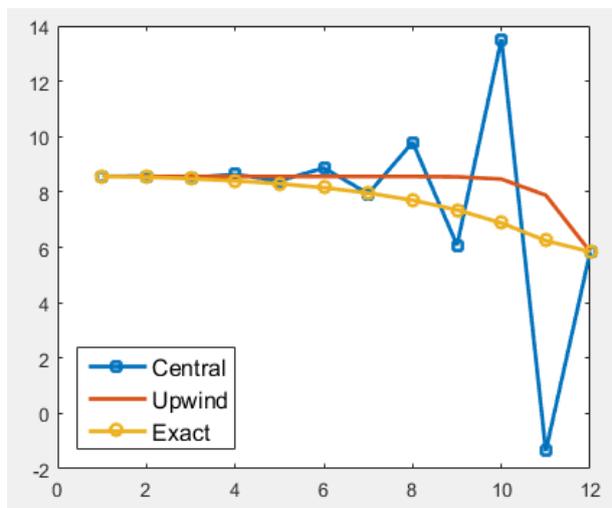

$$c\frac{du}{dx} = \epsilon \frac{d^2u}{dx^2}$$

$$Pe = \frac{c\Delta x}{\epsilon}$$

Exact: $\dfrac{u-u_o}{u_L-u_o} = \dfrac{e^{\frac{c.x}{\epsilon}}-1}{e^{\frac{c.L}{\epsilon}}-1}$

$CDS: c\dfrac{u_{i+1}-u_{i-1}}{2\Delta x} = \epsilon \dfrac{u_{i+1}-2u_i+u_{i-1}}{(\Delta x)^2}$

$UDS: c\dfrac{u_i-u_{i-1}}{\Delta x} = \epsilon \dfrac{u_{i+1}-2u_i+u_{i-1}}{(\Delta x)^2}$

*Figure 1.3: Advection-diffusion solutions*

Conventionally, advection diffusion equation could be solved using central difference scheme only if the problem is not advection dominant. If local Peclet number > 2, then instability is clearly evident. The unphysical oscillations increase heavily with the increase in Peclet number. Upwind difference scheme is clearly successful in damping the oscillations but the solution is dissipated. The dissipation too decreases with decreasing Peclet number. So, lower order methods do not work well for advection dominant cases. Higher order methods are used to give much accurate solution but computation becomes very expensive. This thesis is all about developing an alternate method with special interest on advection dominant problems.

# CHAPTER 2
# DATA DRIVEN METHODS

## 2.1.1. NON-LINEAR REGRESSION OF PHYSICAL SOLUTION FOR ADVECTION-DIFFUSION EQUATION

The exact solution for advection-diffusion equation with Dirichlet boundary conditions could be found out easily.

$$\text{Exact: } \boldsymbol{u} = \frac{u_L - u_o}{e^{\frac{c.L}{\epsilon}} - 1} e^{\frac{c.x}{\epsilon}} + \frac{u_o e^{\frac{c.L}{\epsilon}} - u_L}{e^{\frac{c.L}{\epsilon}} - 1}$$

The exact equation can fit into the form:

$$\boldsymbol{\phi = a\mathrm{e}^{bx} + c}$$

where a, b and c can be determined by regression. For regression we need to generate numerous exact value points using the exact equation. We have three unknowns and with just three points we can solve the equations to get a, b and c exactly. But it is interesting to know the behaviour of partial gradients with respect to the unknown variables.

Various non-linear regression techniques could be used for this problem. The techniques that were used are:

Gauss-Newton Algorithm: $(J^T J)\delta = J^T [y - f(\beta)]$

Marquardt Algorithm: $(J^T J + \lambda I)\delta = J^T [y - f(\beta)]$

Levenberg-Marquardt Algorithm: $(J^T J + \lambda \, diag(J^T J))\delta = J^T [y - f(\beta)]$

Tikhonov's Regularisation: $(J^T J + \lambda I)\delta = J^T [y - f(\beta)] + \lambda A$

Jacobian Matrix (**J**) = $\left[ \left[\frac{\partial \phi i}{\partial a}\right] \quad \left[\frac{\partial \phi i}{\partial b}\right] \quad \left[\frac{\partial \phi i}{\partial c}\right] \right]$     **A** = { a, b, c}     δ = {δa, δb, δc}

$\left[\frac{\partial \phi i}{\partial a}\right]$ is a column matrix, where each element accounts for each exact value point.

To get the fitting variables, the unknown variables are randomly initialised and δ vector is solved in each iteration. Fitting variables are updated using **A\* = A + δ**.

## 2.1.2. OBSERVATIONS

Ideally the fitting should produce very accurate solution but it is not that straight forward. δ vector is not always easily solvable. Three types of solutions could be seen:

1. Properly converged
2. Unstable
3. Singular $J^TJ$

Apart from difficulty in solving, solution depends on initialization, boundary conditions and tuning parameters as well.

Using the Gauss-Newton Algorithm (GNA), for large domain it was clearly evident that the $J^TJ$ matrix became singular. Especially, if $\epsilon < 0.2$, then finding the regression becomes extremely difficult. The small domains over which solution is stable are distributed all over the $\mathbf{R}^1$. However, for negative values $\epsilon$ of the occurrence of singularity is remarkably rarer. This is a strange behavior, which comes into frame in further chapters of this thesis as well.

All other techniques mentioned above have regularization parameter ($\lambda$) which helps in conditioning the $J^TJ$ matrix to avoid singularity. Even after adjusting the tuning parameter the regression is unstable for most of the portion of the domain. The Marquardt Algorithm (MA) and Tikhonov's Regularization (TR) perform better than the Levenberg-Marquardt Algorithm (LMA) though LMA is typically assumed to be more robust than the former algorithms. MA, TR and LMA perform appreciably better than GNA but none of them are robust for our problem. If $\epsilon < 0.02$ none of them are able to produce appreciable regression.

## 2.2.1. ENHANCEMENT OF SOLUTION FROM CENTRAL DIFFERENCE SCHEME

Central difference scheme (CDS) produces unstable solution for advection dominant problems. Central differencing is a good approximation for isotropic spread of the property. A map between CDS solution and exact solution can be easily built using a single layered neural network since CDS and exact solutions can be easily found out. If a generalized map could be found out, then it could enhance CDS solution for any advection-diffusion problem. This is the key advantage of the model.

Let N finite grid points be taken over the test domain. A single layered neural network is trained with the exact solution for the cases with different $\epsilon$ and different boundary conditions as ground truth. The input takes in the CDS prediction at the N points through N input nodes as feed. The ground truth at the output layer is the exact values at the N finite points.

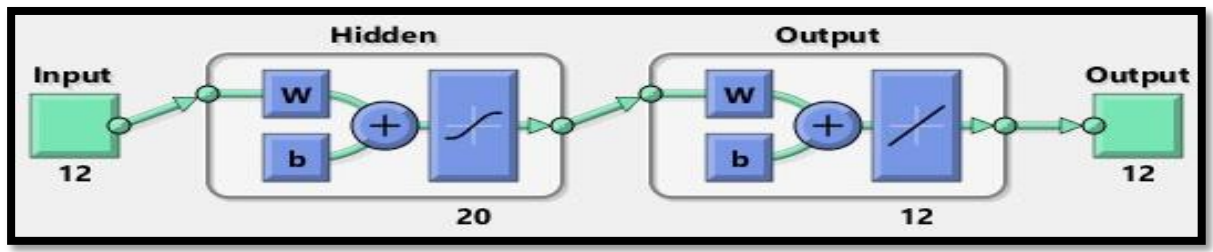

*Figure 2.1: Architecture for solution enhancement*

### 2.2.2. OBSERVATIONS

On training the model over sufficient number of cases, the approximation is actually better than upwind case. If cases are for specific boundary values then the approximation is appreciably better. Approximation is even better if all the cases have a specific Pe value. General cases are the major concern here. The mapping ability of this model is very good but the solution is not very accurate for our purpose. Slight deviations around the exact solution could be easily noticed.

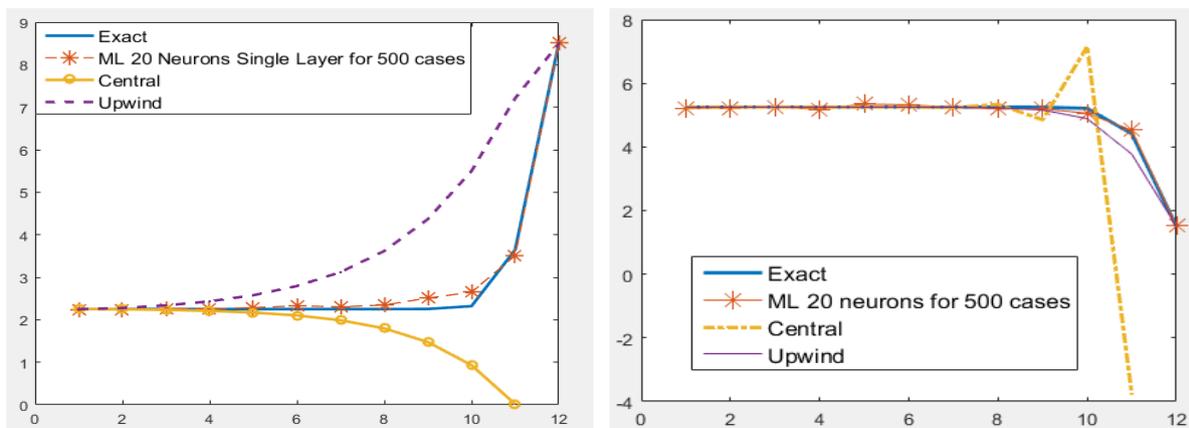

*Figure 2.2: Comparisons of neural network enhanced predictions vs upwind solution for two cases*

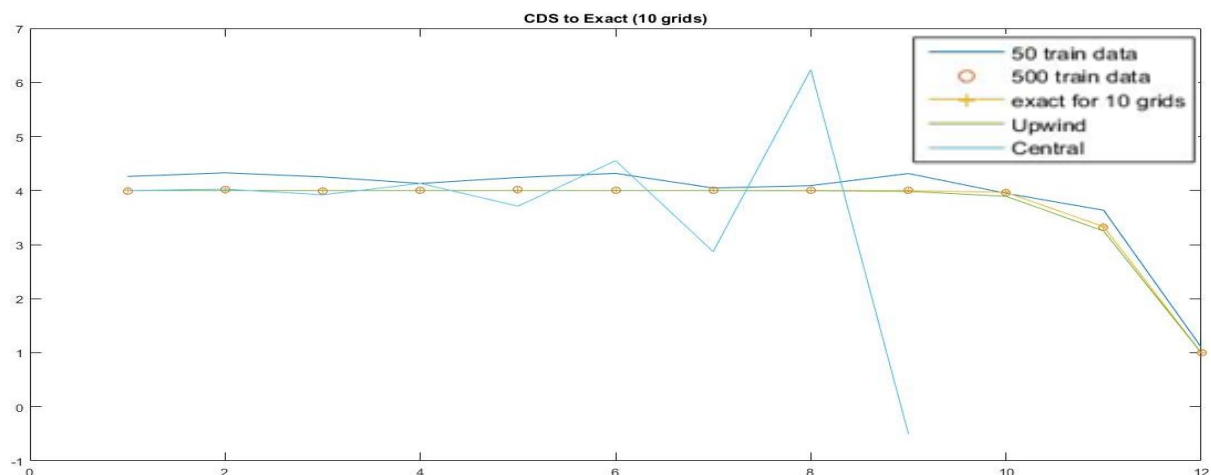

*Figure 2.3: Effect of number of training data enhancement.*

The prediction would improve with the increase in the depth of network and increase in the number of grid points. There are two major limitations of this method. The predictions are limited to the range of the cases over which the network has been training. It cannot be used to enhance adverse cases. The second major limitation is inability to enhance the general second order differential equation (**au'' + bu' + cu + d = 0**). The complexity of the CDS to exact map increases when generalized. Deeper networks may help on capturing the complexity.

The underlying advantages of this method are ability to predict for very high Pe cases. The second major advantage is ability to speed up prediction by obtaining coarse CDS solution and enhancing it to an estimate that we would have obtained from higher resolution solution. In fact, neural networks could be used to generate higher resolution solution from low resolution solution. In an experiment, it was observed to be giving better estimate than Richardson's extrapolation.

### 2.3.1. PREDICTION OF ERROR IN CDS SOLUTION

The error in the CDS solution can be exactly found out analytically. Plugging the exact solution in the central difference equation gives the exact residual. The residual is in the form of a false diffusion. If artificially extra diffusion (artificial diffusion) is introduced then the CDS would produce the exact solution.

$$a \frac{d^2 \emptyset}{dx^2} = b \frac{d\emptyset}{dx} \qquad a^* = a + \mu_{artificial} \qquad Pe = \frac{b \Delta x}{a}$$

$$\boxed{\mu_{artificial} = \frac{a}{e^{Pe} + e^{-Pe} - 2} \left[ \frac{Pe}{2}(e^{Pe} - e^{-Pe}) - (e^{Pe} + e^{-Pe} - 2) \right]}$$

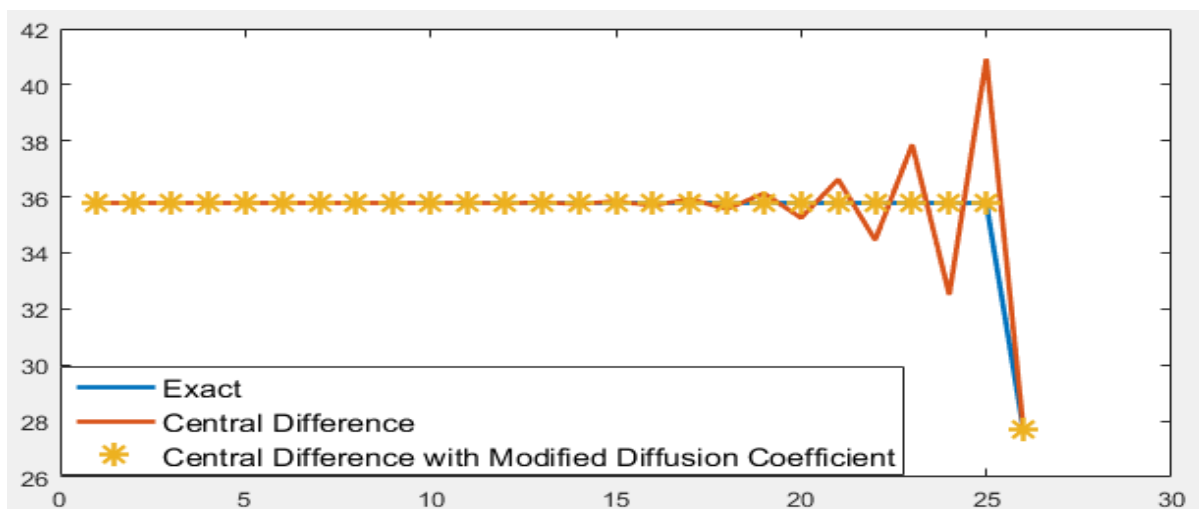

*Figure 2.4: Prediction of exact solution by adding artificial diffusion.*

$\mu_{artificial}$ is a function of a, b and $\Delta$x. In fact, the function is very similar to the activation function tanh(x). We designed few architectures which were essentially fed with different combinations of these three input variables. For instance, (a and Pe), (a, b and Pe), (a, b and $\Delta$x) etc. These are called different representations for input. We would actually create a function estimator. This experiment will help us to figure out how easy it is to create a surrogate model to estimate the artificial diffusion. If such a model works well, it could be used to find the error in estimation through an extended variant of our model.

### 2.3.2. OBSERVATIONS

On training the model over sufficient number of cases the regression would give sort of acceptable result. The correlation coefficient for the regression is not much appreciable. Though the obtained artificial diffusion coefficient helps in stabilizing the CDS, but it is not a good consideration when we have an intention for developing an alternate approximation method.

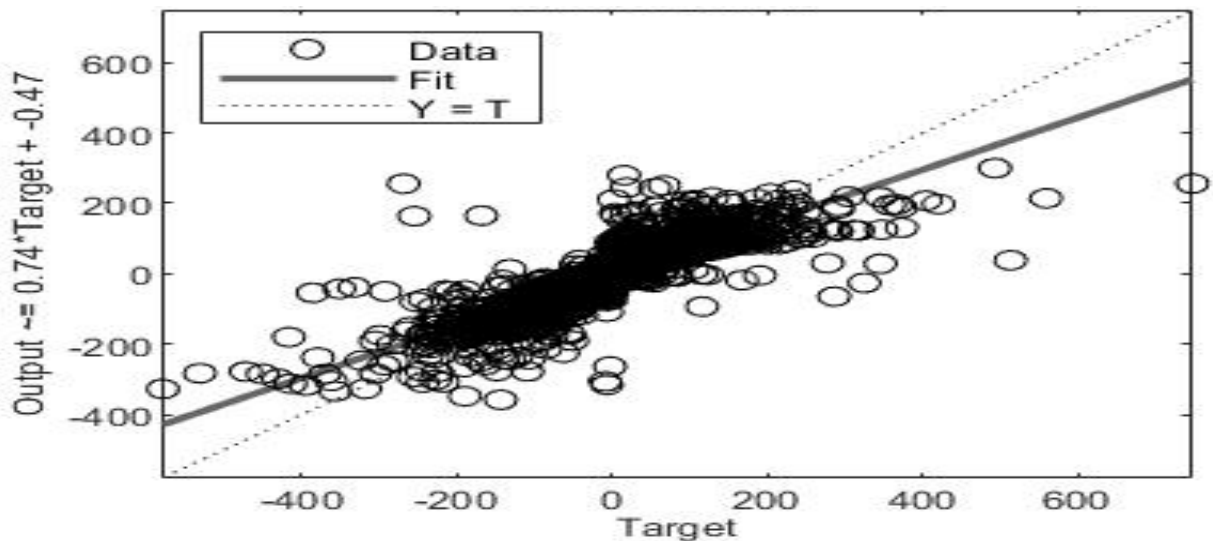

*Figure 2.5: Regression plot for prediction with the architecture [3, 50, 30, 1].*

| Correlation Coefficient | Representations | | |
|---|---|---|---|
| | a, Pe | a, b, $\Delta$x | a, b, Pe |
| 25 neurons | 0.63 | 0.68 | 0.8 |
| 50 neurons | 0.72 | 0.74 | 0.87 |
| 150 neurons | 0.76 | 0.79 | 0.96 |

*Table 2.1: Correlation coefficient of regression for different number of neurons.*

Dependence of physical parameters in the exact solution is unknown for most of the problems. It is not easy to generate the exact error or artificial diffusion. Training needs to be done with historical data. The prediction seems to improve with addition of layers to the neural network architecture.

## 2.4. FUNCTION APPROXIMATION OF EXACT PHYSICAL SOLUTION

A function approximator for advection diffusion can also be devised by training a single layered neural network with the exact solutions or higher order accurate solutions. If the representation for the map is x (independent variable) to u (dependent variable), the regression is very much perfect but minor oscillations could be seen all around the exact solution. For high Pe cases, larger number of cases were used to train the network.

The benefit of this network is that it provides us a surrogate function for the exact function of the governing differential equation. This function could be repeatedly used once it is trained. The drawback of this method is the requirement of highly accurate solution for training. That means the problem has to be solved using known exact function or using higher order method. It is not at all an alternative approximation method in true sense.

## 2.5. CONCLUSION

Data driven methods are useful for solution enhancement, function approximation and error estimation. No doubt, these are strong abilities but generation of data requires us to use conventional methods. The scope for direct reduction in computational expense is not possible if a method requires a plenty of data to train upon. For an alternate approximation method, it is essential to cut short the use of accurate data. There is a need for seeking methods that do not rely much on available solution. So, the physics informed learning comes into picture. In has been explored in depth in upcoming chapters.

# CHAPTER 3
# PHYSICS INFORMED LEARNING

## 3.1. PHYSICS INFORMED NEURAL NETWORK FOR SOLVING ODE AND PDE

The usual neural networks are based on data that are fed. The model just tries to fit into the data. There is no need for understanding the physics behind the problem to train a neural network. This is the reason why there are small noise like oscillations around the exact solution. It is essential to make the neural network function learn the physics of the problem.

Solving a regression problem is an optimization problem. The easiest way to introduce the physics into the problem is to somehow introduce it in the optimization objective. I.E. Lagaris et. al. in 1997 suggested a way to do so. Solving a differential equation is treated as an optimization problem over a neural network function and not really treated as a training problem. The beauty of this algorithm is that, it does not require any data to get the required neural network trained. Only the values of independent variables are fed for training. The initial and boundary conditions are also required to be fed into the optimization objective.

## 3.2.1. LAGARIS' ALGORITHM

A function approximator for any differential equation is proposed to be build. It should take in the values of independent variables like x, y, z etc. The method suggested by Lagaris et. al. in 1997 uses a single layered neural network with sigmoid activation function to satisfy the differential equation. The residual after plugging in the neural network function ($N(\vec{x_i}, \vec{p})$) is a good measure of error. To inform the boundary constraints a trial function ($\psi$) is proposed to be plugged in, instead of plugging in the neural network function directly. The trial function is a mathematically transformed function of the neural network function that satisfies the constraints exactly. A properly trained network would produce extremely low residual. Minimization of this residual ($G_i$) is a beautiful way to impose physics into the neural network. So, the least squared sum of residual over the grid points is the loss function. The points over which the network is trained are called collocation points.

$$L = \min_{\vec{p}} \sum G^2\left(\vec{x_i}, \psi(\vec{x_i}, \vec{p}), \nabla\psi(\vec{x_i}, \vec{p}), \nabla^2\psi(\vec{x_i}, \vec{p})\right)$$

$$\psi(\vec{x}, \vec{p}) = A(\vec{x}) + F(\vec{x}, N(\vec{x}, \vec{p}))$$

The parameters ($\vec{p}$) contains the weights and biases for the network. The gradients are calculated using the AutoGrad technique, easily available on TensorFlow now a days.

### 3.2.2. LAGARIS' IMPLEMENTATION OF ADVECTION DIFFUSION

The base problem of steady one-dimensional advection diffusion equation is solved under Dirichlet boundary conditions $\{u(0) = u_L, u(1) = u_R\}$. So, the trial function and the loss function are:

$$\psi(x,\vec{p}) = (1-x)u_L + xu_R + x(1-x)N(x,\vec{p})$$

$$L(\vec{p}) = \sum \left[\epsilon \frac{\partial^2 \psi(x,\vec{p})}{\partial x^2} - \frac{\partial \psi(x,\vec{p})}{\partial x}\right]^2$$

Regular optimizers like gradient descent could be used as regular neural network. It works really well for $\epsilon > 0.6$ but with decrease in the value the estimation deviates from the exact solution. The boundaries are always intact since we have forced them in the trial function. If $\epsilon$ is very low like 0.1 or less then the model fails and it is unable to converge. So, keeping the boundaries in place makes it difficult for optimization. This is a serious problem and there will be multiple evidences in further chapters as well. Not much improvement could be noticed with increased number of neurons in the hidden layer. Since, there are no much other parameter which could be controlled, it proves the incapability of the method to solve such problems.

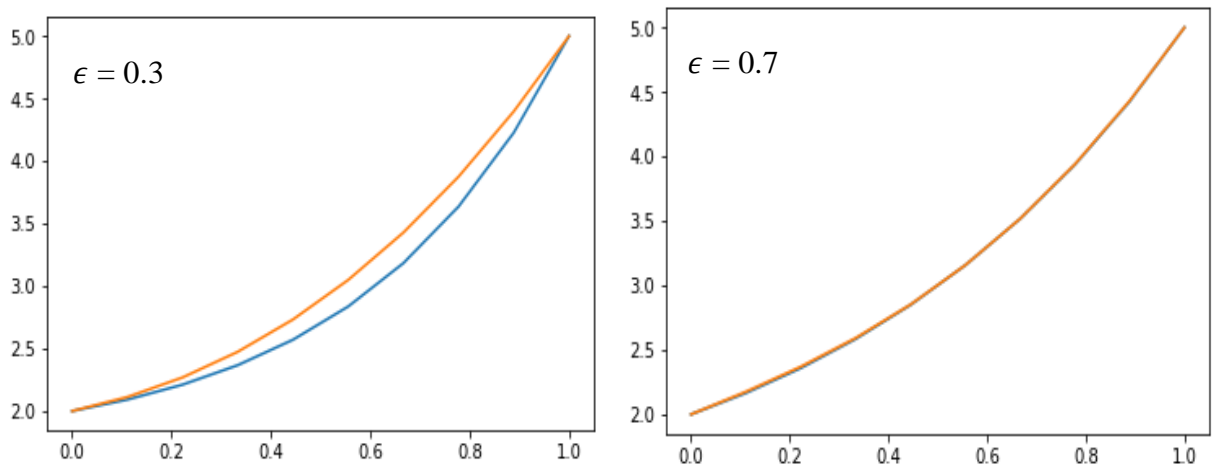

*Figure 3.1: Comparison of prediction using Lagaris' concept for $\epsilon$ = 0.3 and 0.7.*
 *{blue curve represents the exact solution while orange curve is the prediction}*
 *The curves on the right seems to be overlapping on each other.*

Lagaris' method can be used for a wide class of differential equations but it cannot provide solution for extreme cases, like in the case advective dominance. A strong limitation of this approach is that the trial function needs to defined appropriately for each problem. This makes the model less generalizable. The other major limitation is the inability to solve time variant problems. Solving initial value problems using this method seems to be very inefficient.

### 3.3.1. PHYSICS INFORMED DEEP LEARNING BY MAZIAR RAISSI ET.AL.

Maziar Raissi et. al., in 2017, came up with a development over the method proposed by Lagaris et. al. The proposed method addresses the limitations of Lagaris' method that has been mentioned at the end of the last segment. The core improvements are:

1. Additional loss terms are computed as mean squared error at boundaries to satisfy the boundary conditions. The makes the method generalizable to almost all the classes of ODEs and PDEs. Numerous boundary conditions could be enforced as additional mean squared error loss terms. There is no need for altering the trial function, the neural network itself can do the job of trial function.
2. The second benefit of using additional loss term for boundary constraints is allowing relaxation at boundaries during the course of optimization makes the convergence easier and simpler. Though there will be slight deviation from boundary values but as the optimization converges the values start matching the exact values very accurately.
3. Initial conditions could be imposed by addition of extra loss terms as mean of squared errors of the initial value data, very much like boundary constraints. This improvement makes it usable to solve time variant problems like advection and burgers.
4. The architecture in this model allows deeper layers, which makes it possible to solve complex and non-linear functions.

The formulations for trail function ($\psi$), loss function (L), collocation loss($L_f$), boundary constraint loss ($L_b$) and initial constraint loss ($L_i$) are:

$$\psi(x, t, \vec{p}) = \text{Neural } Network(x, t, \vec{p})$$

$$L_f(\vec{p}) = \frac{1}{N_f} \sum \left[ f_{DE}\left(\frac{\partial^n \psi(x,t,\vec{p})}{\partial x^n}, \ldots, \frac{\partial^2 \psi(x,t,\vec{p})}{\partial x^2}, \frac{\partial \psi(x,t,\vec{p})}{\partial x}\right) \right]^2$$

$$L_b(\vec{p}) = \frac{1}{N_b} \sum \left[\psi(x_b^k, t, \vec{p}) - \psi_b^k\right]^2$$

$$L_i(\vec{p}) = \frac{1}{N_i} \sum \left[\psi(x_i^k, t, \vec{p}) - \psi_i^k\right]^2$$

$$L(\vec{p}) = L_f(\vec{p}) + L_b(\vec{p}) + L_i(\vec{p})$$

$N_f$, $N_b$ and $N_i$ are numbers of collocation points, boundary points and initial points respectively.

### 3.3.2. PINN IMPLEMENTATION FOR ADVECTION DIFFUSION

For advection diffusion,

$$f_{DE} = \epsilon \frac{\partial^2 \psi(x,t,\vec{p})}{\partial x^2} - \frac{\partial \psi(x,t,\vec{p})}{\partial x}$$

PINN's implementation makes it possible to solve problems for $\epsilon > 0.14$. This is a remarkable improvement, since the previous method could produce solution for $\epsilon > 0.6$. Practically speaking, $\epsilon > 0.1$ is not a sufficient range.

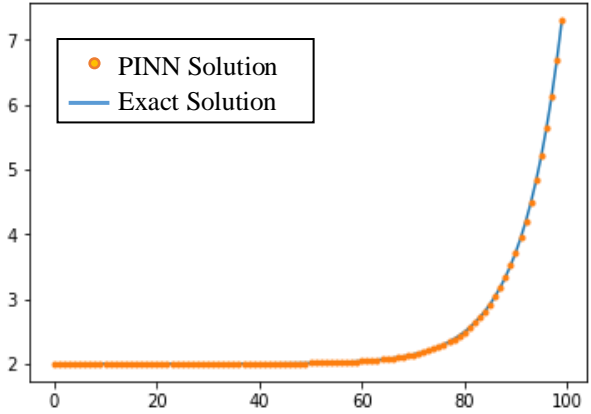

| Architecture | | Valid for: |
|---|---|---|
| Layers | Neurons per Layer | |
| 1 | 2 | 1/e < 3.8 |
| 2 | 2 | 1/e < 4.8 |
| 3 | 2 | 1/e < 6.3 |
| 3 | 3 | 1/e < 6.5 |
| 3 | 10 | 1/e < 6.6 |

*Figure 3.2: Prediction using PINN for $\epsilon$ = 0.15*     *Table 3.1: Range of $\epsilon$ for different architectures*

Keeping the number of neurons per layer high and keeping very deep architecture do not seem to increase the capability of this method. Unless a method works for extreme cases, the usability gets very much restricted. The next chapter proposes a novel method to increase the range of $\epsilon$ for prediction of extremely advection dominant cases.

# CHAPTER 4
# DISTRIBUTED PHYSICS INFORMED NEURAL NETWORK (DPINN)

## 4.1. INTRODUCTION

The available methods cannot provide solution for $\epsilon < 0.1$. The optimization becomes very tough for the method because for extremely low $\epsilon$ the architecture could not capture the sharp gradient that appears quite sudden in the course of domain. To reduce the burden over a network it has been decided to go for a series of small networks. The idea is to fit the equation piece-wise. It is analogous to piece wise curve fitting. Each piece is essentially a sub-domain and there will be a network for each sub-domain to approximate for the sub-domain.

## 4.2. DATA DRIVEN DISTRIBUTED NEURAL NETWORK

The basic test for attempting piece-wise approximation is done by fitting into the physical solution. This is done by taking a series of typical neural network fed with input and output data. It gives good result with minor oscillations around the exact solution, which is typical to happen. It was observed that the fitting was quite good in fact with lesser number of training cases one could get a good fit compared to undivided regression. This gives a motivation to continue over the idea of splitting the domain in physics informed approach.

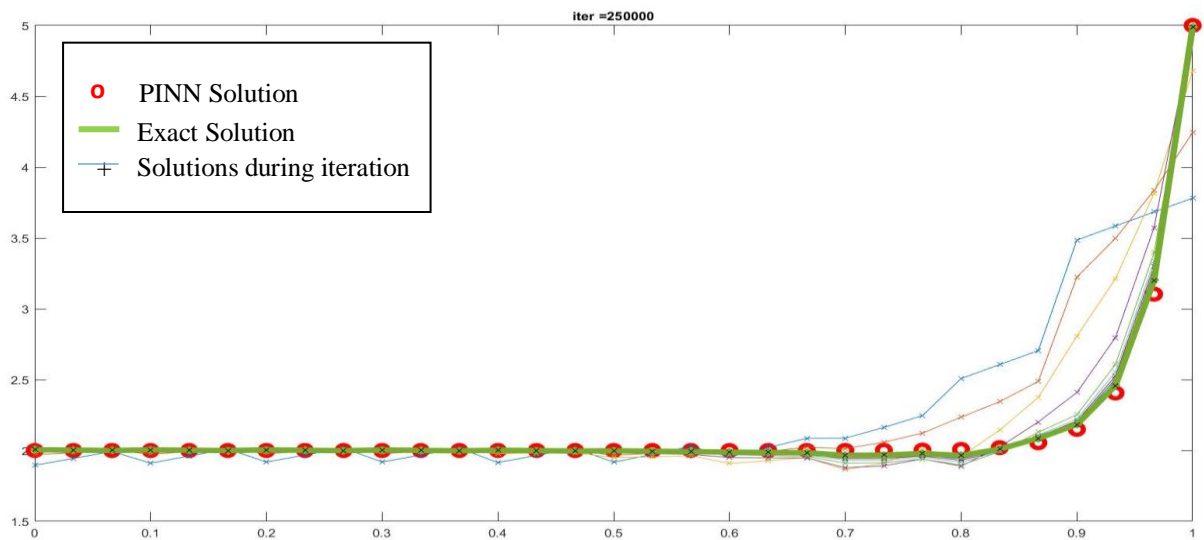

*Figure 4.1: Prediction using distributed neural network for $\epsilon = 0.03$*

It works well because each network has to approximate a part of the curve. So, each network has to approximate a simpler function.

## 4.3. DISTRIBUTED PHYSICS INFORMED NEURAL NETWORK (DPINN)

Distributed physics informed neural network is essentially piece-wise physics informed learning. If the domain is divided into N sub-domains then N neural networks are considered for regression over each sub-domain. In the previous method, physics was informed by forcing the ODE/PDE at collocation points and forcing the constraints like boundary conditions and initial conditions. But since the domain is split, the interfaces need to be bound by constraints like continuity and differentiability.

| | | | | |
|---|---|---|---|---|
| $i = 1$, $j = 1$ | $i = 2$, $j = 1$ | .. | $i = N_{bx}$, $j = 1$ | $t_f$ |
| $i = 1$, $j = 2$ | $i = 2$, $j = 2$ | .. | $i = N_{bx}$, $j = 2$ | |
| .. | .. | .. | .. | |
| $i = 1$, $j = N_{bt}$ | $i = 2$, $j = N_{bt}$ | .. | $i = N_{bx}$, $j = N_{bt}$ | $t_i$ |
| $x_L$ | | | | $x_R$ |

*Figure 4.2: Split of domain into blocks for a two-variable system*

The experiments for the thesis are conducted with only two neurons over a single hidden layer. The activation function used in the neurons are 'tanh'. So, the trial function for each network is formulated as:

$$\boldsymbol{\phi}_{ij}(x_{ij}, t_{ij}, W_{ij}, B_{ij}) = W^2_{ij}.tanh(W^{1,1}_{ij}.x_{ij} + W^{1,2}_{ij}.t_{ij} + B^1_{ij}) + B^2_{ij}$$

where $\boldsymbol{\phi}_{ij}$ is the prediction over the i$^{th}$ sub-domain in x for i = 1, 2, .. , $N_{bx}$ and the j$^{th}$ sub-domain in t for i = 1, 2, .. , $N_{bt}$

Apart from forcing the differential equation at collocation points through mean of squared residual ($L_f$) and forcing the boundary points and initial points by mean of squared errors ($L_b$ and $L_i$) at boundaries and initial points respectively. For continuity and differentiability at interfaces in the domain of x, loss terms ($L_{vm,x}$ and $L_{sm,x}$) are introduced as mean square difference of values and slopes at the interfaces in domain of x. So, the gross loss function (L) is sum of these partial loss terms.

$$L(\vec{p}) = L_f(\vec{p}) + L_b(\vec{p}) + L_i(\vec{p}) + L_{vm,x}(\vec{p}) + L_{sm,x}(\vec{p})$$

$$L_f(\vec{p}) = \frac{1}{2m}\sum_{i=1}^{N_{bx}} \sum_{j=1}^{N_{bt}} \sum_{p=1}^{m} \left( f\left(\frac{\partial^n \phi_{ij}^p}{\partial x^n}, \ldots, \frac{\partial^2 \phi_{ij}^p}{\partial x^2}, \frac{\partial \phi_{ij}^p}{\partial x}, \frac{\partial \phi_{ij}^p}{\partial t}\right) \right)^2$$

$$L_b(\vec{p}) = \frac{1}{2}\sum_{j=1}^{N_{bt}} \left[ \left(\phi_{1,j}^1 - \phi_{exact(L,j)}\right)^2 + \left(\phi_{Nbx,j}^m - \phi_{exact(R,j)}^m\right)^2 \right]$$

$$L_i(\vec{p}) = \frac{1}{2}\sum_{i=1}^{N_{bx}} \sum_{p=1}^{m} \left(\phi_{i,1}^p - \phi_{exact(i,1)}^p\right)^2$$

The net loss function is minimized to produce the approximation. On testing over steady advection-diffusion, unsteady advection and non-linear burgers equations, it performs better than existing methods. The extreme cases for advection-diffusion problem are discussed in details in upcoming segments.

### 4.4.1. INTERFACE CONDITIONS: VALUE MATCHING

The loss term for value matching at interfaces is very important. It ensures that the predictions at interfaces due to each network matches to make the regression continuous. The loss term is the mean-square of the difference in the predictions at the interfaces. Value matching criterion is a must for keeping the physics intact. It transforms the problem to be solved as multiple boundary value problems where intermediate boundary values must be shared. So, without this criterion physics would fail.

The formulation of loss term for continuity at interfaces is cumulated square difference in prediction:

$$L_{vm,x}(\vec{p}) = \frac{1}{2}\sum_{i=1}^{N_{bx}-1} \sum_{j=1}^{N_{bt}-1} \left(\phi_{ij}^m - \phi_{i+1,j}^1\right)^2$$

### 4.4.2. INTERFACE CONDITIONS: SLOPE MATCHING

The loss term for slope matching at interfaces is not always a necessary condition. It ensures that the approximate function is differentiable at interfaces. For problems with shocks and very sharp gradients, solution is highly diffused. During experiments, it is clearly evident in linear unsteady advection of heaviside function or square pulse. While for continuous and differentiable functions, using this additional loss term smoothens the optimization. Even using slope matching loss term helps in equations with diffusion terms like advection-diffusion and burgers.

The formulation of loss term for differentiability at interfaces is cumulated square difference in derivative:

$$\mathbf{L}_{\text{sm,x}}(\vec{p}) = \frac{1}{2} \sum_{i=1}^{N_{bx}-1} \sum_{j=1}^{N_{bt}-1} \left( \frac{\partial}{\partial x} \phi_{i,j}^m - \frac{\partial}{\partial x} \phi_{i+1,j}^1 \right)^2$$

In case of advection dominant problems, using this additional loss term is a great help. $L_{\text{sm,x}}$ becomes kind of necessary for such problems since it helps in increasing the range of $\epsilon$ over which method could approximate.

### 4.4.3. INTERFACE CONDITIONS: SECOND DERIVATIVE MATCHING

The formulation of loss term for second derivative at interfaces is cumulated square difference in second derivative:

$$\mathbf{L}_{\text{sdm,x}}(\vec{p}) = \frac{1}{2} \sum_{i=1}^{N_{bx}-1} \sum_{j=1}^{N_{bt}-1} \left( \frac{\partial^2}{\partial x^2} \phi_{i,j}^m - \frac{\partial^2}{\partial x^2} \phi_{i+1,j}^1 \right)^2$$

This additional loss term does not seem to help at all. In fact, this makes the convergence more difficult. This may help in convergence of higher order problems.

### 4.4.4. INTERFACE CONDITIONS: FLUX MATCHING

Flux is conserved throughout the domain. Hence, equating flux at interface is a way to inform the continuity of flux. Flux is calculated as the integral of the governing equation. Continuity and differentiability are essentially continuity of individual terms in the flux equation. So, flux equation determines the highest order of partial derivative that could be matched at interface. This suggests that highest order of advection equation is zero while for advection-diffusion and burgers, highest order of partial derivative is one. Formula for flux term is:

$$F_{ij}^p = \int f \left( \frac{\partial^n \phi_{ij}^p}{\partial x^n}, \ldots, \frac{\partial^2 \phi_{ij}^p}{\partial x^2}, \frac{\partial \phi_{ij}^p}{\partial x}, \frac{\partial \phi_{ij}^p}{\partial t} \right) dx$$

The formulation of loss term for flux continuity at interfaces is cumulated square difference in flux:

$$\mathbf{L_{fm,x}}(\vec{p}) = \frac{1}{2} \sum_{i=1}^{N_{bx}-1} \sum_{j=1}^{N_{bt}-1} \left( F_{ij}^m - F_{i+1,j}^1 \right)^2$$

This additional loss term has nearly similar effect of slope matching at interfaces. Slope matching and flux matching could be used interchangeably. Keeping or discarding this loss term does not seem to make significant change in convergence.

## 4.5. DPINN IMPLEMENTATION OF STEADY ADVECTION DIFFUSION

Steady advection diffusion is a time invariant problem. Hence, number of inputs is only one (x). The architecture used for each block in the domain for two neurons per hidden layer is:

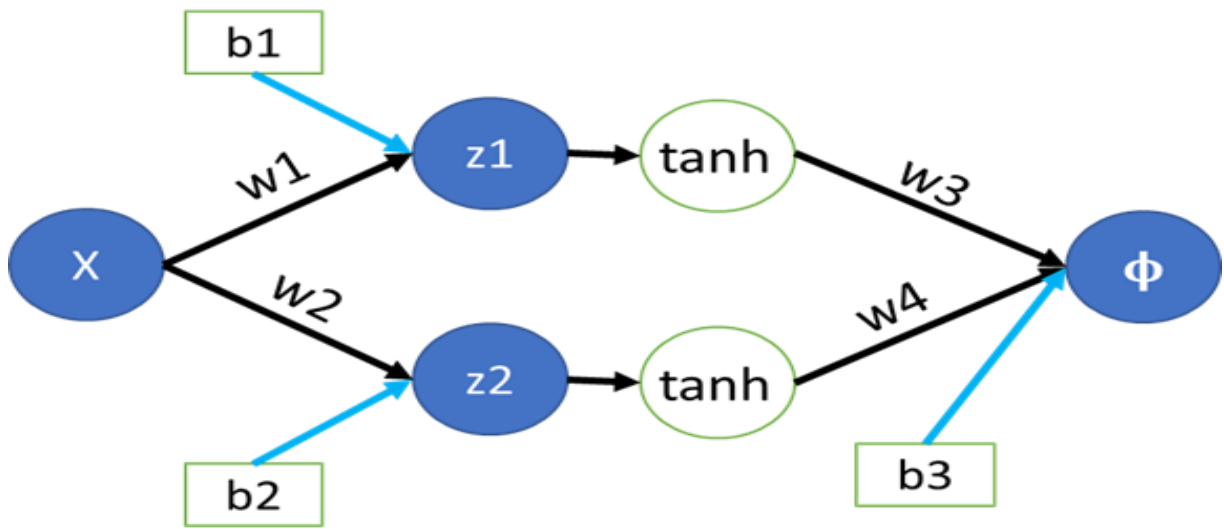

*Figure 4.3: Neural network architecture for each block in DPINN for a specific case.*

The neural network function for each block:

$$\boldsymbol{\phi}_i(x_i, w_i, b_i) = w3_i.tanh(w1_i.x_i + b1_i) + w4_i.tanh(w2_i.x_i + b2_i) + b3_i$$

where $\boldsymbol{\phi}_i$ is the prediction over the $i^{th}$ sub-domain for i = 1, 2, .., N

The loss terms are:

$$\mathbf{L} = \mathbf{L_f} + \mathbf{L_b} + \mathbf{L_{vm}} + \mathbf{L_{sm}}$$

$$L_f = \frac{1}{2m}\sum_{i=1}^{N} \sum_{p=1}^{m} (\varepsilon \frac{\partial^2 \phi_i^p}{\partial x^2} - \frac{\partial \phi_i^p}{\partial x})^2$$

$$L_b = \frac{1}{2}(\phi_1^1 - \phi_L)^2 + \frac{1}{2}(\phi_N^m - \phi_R)^2$$

$$L_{vm} = \frac{1}{2}\sum_{i=1}^{N-1} (\phi_i^m - \phi_{i+1}^1)^2$$

$$L_{sm} = \frac{1}{2}\sum_{i=1}^{N-1} (\frac{\partial}{\partial x}\phi_i^m - \frac{\partial}{\partial x}\phi_{i+1}^1)^2$$

Even with shallower networks, DPINN could approximate for lower $\epsilon$ when compared with PINN and Lagaris' approach. DPINN approach gives more flexibility to the networks to accommodate complex regressions. With just single hidden layer with two neurons the model could easily approximate if $\epsilon > 0.14$. For lower values, the interface constraints and boundary constraints seem to be relaxed and steps appear at the interfaces. With increase in the number of loss terms the optimization seems to get tougher. The loss terms seem to compete with each other for lower $\epsilon$.

### 4.6. SUMMARY OF DPINN

Though Lagaris' algorithm could generate a good function approximator but it fails for advection dominant problems. Then arrives physics informed deep network by M. Raissi et. al. which could be used for advection dominant problems but it could not approximate cases for highly advective problems even with deep neural architecture. DPINN performs better than the above two methods even with a single layer architecture with only two neurons in the hidden layer.

Still it has been difficult to approximate extreme cases. Due to the involvement of multiple loss terms, the problem of optimization has become a tough multi-objective optimization problem. This makes the problem difficult to converge to the required solution, Pareto optimal solution. The optimization faces saddle points or local optimums which keeps obstructing the convergence strongly. This makes the learning process take much larger number of iterations to converge than it should ideally take.

# CHAPTER 5
# EXPERIMENTS TO ENHANCE DPINN

## 5.1. INTRODUCTION

In this chapter various reports of various trial and experiments conducted to enhance the DPINN algorithm are recorded topic wise.

## 5.2. RANDOM COLLOCATION POINTS FOR FORCING THE EQUATION

Overfitting is evident in the vicinity of shocks and sharp gradients. It could be seen in around the edges square pulse in linear advection. Initially, uniformly domain points were fed for forcing the differential equation at only specific points as a result the behaviour of the function at neighbouring points could not be captured.

New set of random points are generated within each block before each iteration of optimization. Considering random points within blocks helps in protecting the network from over fitting. It also helps in speeding up the convergence. The predictions at edges becomes neater. It should be noted that, forcing the differential equation at end points perform better than otherwise.

## 5.3. EFFECTS OF ACTIVATION FUNCTIONS

| **SIGMOID** | **TANH** | **EXPONENT** |
|---|---|---|
| $S(x) = \frac{1}{1+e^{-x}}$ | $T(x) = \frac{1}{1+e^{-x}}$ | $E(x) = e^x$ |
| Can be used for regression. | Regression results are very appreciable and better than sigmoid. | Regression gives best result because exact solution matches with the equation form of advection diffusion. |
| Can be generalised for different cases. | Can be generalised for different cases. | Can't be generalised easily. |

*Table 5.1: Comparison between various activation function for DPINN*

## 5.4. REGULARIZATION

Upon looking into the patterns of the weights and biases for each panel it was noticed that:

- wherever the gradients are low, the weights and biases are low and none of them has extremely high value of higher order
- however, where the gradients are high, few of the weights and biases have much higher value than others
- The appearance of few large weights and biases could be sometimes noticed for low Pe cases but they are mostly observed in case of highly advection dominant cases. They destabilize the optimization.

**Weight & Biases Matrices for two neurons in a hidden layer**

| $\epsilon = 0.42$ (without regularisation) | | | |
|---|---|---|---|
| 1.7805 | -1.7912 | -1.8084 | w1 |
| -1.5659 | 2.1673 | 2.7765 | b1 |
| -0.2643 | -0.2407 | 0.2207 | w2 |
| -1.1908 | -0.8234 | 0.5371 | b2 |
| 1.8125 | -4.0072 | -8.7473 | w3 |
| -2.2404 | -3.6744 | 7.0361 | w4 |
| 1.8106 | 3.4170 | 7.0328 | b3 |

| $\epsilon = 0.1$ (without regularisation) | | |
|---|---|---|
| -0.3066 | 17.0033 | 2.3722 |
| -0.0113 | 2.3128 | -3.3988 |
| 3.2864 | 0.9377 | -6.8814 |
| -2.5445 | -2.1815 | 7.8949 |
| 0.0131 | 3.3761 | 3.1859 |
| 0.6595 | 5.0087 | -5.6486 |
| 2.8943 | 3.9407 | 11.5863 |

| $\epsilon = 0.14$ (without regularisation) | | | |
|---|---|---|---|
| 1.4509 | 0.4490 | 35.3919 | w1 |
| 0.2307 | 0.3516 | -18.1995 | b1 |
| 1.3852 | -14.6046 | -7.4035 | w2 |
| -0.1523 | -1.8173 | -0.6267 | b2 |
| -0.1508 | 0.4808 | 4.0374 | w3 |
| 0.2830 | -1.5808 | -0.8216 | w4 |
| 2.8100 | 1.7204 | -0.5917 | b3 |

| $\epsilon = 0.14$ (with regularization of 1st layer) | | |
|---|---|---|
| 0.6522 | -1.9664 | 4.5726 |
| 0.1691 | 2.8057 | -5.6534 |
| 0.5617 | 0.3314 | -0.2633 |
| -0.2802 | -0.7681 | -0.6523 |
| -0.3587 (Relaxed Weights) | -8.2259 | 11.2435 |
| 0.4944 | -3.2245 | -5.7162 |
| 2.3281 | 8.4468 | 9.7802 |

*Figure 5.1: Weight matrices for different cases.*

To stabilize the weight matrix, the large weights and biases are penalized by regularization. The regularization is similar to Tikhonov's regularization where an additional term

$$L^* = L_f + L_b + L_{vm} + L_{sm} + \frac{\lambda}{2(i*j*k)}\sum\sum\sum(W_{ij}{}^k)^2$$

There is a need for regularization of weights but it does not help in optimization of very low $\epsilon$.

## 5.5.1. WEIGHED LOSS TERMS

The loss function in case of DPINN is composed of many loss objectives. Linearly weighed sum of objectives is a good way to scalarize multi-objective optimization problem. This is a simple yet effective linear scalarization technique.

$$\min_{x \in X} \sum_{i=1}^{k} w_i f_i(x),$$

$$L^* = w_f L_f + w_b L_b + w_{vm} L_{vm} + w_{sm} L_{sm} + \lambda \sum \sum \sum (W_{ij}^k)^2$$

Weights of each objective can be kept equal to '1' for simplicity. The optimization becomes tougher with increasing number of objectives because of rise in number of local minima and because each objective may have different range and may be scaled as equivalent. There cannot be apple to apple comparison between any two objectives. Normalisation is a way to provide equal weightage to each objective.

Weights of loss terms controls dominance of each objective. The net loss function may not produce the proper minimum (Pareto Optimum) where each objective is minimised. Proper weights need to be assigned for Pareto optimal solution but finding proper weights is a tough task. Assigning weights by tuning through trials helps as well.

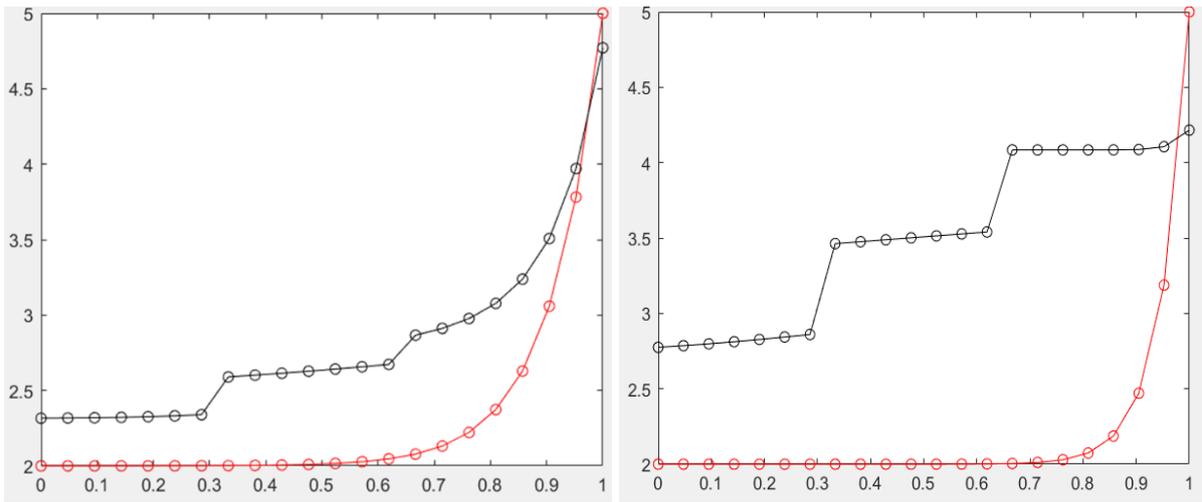

*Figure 5.2: Predictions using DPINN for $\epsilon = 0.1$ and $\epsilon = 0.075$*

In strong convective dominance, struggle between various objectives is evident from the graph. The loss objectives seem to be competing with each other. For highly advective problems, the governing equation fitting is very well satisfied while the value matching at intermediate points and boundary points matching fails heavily. So, penalizing the fitting loss term by dividing the loss term by 10-50 helps. Similarly, prioritizing the boundary point loss and value matching at intermediate points by multiplying the loss term by 10-50 helps.

Error due to slope at intermediate points is not much. Changing the weight for that case does not seem to respond to well to direct toward the actual solution. Somehow, it is subtly stable weight equals to 1.

### 5.5.2. TUNING WEIGHED LOSS TERMS BY TRIALS

Increasing $W_f$ improvises the governing equation fitting. So, it gives the required curvature while $W_{vm}$ and $W_b$ reduces the steps at intermediate points. Proper weights can direct towards convergence but adjusting so many weights manually is a tough job.

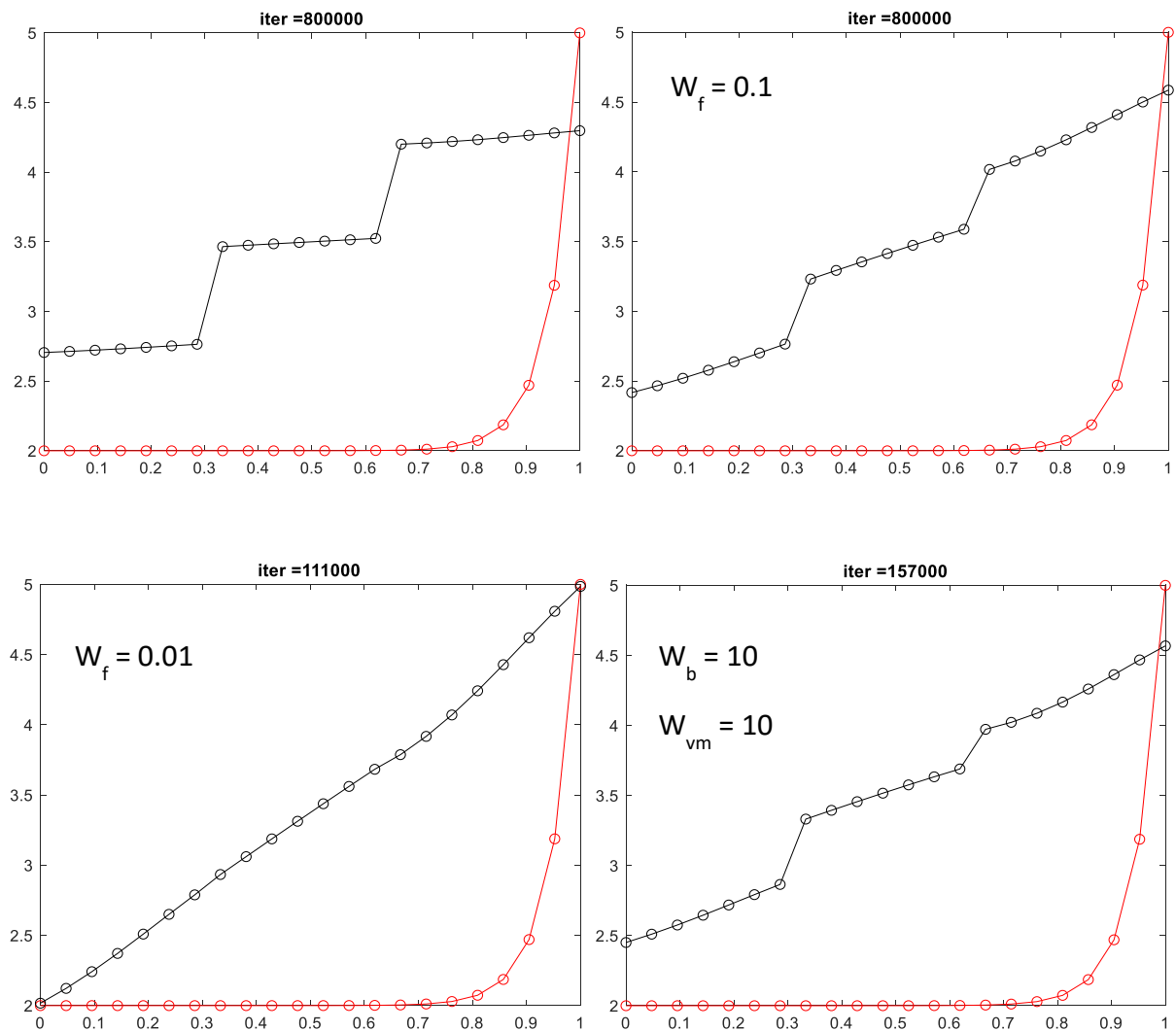

*Figure 5.3: Predictions using DPINN for different weights assigned to loss terms for ϵ = 0.05 and lr = 0.01*

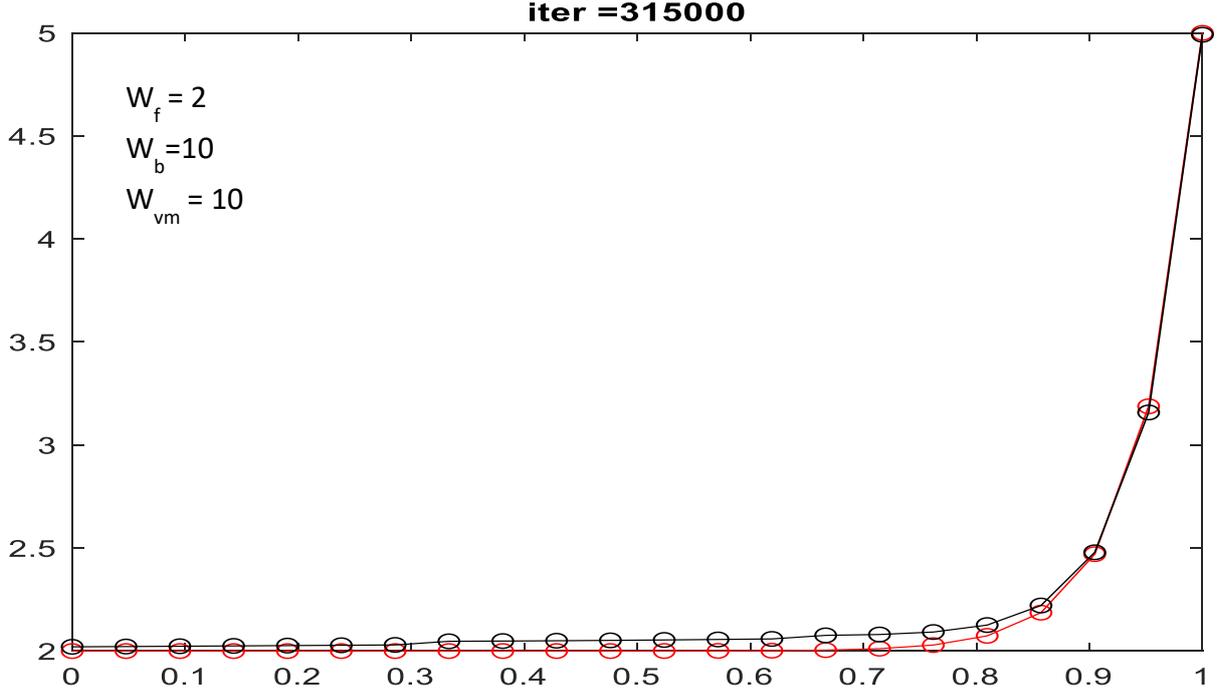

*Figure 5.4: Predictions using DPINN for weights decided by trials assigned to loss terms for $\epsilon$ = 0.05, lr = 0.01*

These trials prove that assigning suitable weights would definitely help the optimization. Each objective must be given appropriate weights such that no objective dominates other.

### 5.5.3. LAGRANGIAN MULTIPLIERS FOR CONSTRAINED OPTIMIZATION

Lagrangian multipliers is a good approach to solve constrained optimization. The weights of loss terms could be appropriately chosen.

$$\mathbf{L}^* = \mathbf{L}_f + \lambda_b \sum l_b + \lambda_{vm} \sum l_{vm} + \lambda_{sm} \sum l_{sm}$$

where $\lambda_b$, $\lambda_{vm}$ and $\lambda_{sm}$ are the Lagrangian multipliers.

$$l_{f,ip} = \varepsilon \frac{\partial^2 \phi_i^p}{\partial x^2} - \frac{\partial \phi_i^p}{\partial x}$$

$$l_{b,k} = \phi_k^b - \phi_b$$

$$l_{vm,i} = \phi_i^m - \phi_{i+1}^1$$

$$l_{sm,i} = \frac{\partial}{\partial x} \phi_i^m - \frac{\partial}{\partial x} \phi_{i+1}^1$$

Normal equations: $\nabla L^* = 0$

$$\Rightarrow \begin{bmatrix} \frac{\partial \sum l_b}{\partial \omega_{ij}} & \frac{\partial \sum l_{vm}}{\partial \omega_{ij}} & \frac{\partial \sum l_{sm}}{\partial \omega_{ij}} \end{bmatrix} [\lambda_b \quad \lambda_{vm} \quad \lambda_{sm}]^T = \begin{bmatrix} -\sum l_f \frac{\partial l_f}{\partial \omega_{ij}} \end{bmatrix}$$

$$\Rightarrow \lambda = (A^T A)^{-1} A^T B$$

The Lagrange's method cannot be exactly imposed in this problem. The result deteriorates in this scheme.

## 5.6. WEIGHT INITIALIZATION BY RECURRENT DOMAIN SPLIT

The domain is divided into one or two pieces and each network is initialised with random weights. The domain splits into double the number of cells (or pieces) and weights are initialised with updated weights from previous case. This is a guided weight initialisation over random initialisation for highly discretized models. So, the chance of convergence significantly increases for heavier discretization. The scheme is not effective for lesser number of panels.

The development of panels and weight initialization is depicted from top to bottom in the figure below.

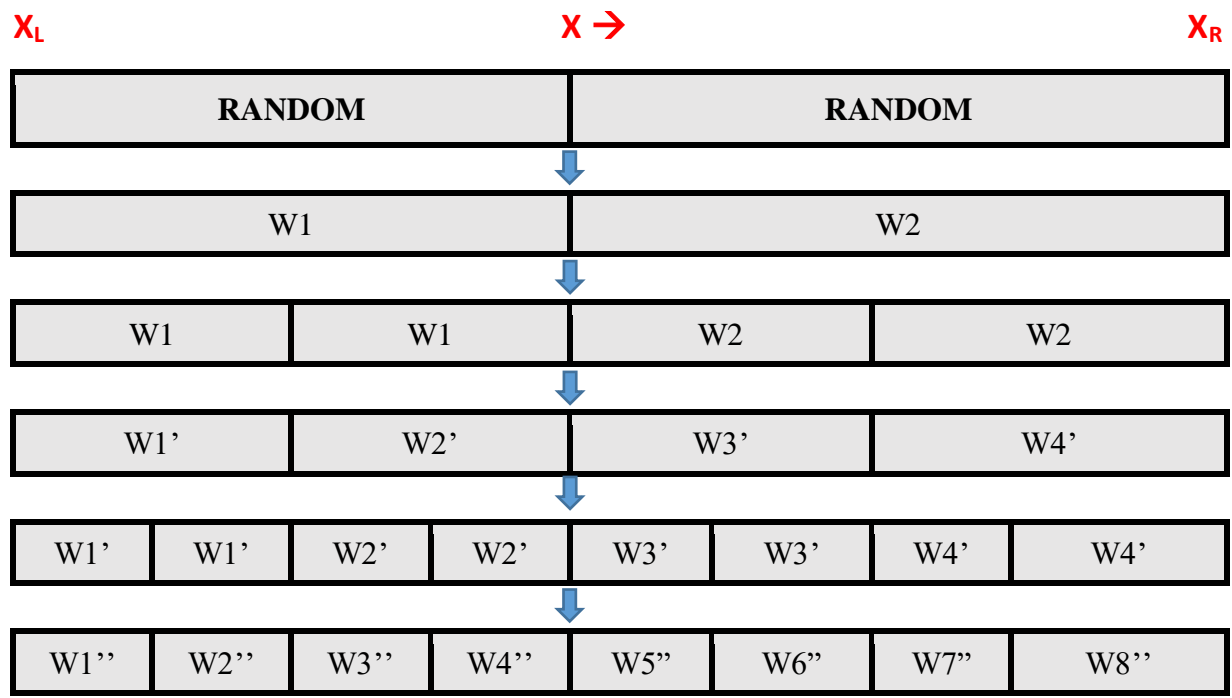

*Figure 5.5: Development of blocks and weight assignment is shown from top to bottom*

## 5.7. GUIDED CONVERGENCE BY UPDATING $\epsilon$ THROUGH ITERATION

The solution for high $\epsilon$ is always accurate. The problems arise as $\epsilon$ reduces to smaller value gradually. The way proposed is to start optimization with descent $\epsilon$ and gradually decrease with every iteration till the required $\epsilon$ reached. This is a way to guide the optimisation and tackle local minimums.

$$\epsilon^* = 1 + \frac{(\epsilon - 1)}{i_{\max}} i$$

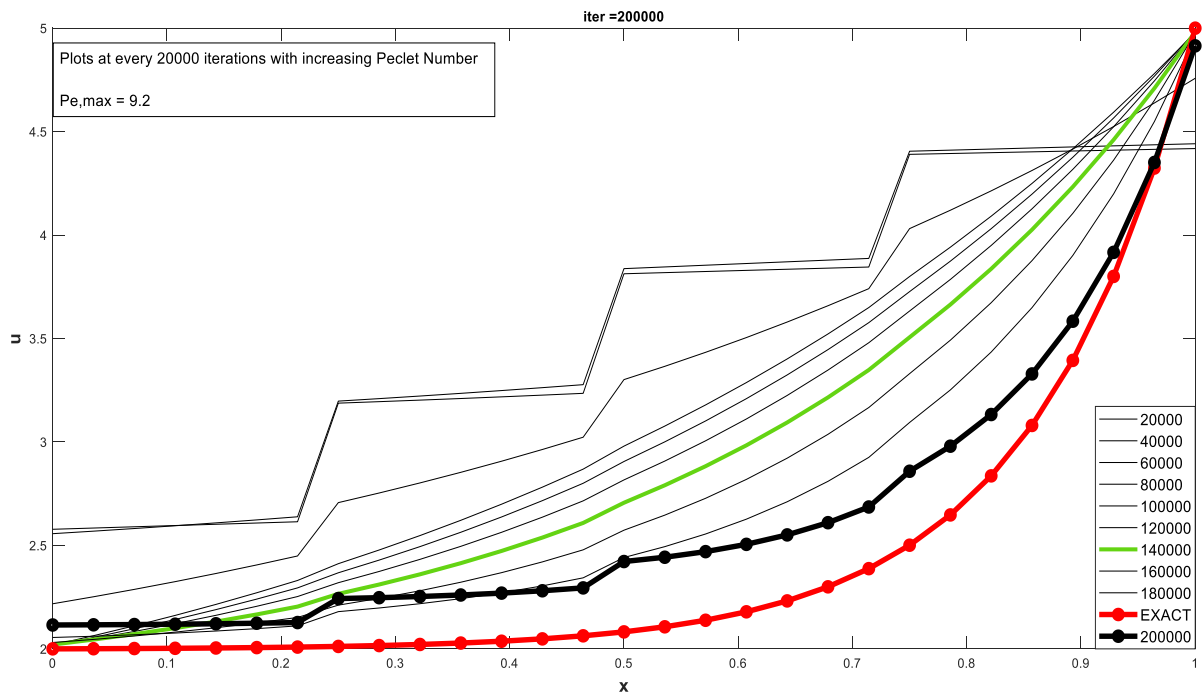

*Figure 5.6: Predictions at various iteration steps*

Initially, the collocation loss is minimised very soon and flat steps are formed. Gradually, the solution converges to a high $\epsilon$. The boundaries and interfaces are properly satisfied as per the loss terms. With decreasing $\epsilon$ the solution starts getting curved gradually with each iteration, as $\epsilon$ reduces. It can be observed that the steps (discontinuities at interfaces) starts appearing as $\epsilon$ starts decreasing. This scheme has a very low impact on optimization. This supports the claim for issues with pareto optimization.

## 5.8.1. MODIFIED TRIAL FUNCTIONS WITH EXTRA LINEAR COMPONENT

Let the trial function have a linear component in the form:

$$\phi(x, \omega, b) = Ax + NN(x, w, b)$$

Where,

$\phi$ : new trial function

NN: old neural network

A: additional weight

The idea behind this approach is to guide the weights update to fit a straight line before the network function starts capturing the trend. Addition of a new term 'Ax' helps the convergence by guiding the weights towards linear fitting at early stage of learning. Can be tried with dominatingly large value of A during weight initialization to increase the impact. Somehow, this approach does not seem to be that effective during optimization

## 5.8.2. MODIFIED TRIAL FUNCTIONS FOR BOUNDARY FORCING

With an inspiration from I.E. Lagaris' approach to exactly force the boundaries, let the trial function be:

$$\phi(x, \omega, b) = A(x) + B(x).NN(x, w, b)$$

Where,

$\phi$ : new trial function

NN: neural network

A(x), B(x): functions to fit boundaries exactly

The case of $\epsilon = 0.1$ could be seen in the figure below. Though boundaries are forced exactly but the problem at interfaces still persists. This model is way slower than the usual DPINN model.

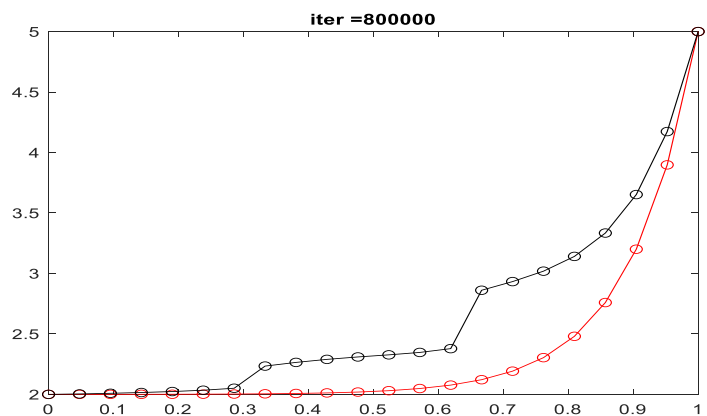

*Figure 5.7: Prediction for $\epsilon = 0.1$ by boundary forcing*

## 5.8.3. MODIFIED TRIAL FUNCTIONS FOR FORCING BOUNDARIES AND INTERFACE MATCHING

With an inspiration from I.E. Lagaris' approach to exactly force the boundaries as well as value matching exactly, let the trial function be:

$$\phi(x, \omega, b) = A(x) + B(x).NN(x, w, b)$$

Where,

$\phi$ : new trial function

NN: neural network

A(x), B(x): fits boundaries and matches interface exactly

Here the trial function for each panel is set such that at interfaces it matches the neighbouring trial value exactly. Similarly, trial function must enforce the boundaries exactly. The case for $\epsilon = 0.1$ could be seen in the figure below.

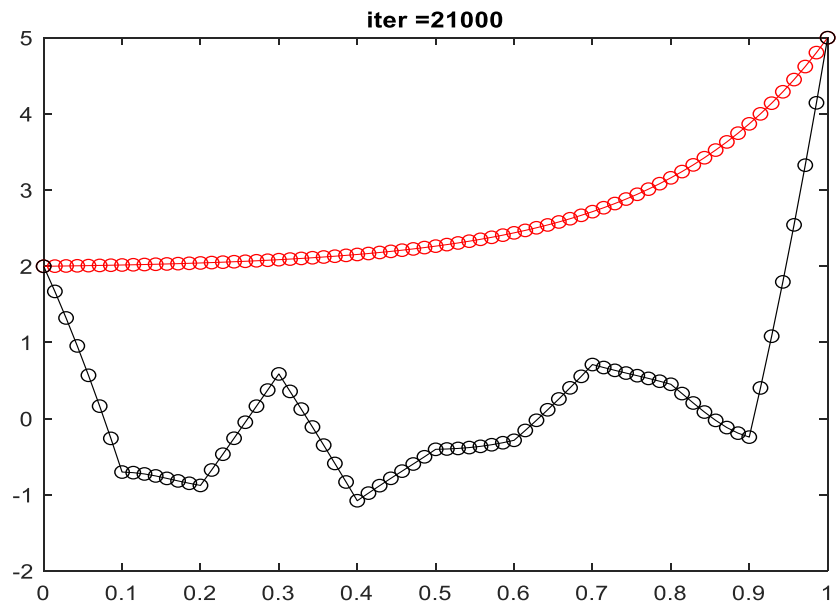

*Figure 5.8: Prediction for $\epsilon = 0.1$ by boundary and continuity forcing*

Though the boundaries and interfaces are forced by setting proper trial function for each panel it could be clearly seen that the collocations points could not fit the differential equation. The regression in each panel do not depict the curve for the concerned differential equation. In fact, slope matching criterion fails as well. Slope matching can be enforced but trial function is tough to find and convergence would be difficult as well. This model is way slower and gradients die out very soon.

## 5.9. EFFECT OF NEGATIVE '$\epsilon$' AND DOMAIN OF 'X'

A strange occurrence that was noticed was, the optimization becomes lot easier on taking negative $\epsilon$ values. The effective equation becomes:

$$|\epsilon|\frac{\partial^2 \phi}{\partial x^2} + \frac{\partial \phi}{\partial x} = 0$$

The convergence happens way faster and very accurate solution is achieved within 2000 to 30000 iterations. Typically, the number of iterations required are in the range of $10^5$ to $3 \times 10^5$. In fact, the range of $|\epsilon|$ is larger if $\epsilon$ is negative. When with positive $\epsilon$ we can get solution for 0.15, while the same method would produce solution for $\epsilon > 0.065$.

By increasing the domain of the problem for say, $\Delta x = 100$, the solution equivalent of Peclet Number 500 to 1500 can be easily achieved with negative Peclet Number.

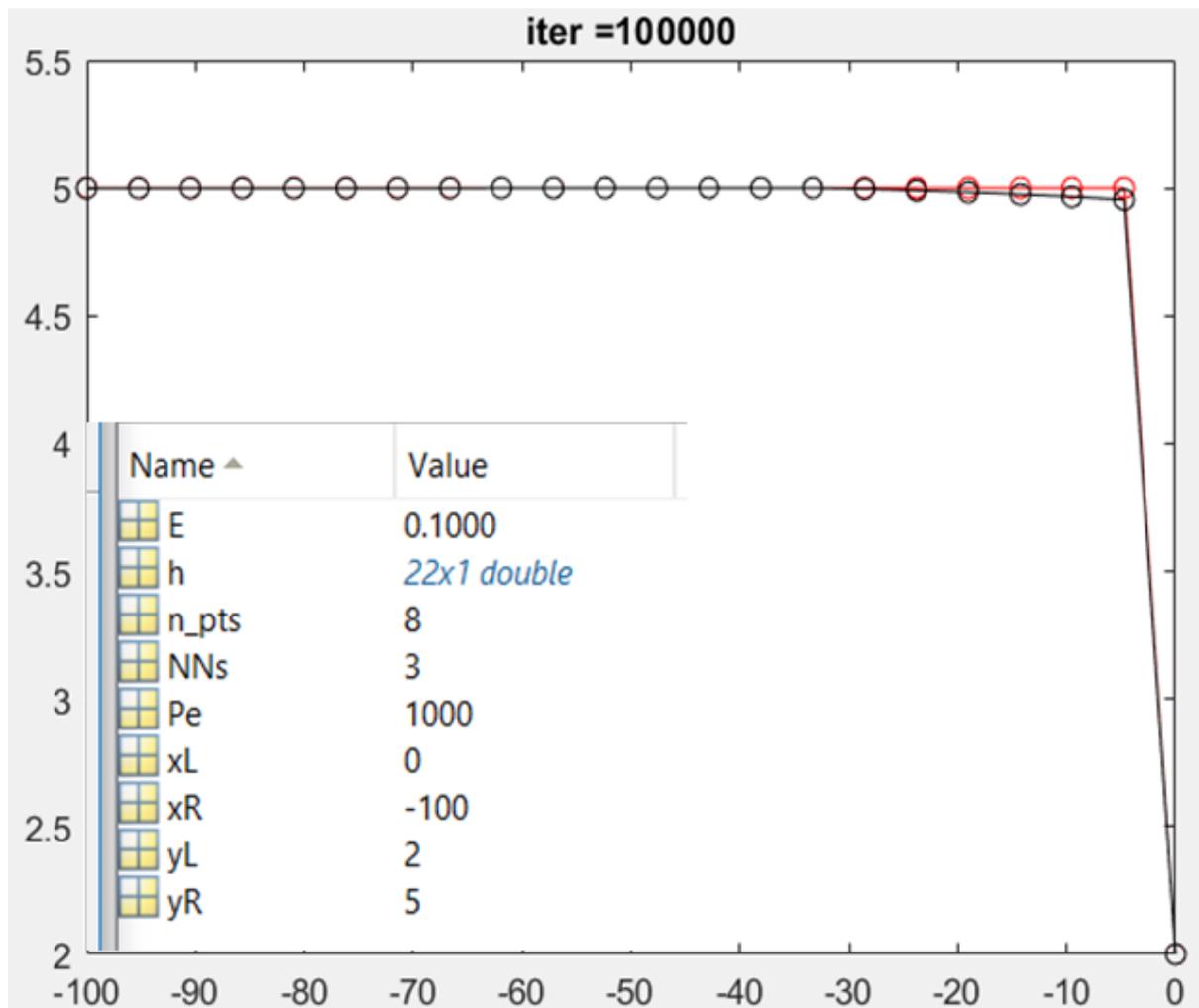

*Figure 5.9: Prediction for negative $\epsilon$ with domain expansion*

## 5.10. USING THE FLUX TERM FOR FORCING AT COLLOCATION POINTS

$$\mathbf{Flux} = [\varepsilon \frac{\partial \phi_i^m}{\partial x} - \phi_i^m]$$

It is not a good option to consider flux instead of the main differential equation for forcing. At least, it detects the curve trend to some extent with many panels. Otherwise, straight stair case used to be formed with many panels. At least it captures the trend to some extent but for low $\epsilon$ trend could not be captured. The curve pattern is different but the boundaries and interfaces conditions are fully satisfied

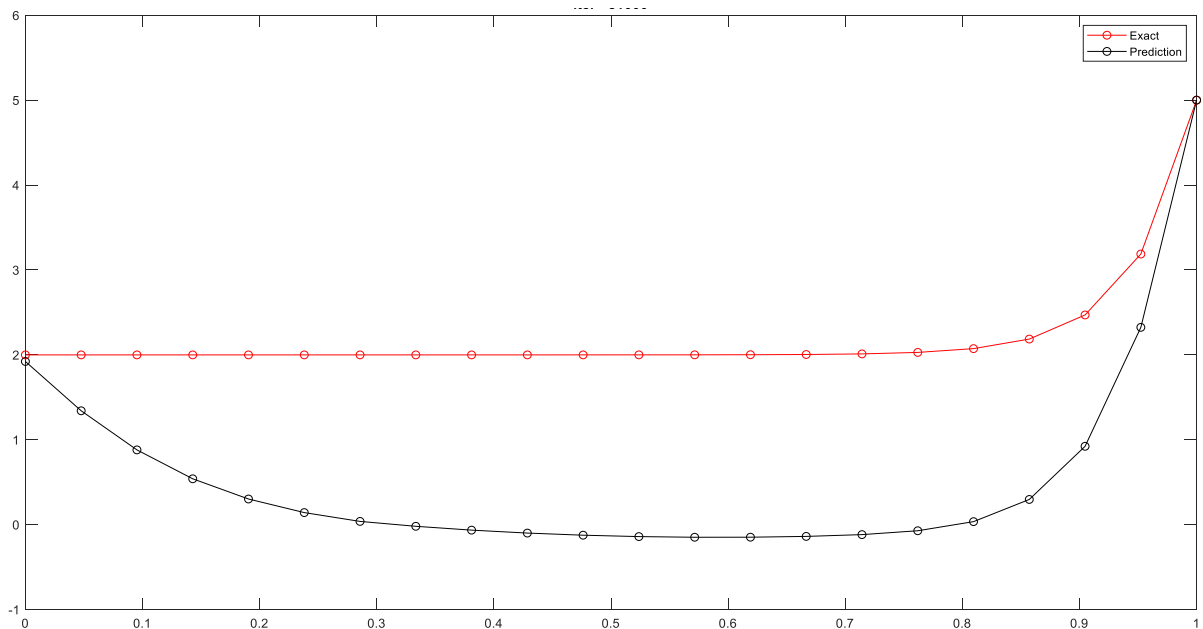
*Figure 5.10: Prediction for flux collocation two points/block*

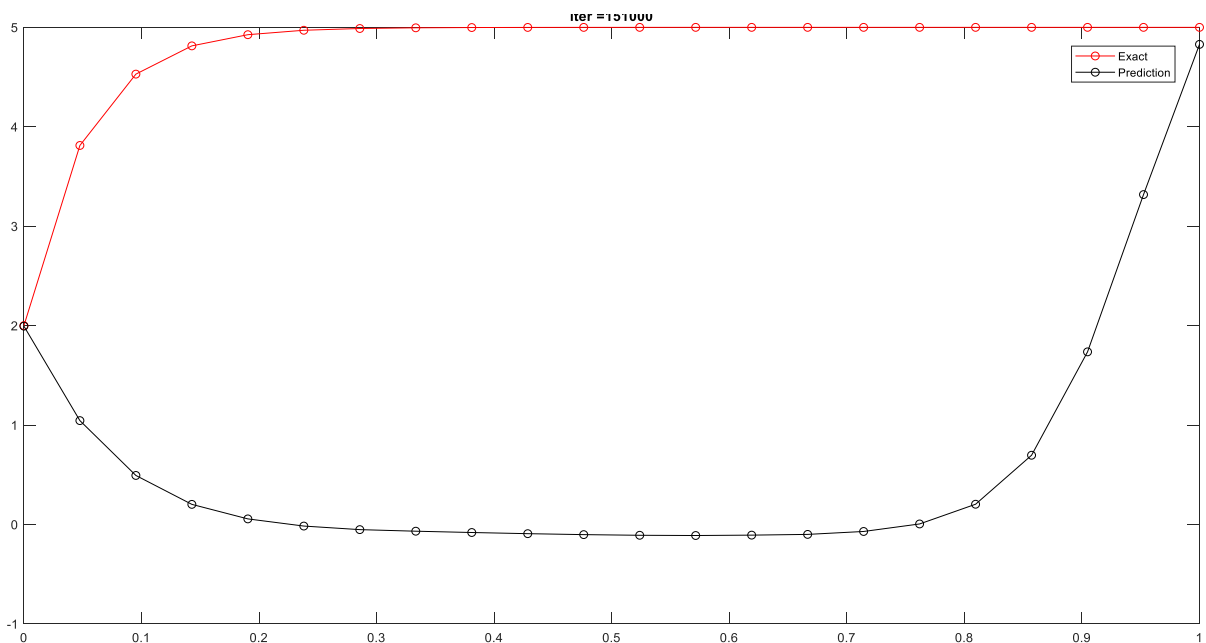
*Figure 5.11: Prediction for negative $\epsilon$ with flux collocation two points/block*

## 5.11. LOSS TERM TREND WITH ITERATION

### I. ROUGHLY APPROXIMATING CASE

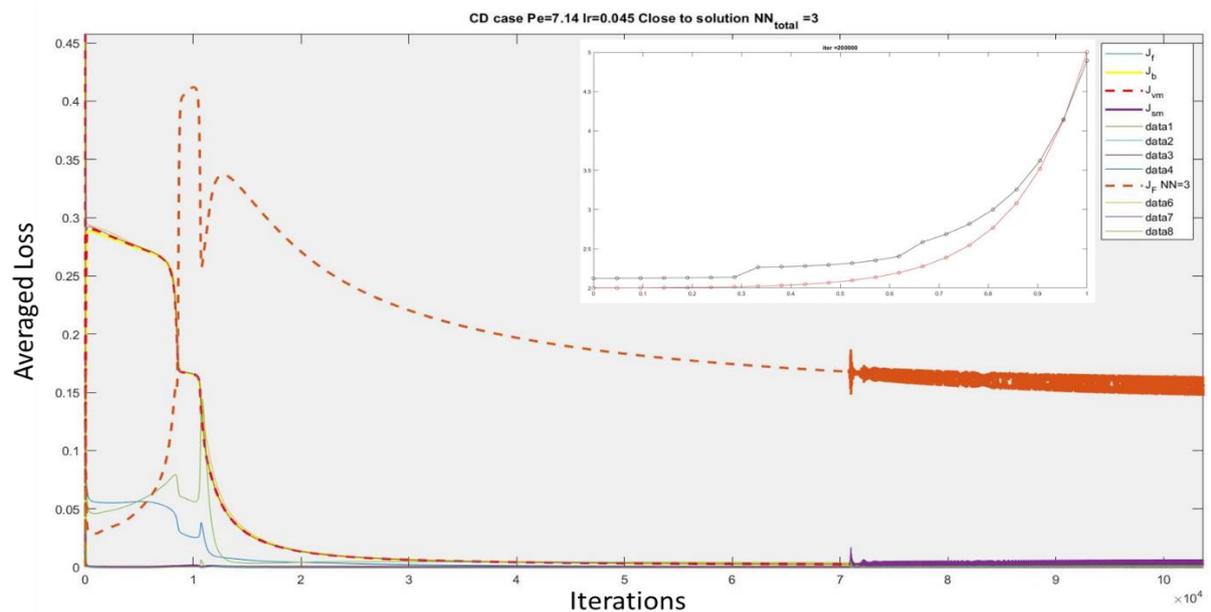

*Figure 5.12: Loss trend when approximation is has minor issues at boundaries and interfaces*

All the loss terms reduce to near zero values while loss due to discontinuity could not converge. It can be noticed that it takes a lot of iterations in order of 10,000-100,000s.

### II. FAILING APPROXIMATION CASE

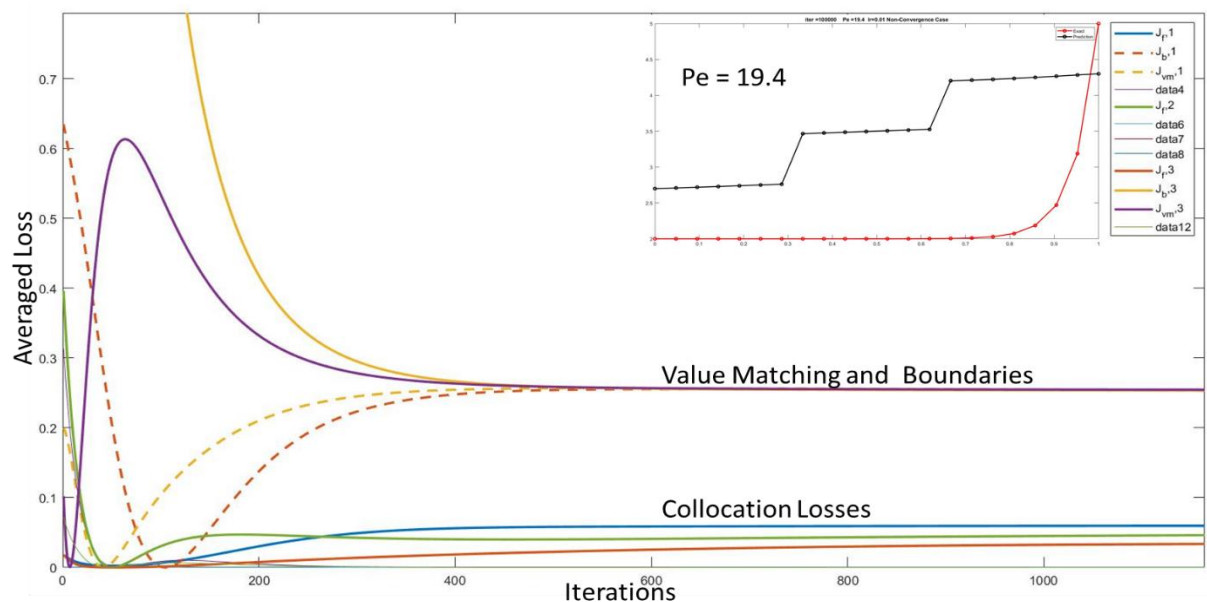

*Figure 5.13: Loss trend for positive $\epsilon$ when boundaries and interfaces fail completely.*

Interestingly all the losses stabilise under 1000 iterations but at high values. Collocation losses are way lesser than continuity losses and boundary losses. Somehow discontinuity losses and boundary losses converge to same value.

## III. NEGATIVE 𝜖 CONVERGING CASE

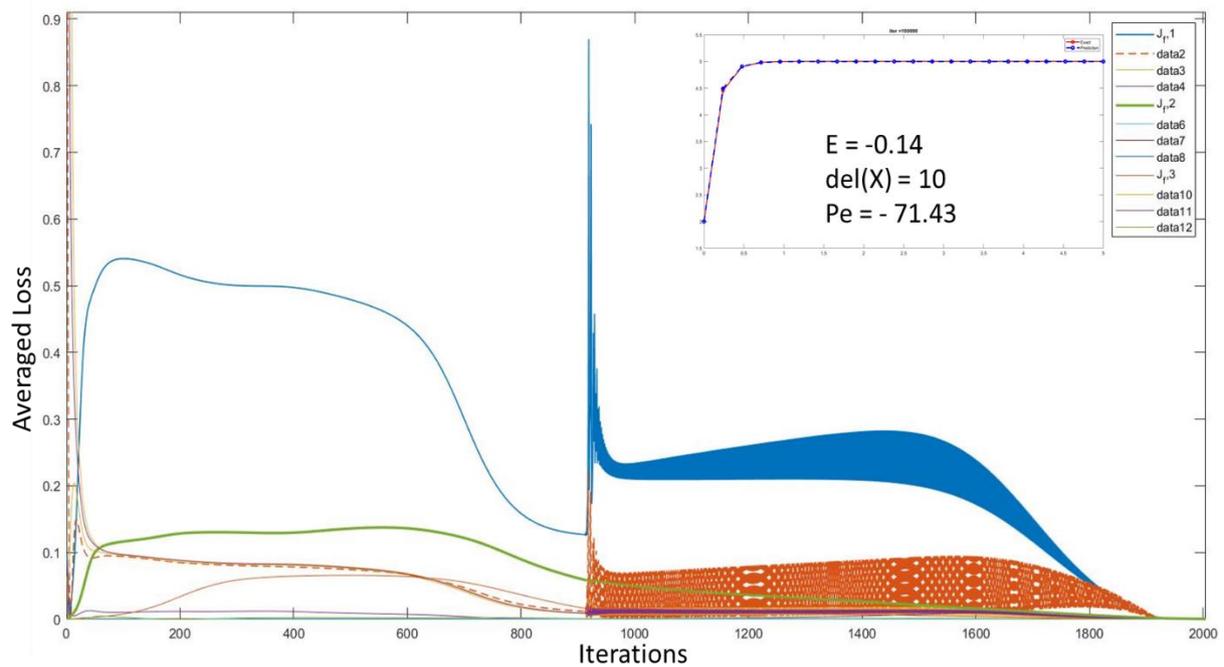

*Figure 5.14: Loss trend for negative 𝜖 when convergence is very exact*

The optimization converges very fast under 20,000 to 30,000 iteration with gradient descent optimizer. Strong oscillations could be seen which gradually stabilises to a low value.

## IV. NEGATIVE 𝜖 NON-CONVERGING CASE

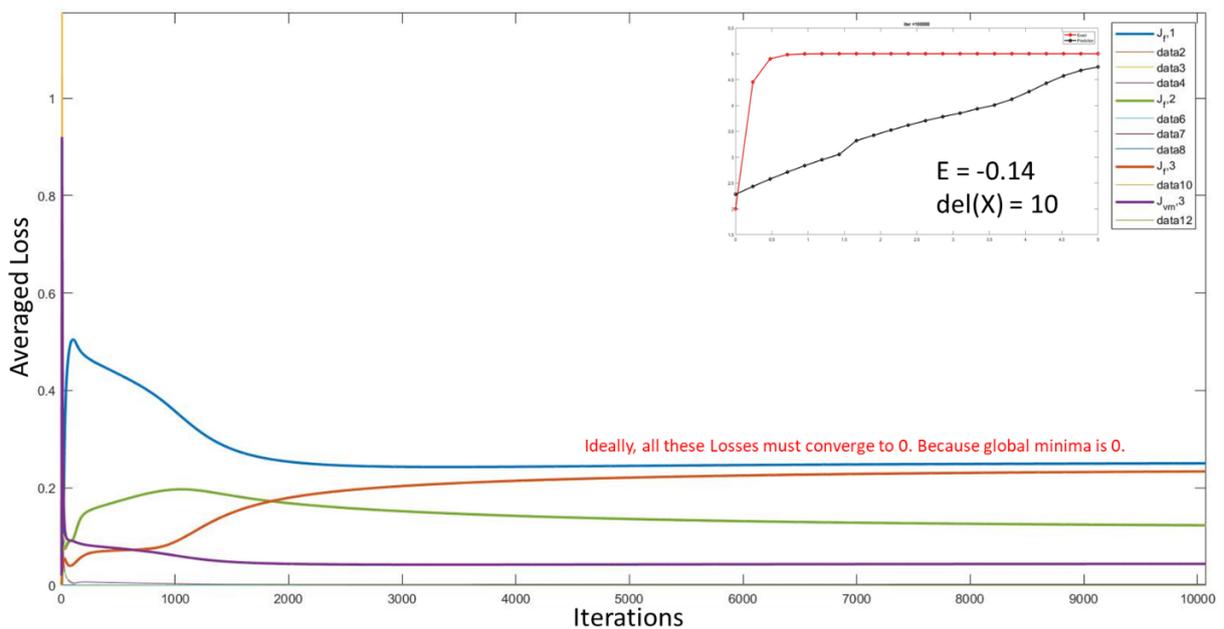

*Figure 5.15: Loss trend for negative 𝜖 when convergence fails*

The loss terms saturate within 10,000-20,000 iterations. This may indicate solution being stuck at local minima.

## 5.12. LOSS GRADIENT TREND WITH ITERATION

### I. FAILING APPROXIMATION CASE

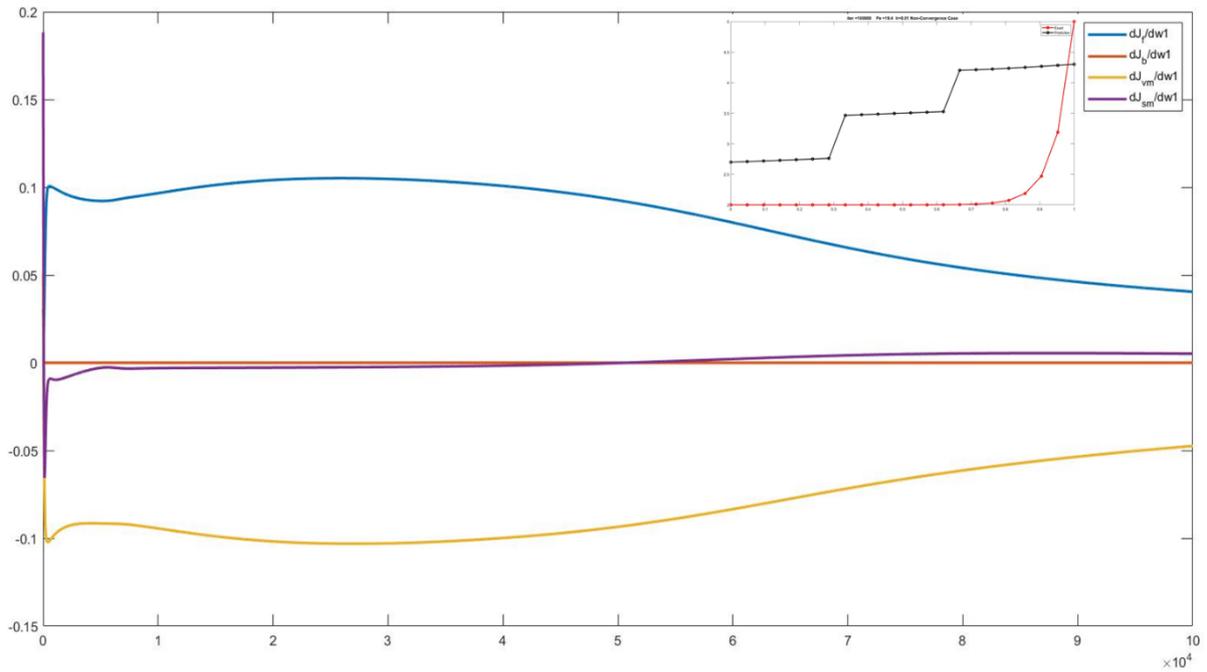

*Figure 5.16: Gradient trend for positive ϵ when convergence fails*

### II. NEGATIVE ϵ CONVERGING CASE

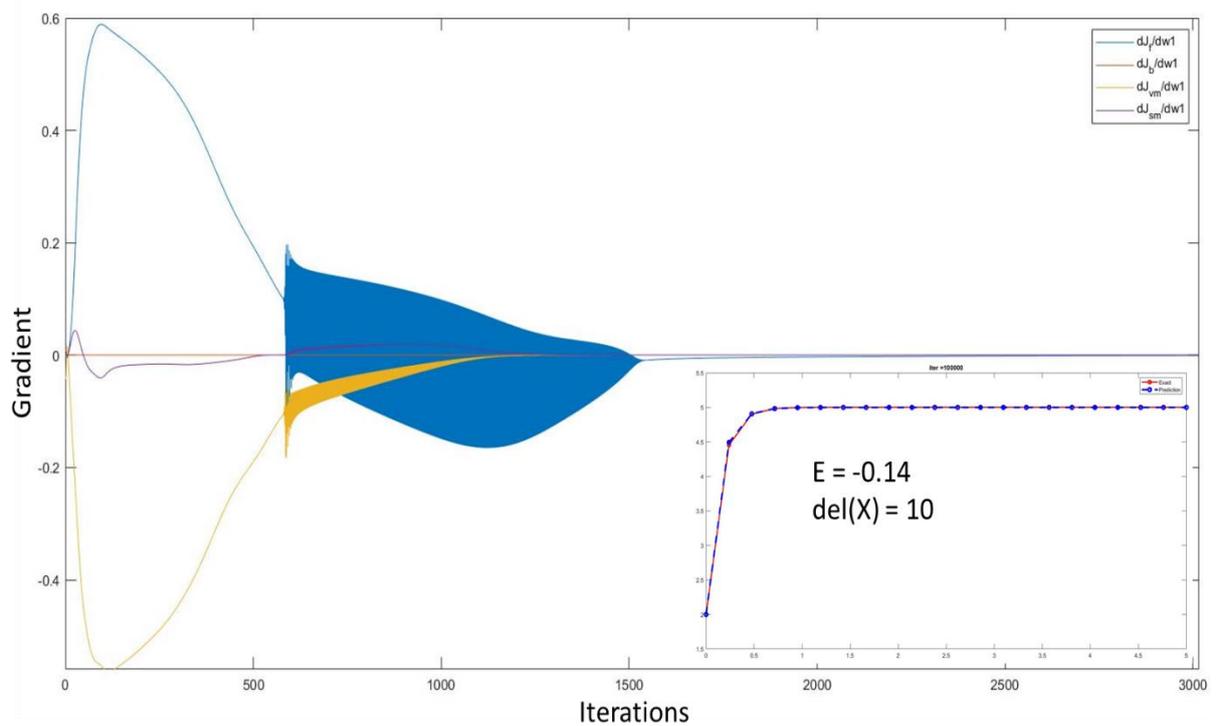

*Figure 5.17: Gradient trend for negative ϵ when convergence is very exact*

## III. NEGATIVE $\epsilon$ NON-CONVERGING CASE

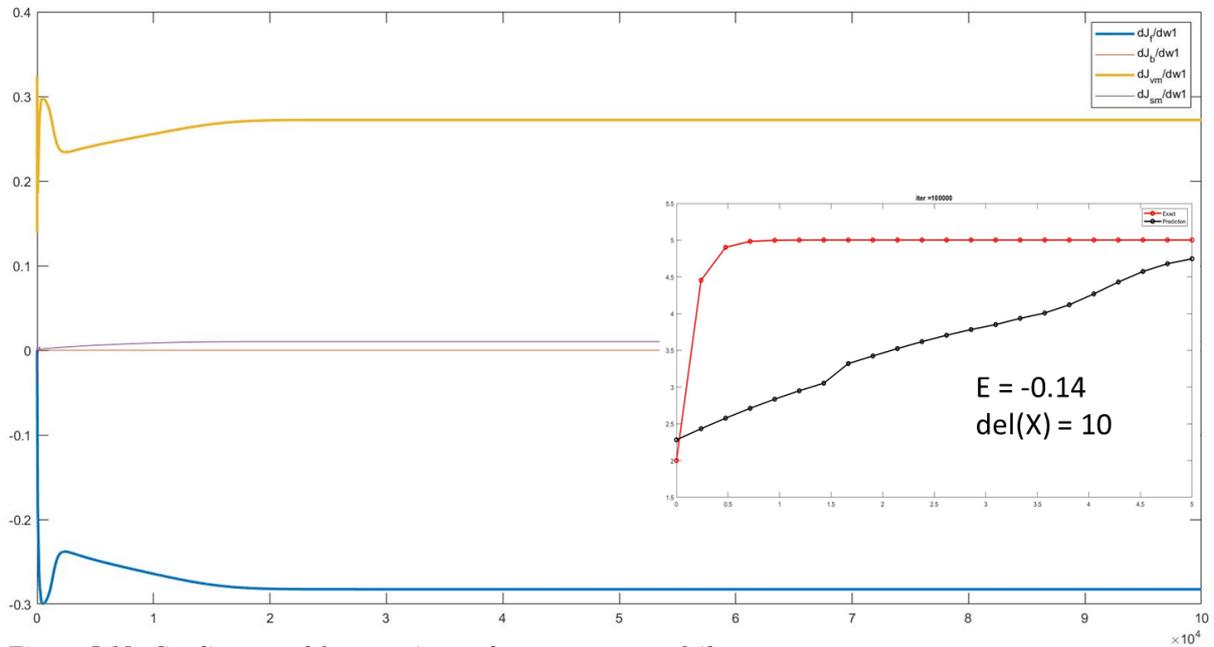

*Figure 5.18: Gradient trend for negative $\epsilon$ when convergence fails*

Somehow the gradients due to collocation losses and gradients due to continuity loss seem to cancel each other in all the cases. The trend of these gradients looks very much symmetric to naked eye. This is a strange occurrence. This leads to saturation of loss terms and vanishing net gradient during each iteration.

## 5.13. NORMALIZATION OF EACH SUB-DOMAIN IN DPINN

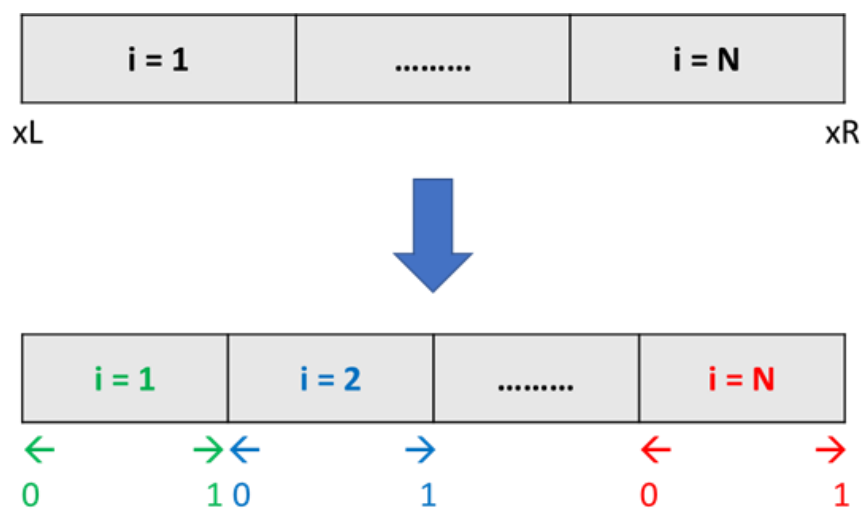

*Figure 5.19: Sub-domains scaled from 0 to 1*

The domain of each block of the DPINN model is scaled from 0 to 1. Inside a block, the domain was $[x_L + (i-1)\Delta x, x_L + i\Delta x]$, while the local variable ($\xi_i$) ranges from 0 to 1.

Within the sub-domain:

$$x = x_L + (i-1 + \xi_i)\Delta x$$

$$\Rightarrow dx = \Delta x . d\xi_i \quad \& \quad (dx)^2 = (\Delta x)^2 . (d\xi_i)^2$$

So, the differential equation becomes:

$$\varepsilon \frac{\partial^2 \phi}{\partial x^2} - \frac{\partial \phi}{\partial x} = 0$$

$$\Rightarrow \varepsilon \frac{\partial^2 \phi}{(\Delta x)^2 \partial \xi_i^2} - \frac{\partial \phi}{(\Delta x) \partial \xi_i} = 0$$

$$\Rightarrow \varepsilon \frac{\partial^2 \phi}{(\Delta x)\partial \xi_i^2} - \frac{\partial \phi}{\partial \xi_i} = 0$$

$$\Rightarrow \left(\frac{N_B . \varepsilon}{x_R - x_L}\right) \frac{\partial^2 \phi}{\partial \xi_i^2} - \frac{\partial \phi}{\partial \xi_i} = 0$$

Normalisation leverages the capability of DPINN method. It helps in increasing the range of '$\varepsilon$' for advection diffusion problems over which solution could be found. If $\varepsilon$ is in the range of 0.5 to 1 then prediction is very much perfect. So, if $\frac{N_B . \varepsilon}{x_R - x_L}$ is in the range of 0.5 to 1 then prediction for extremely low $\varepsilon$, like 0.005 or 0.002, is possible.

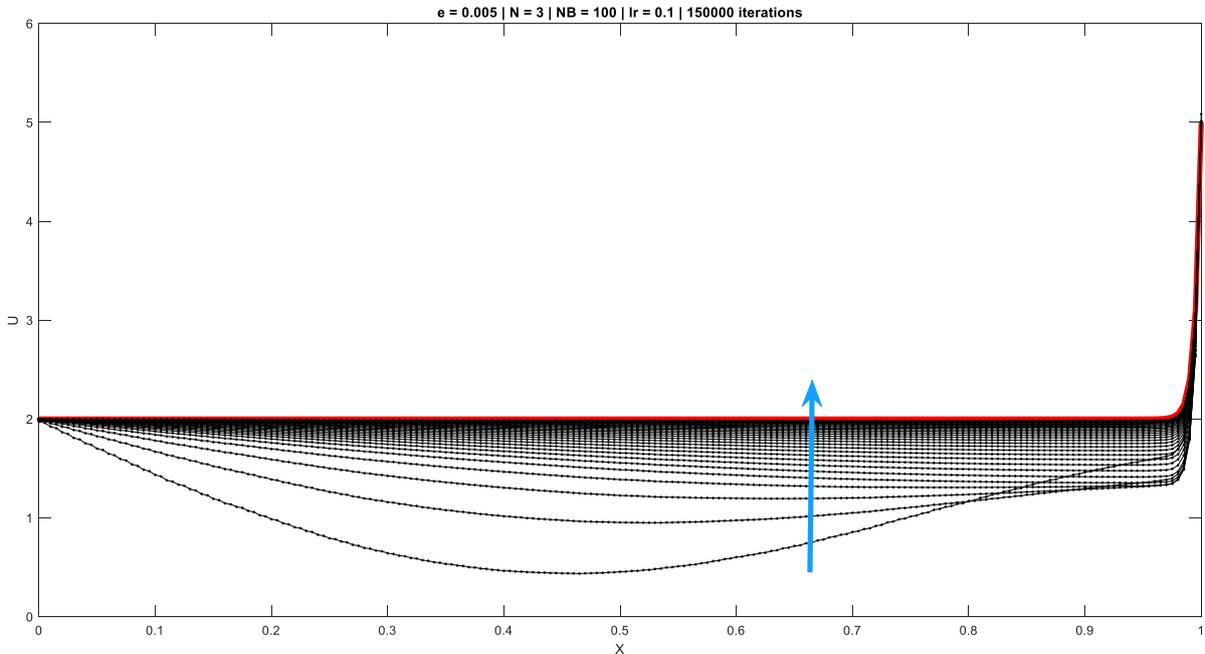

*Figure 5.20: Development of the solution with normalised DPINN*

The solution stabilises very beautifully. The figure shows the curve falling into the right place, the arrow shows the direction of developing curve in every 20000 iterations. Only problem that still persists is that runtime does not reduce.

## 5.14. TESTS FOR FINDING ISSUES

Issues may have rose due to many problems but most probably due to either poor optimisation or ill-posed algorithm for physics information (training). Various tests were conducted to figure out if there is any error due to training.

- **Linear Approximation** by Solving Exactly Required Consistent Discrete Equations
- **Linear Approximation** by Norm Minimisation of Discrete Equations
- **Quadratic Approximation** by Norm Minimisation of Discrete Equations
- Optimization of **Neural Network using Levenberg-Marquardt** Algorithm

### 5.14.1. PIECEWISE LINEAR APPROXIMATION

Let us have N panels and approximating function in each panel be $Y = A_i x + B_i$, where $A_i$ and $B_i$ are to be determined to find the fit, where i = panel number.

Governing equations: $\in \frac{\partial 2Y}{\partial x2} - \frac{\partial Y}{\partial x} = 0 \Rightarrow 0 - A_i = 0$ ( **invalid** )

So, Flux equation is used as governing equation: $\in \frac{\partial Y}{\partial x} - Y = 0$

$$\Rightarrow \in A_i - A_i x - B_i = 0$$

Interface continuity equation,

$$Y(x=x_{i,R}) = Y(x=x_{i+1,L})$$

$$\Rightarrow A_i \cdot x_{i,R} + B_i = A_{i+1} \cdot x_{i+1,L} + B_{i+1}$$

Boundary equation,

$$Y(x=x_L) = Y_L \quad \& \quad Y(x=x_R) = Y_R$$

$$\Rightarrow A_1 \cdot x_L + B_1 = Y_L \quad \& \quad A_N \cdot x_R + B_N = Y_R$$

The number of unknowns that needs to be determined is 2N. For exact solution, 2N equations are required to be solved. There are two compulsory equations for boundary conditions, N-1 compulsory equations for value match at interfaces for continuity. N – 1 more equations are required for exact solution. So, N-1 distinct collocation points are used over governing differential equation but there are infinite points for collocation so, equations will be inconsistent.

Writing the equations in Matrix form, MX = C where X = coefficient matrix,

$$X = M^{-1}.C$$

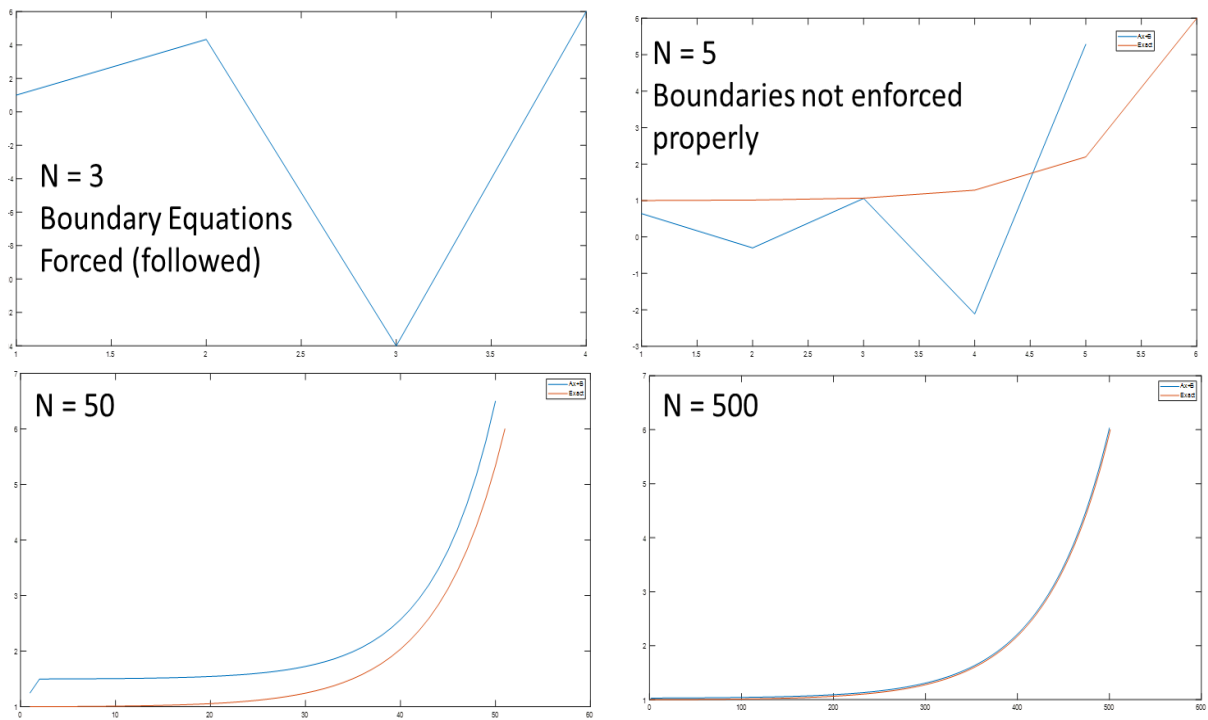

*Figure 5.21: Effect of number of collocation points for exact solution using linear approximation*

In case of inconsistent equations, M is not invertible. So, pseudoinverse of M is taken instead. Penrose-Moore pseudoinverse solution give the solution in least-norm sense.

$$X = pinv(M).C$$

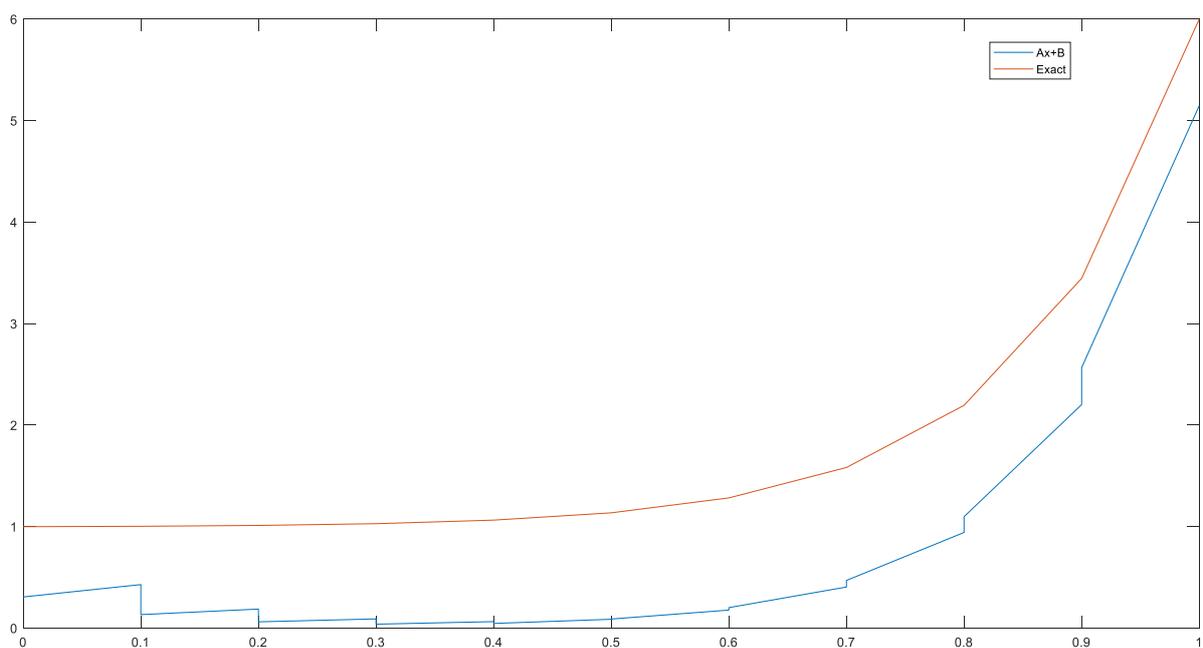

*Figure 5.22: Solution using of pseudoinverse of M with 10 blocks*

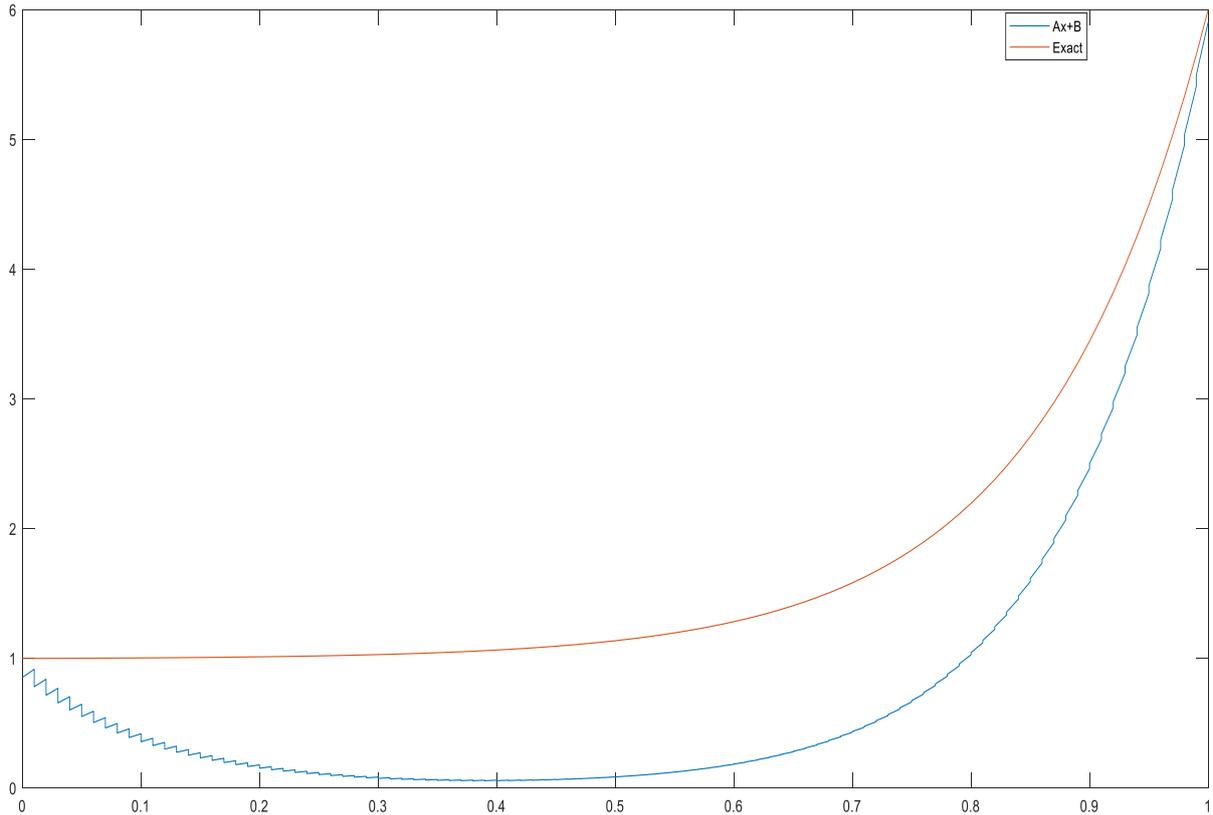

*Figure 5.23: Solution using of pseudoinverse of M with 100 blocks*

Equations solved by pseudoinverse, essentially, computes the solution for best fit by least squared norm minimization. MATLAB's pinv() is used for the purpose.

Discontinuities are clearly evident. Even boundaries are not met. Problems are similar to that faced in DPINN.

### 5.14.2. PIECEWISE QUADRATIC APPROXIMATION

Let us have N panels and approximating function in each panel be **Y = $A_i x2$ + $B_i$x + $C_i$**, where $A_i$ and $B_i$ are to be determined to find the fit, where i = panel number.

Governing equations: $\epsilon \frac{\partial 2Y}{\partial x2} - \frac{\partial Y}{\partial x} = 0$   => $2\epsilon A_i - 2A_i x - B_i = 0$

If flux is used as governing equation: $\epsilon \frac{\partial Y}{\partial x} - Y = 0$  => $2\epsilon A_i x + \epsilon B_i - 2A_i x - B_i = 0$

The other equations remain the same as used in linear approximation.

 Somehow the prediction by forcing the governing equation fails. The figure below shows the solution by taking pseudoinverse.

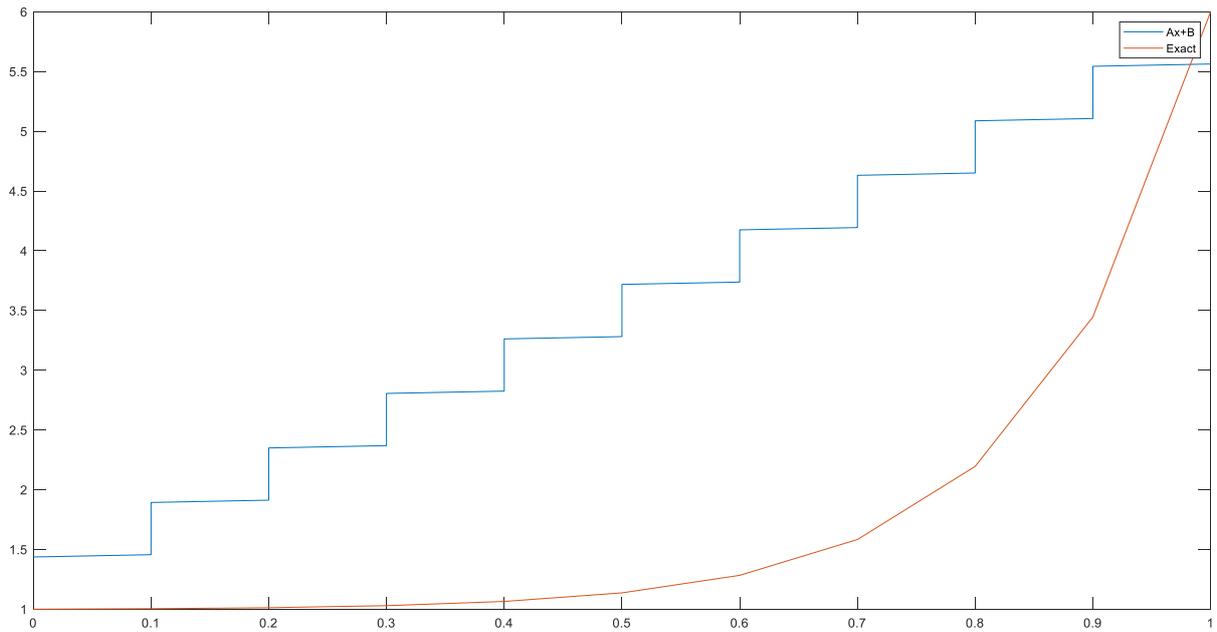
*Figure 5.24: Solution by piecewise quadratic approximation with direct differential equation*

But by forcing the flux equation better solution is obtained. The figure below gives the solution.

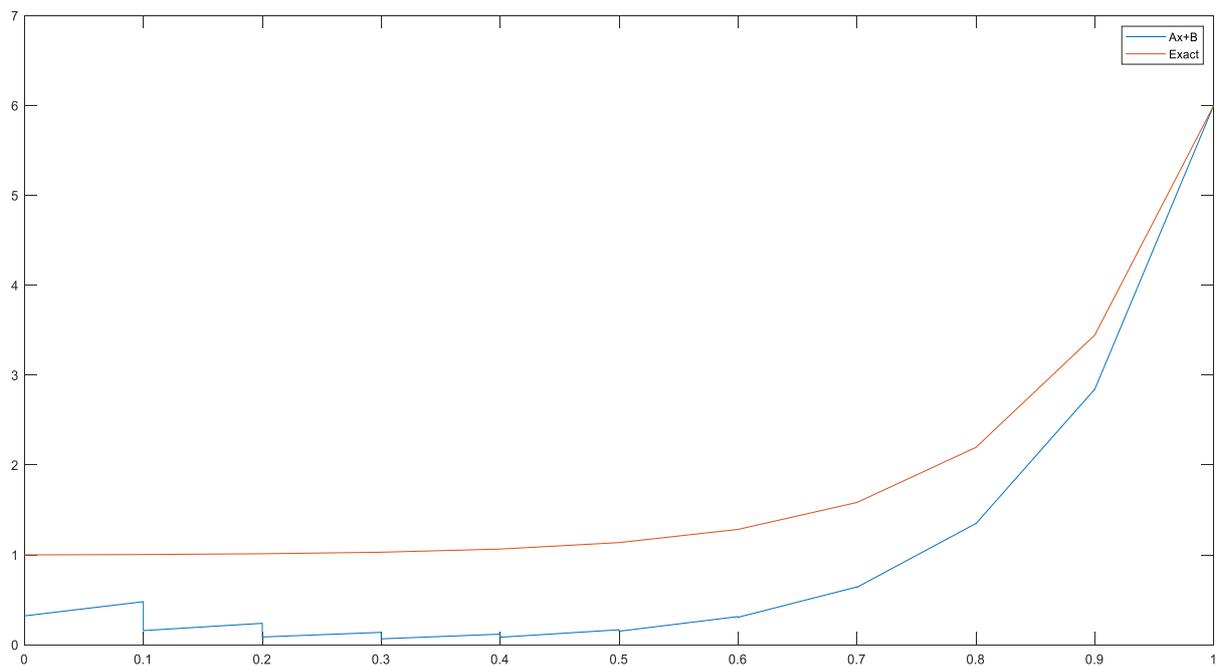
*Figure 5.25: Solution by piecewise quadratic approximation with flux equation*

### 5.14.3. CONCLUSION FROM EQUATION SOLVING

The cases showed similar problems as those evident while prediction using neural network. It is very much evident that the problem is with optimisation as well as physics posing. With least squared (norm minimised) solution of the discrete equations, discontinuities are seen at boundaries and interfaces.

## 5.15. LEVENBERG-MARQUARDT ALGORITHM FOR WEIGHT UPDATE

Levenberg-Marquardt Algorithm (LMA) is a much more robust algorithm for optimization. It can work faster than gradient descent optimizer in some cases. But the associated problems are large memory involved in the process and works well only if loss function is in terms of least squares and network has single layer architecture.

The weight update equation is:

$$w_{k+1} = w_k - \left(J_k^T J_k + \mu I\right)^{-1} J_k e_k$$

For single layer networks, LMA seemed to work faster than gradient descent in case of advection-diffusion problems. Diffusion problem works very fast and very accurately with our shallow network. However, LMA does not seem to be giving better solution than gradient descent for advection-diffusion problem. The prevalent problems still persist.

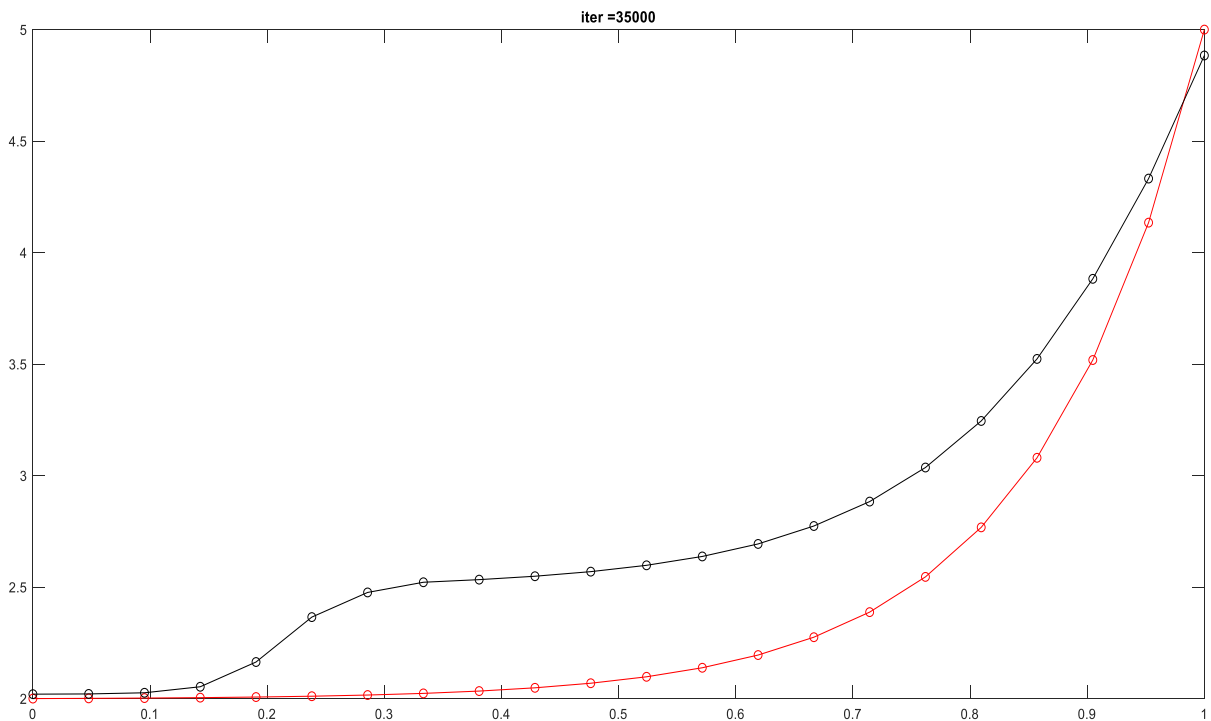

*Figure 5.26: Solution by LMA for $\epsilon = 0.14$*

The prediction seems to deviate from the solution as the $\epsilon < 0.15$. Very much like we had observed earlier. Steps starts to appear as even lower values are taken.

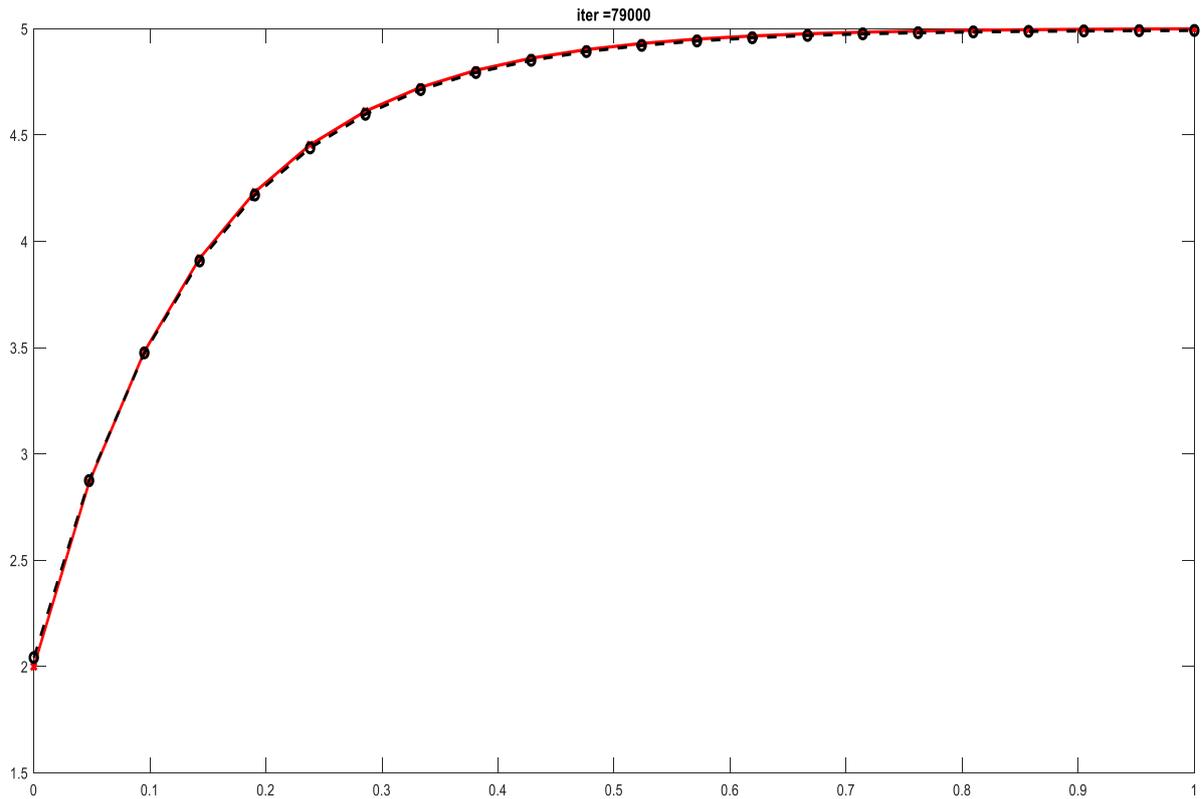

*Figure 5.27: Solution by LMA  for $\epsilon$ = -0.14*

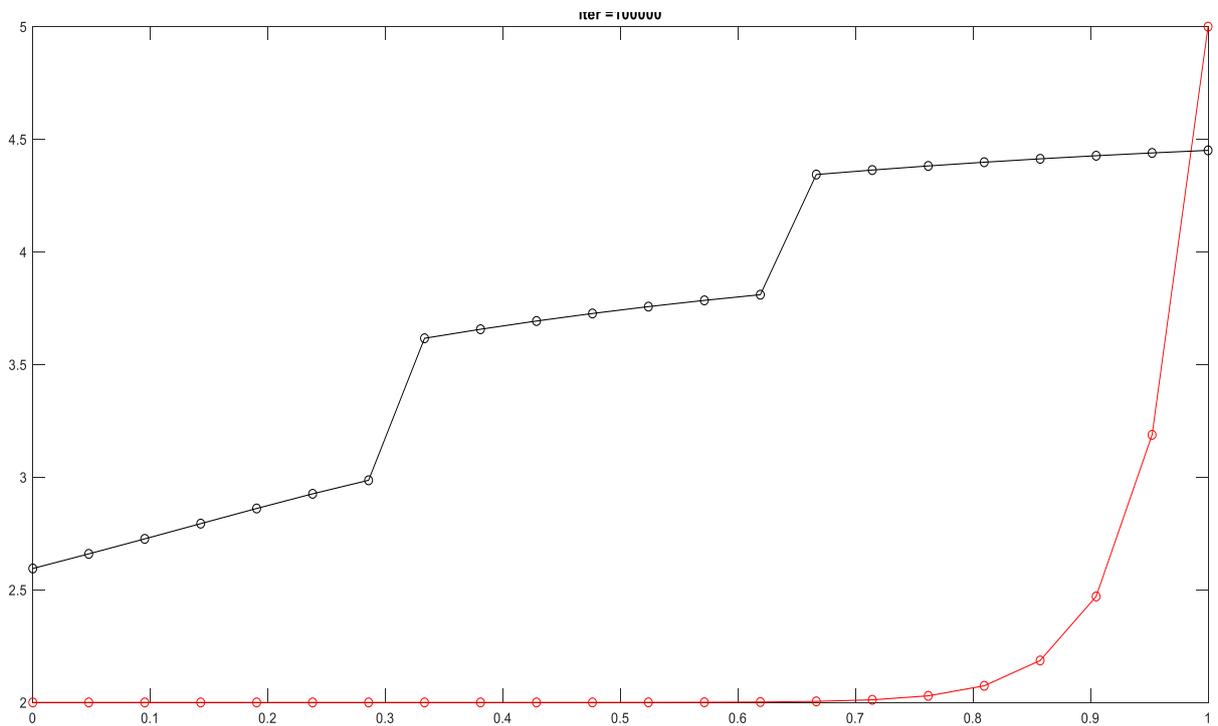

*Figure 5.28: Solution by LMA  for $\epsilon$ = 0.1*

It can be observed that $J^TJ$ is singular for most of the cases with advection dominance. Similar problem persists in LMA as well as in case of gradient descent optimizer. Convergence is better for negative Peclet number cases because $J^TJ$ does not become singular.

# CHAPTER 6
# EXTREME LEARNING MACHINE BASED DPINN

## 6.1. EXTREME LEARNING MACHINE (ELM)

Extreme learning machines are special feedforward neural networks with a single layer or multiple layers of hidden nodes, where the parameters of hidden modes are not tuned. These hidden nodes can be randomly assigned or inherited from another model and never updated. The weights of final layer are learned to produce results as linear combination of non-linear entities. The term extreme has been coined because of its fast training speed.

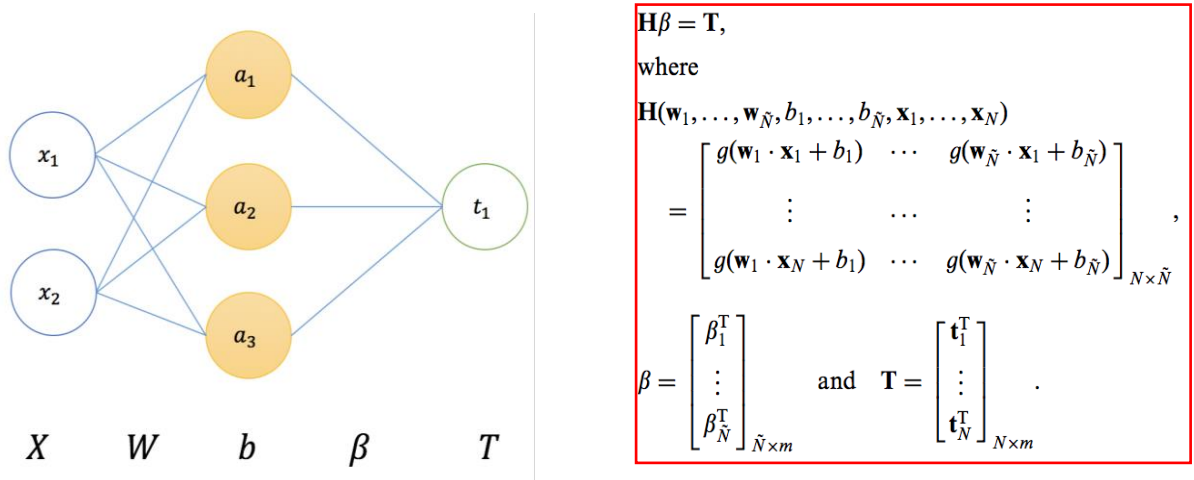

*Figure 6.1: Architecture and algorithm for ELM*

The network is trained by finding least square solution **β'** to the linear system **Hβ = T**. Output weights can be determined exactly or by pseudo-inverse or by least squared regression. Prediction is a linear combination of randomly set non-linear entities. Sometimes, fixed random weights in the first layer makes the prediction uncertain when number of neurons is insufficient. So, taking large number of neurons helps.

The comparison between ELM and normal method can be referred in the table below.

| Extreme Learning Machine (ELM) | Gradient based Algorithms e.g. Backpropagation |
|---|---|
| Unique minimum solution | Prone to local minima convergence trap |
| Gives smallest training error & weight norm | Minimizes Error alone |
| Does not need a stopping method | Overtraining occurs if improper stopping methods and validation |

*Table 6.1: Comparison between ELM and typical neural network*

## 6.2.1. ELM IN PINN FOR ADVECTION DIFFUSION

The architecture for ELM is shown in the figure below.

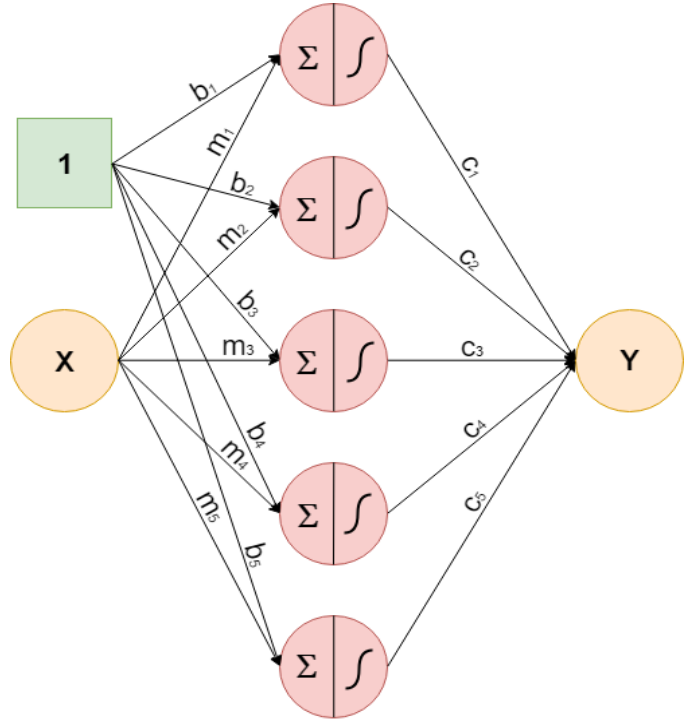

*Figure 6.2: Architecture for ELM-PINN*

The formulation of neural network in single layered ELM is:

$$NN(x) = \Sigma c_i \tanh(m_i x + b_i)$$

Equations at collocation points:

$$NN'(x) - \varepsilon NN''(x) = 0$$

$$\Rightarrow \Sigma c_i (m_i - \varepsilon m_i^2)\tanh(m_i x + b_i) = 0$$

Equations for boundary conditions:

$$NN(x_b) = \Sigma c_i \tanh(m_i x_b + b_i) = Y_b$$

It can be noticed that all the equations are linear in $c_i$. When represented as a system of linear equations $[A][C] = [R]$. So, finding $[C]$ matrix becomes easy and fast.

$$[C] = [A]\backslash[R]$$

If the number of collocation equations is N, and the number of boundary equations is 2 then for exact solution, we need N+2 unknowns for exactness. So, N+2 neurons are taken in the architecture.

ELM preserves the non-linearity but since only the second layer of weights are variable the optimization objective becomes simpler. The set of equations are linear combinations of the weights; hence the varying weights can be solved exactly if the number of neurons equals the number of equations. There is no need for backpropagation. If a greater number of training equations/points are taken than the number of neurons then least squared solution is a very good approximate. This is the reason why this algorithm is way faster.

**6.2.2. OBSERVATIONS**

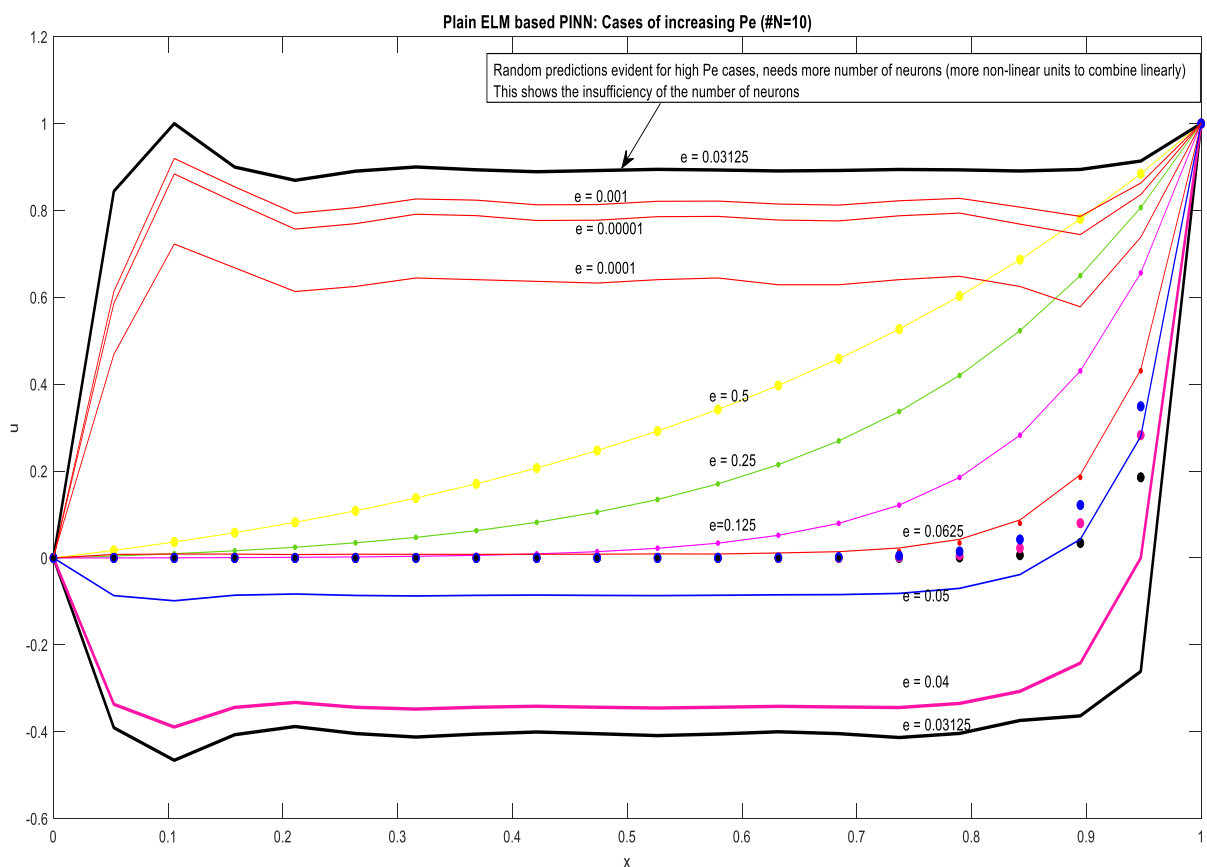

*Figure 6.3: ELM-PINN predictions for different ε*

Optimization was a difficulty in normal PINN. ELM makes the optimization objective simpler since it becomes linear function of variables while the ability to capture complex functions is still preserved to a great extent as a result ELM based PINN performs better and way faster. The figure above shows the predictions with 10 collocation points for various 'ε' values over exact criterion. With these conditions, prediction is very accurate for ε > 0.06, otherwise random curves appear.

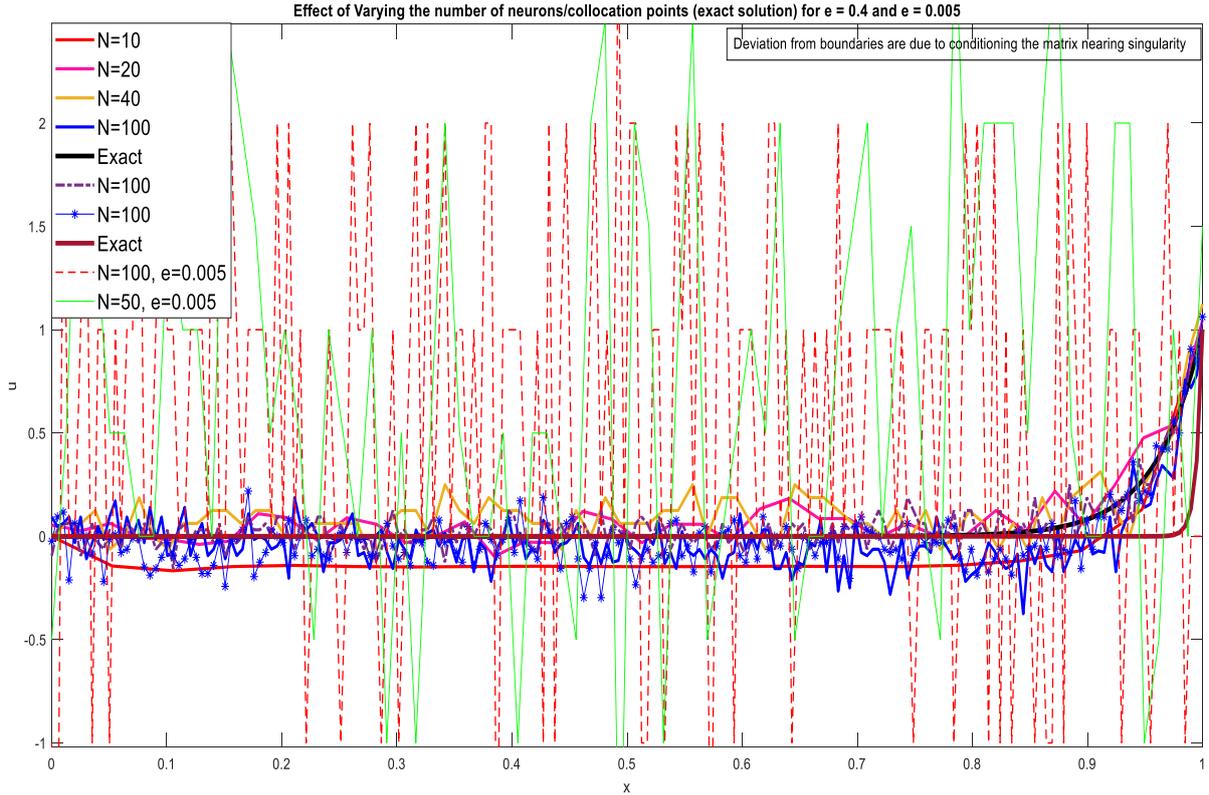

*Figure 6.4: ELM-PINN predictions for different number of collocation points*

The previous figure shows the effect of increase in number of collocation points with exact criterion for ε = 0.4. With increasing number of collocation points oscillations seems to start appearing around the accurate solution. If ε = 0.005 the oscillations seem to be very large and unstable. It could also be noticed that the boundaries are always exactly fit while minor deviation due to matrix conditioning while inversion in case of approaching singularity. With increasing number of collocation points the matrix inversion becomes difficult since it approaches singularity similar to the observations noted in solving the equations exactly. Since, conditioning is being done for very low ε, the boundaries (and interfaces in DPINN) could not be matched. Different objectives need different level of conditioning to help in optimizing this multi-objective problem. These objectives include the boundaries, interfaces equations and equation forcing. The oscillations are very much random and depend on the fixed weights in the fixed layer.

The use of pseudoinverse in these cases help since they provide the least-squared norm solution of the equations. They stabilise the oscillations. The figure below shows the effect of pseudoinverse on these problems. If the number of collocation points is very high then, since, the solution is found out using squared norm minimisation the effect of boundary constraints (only 2 equations) are washed out gradually with increase in the number of collocation points. Similar issue occurs at the interface points as well.

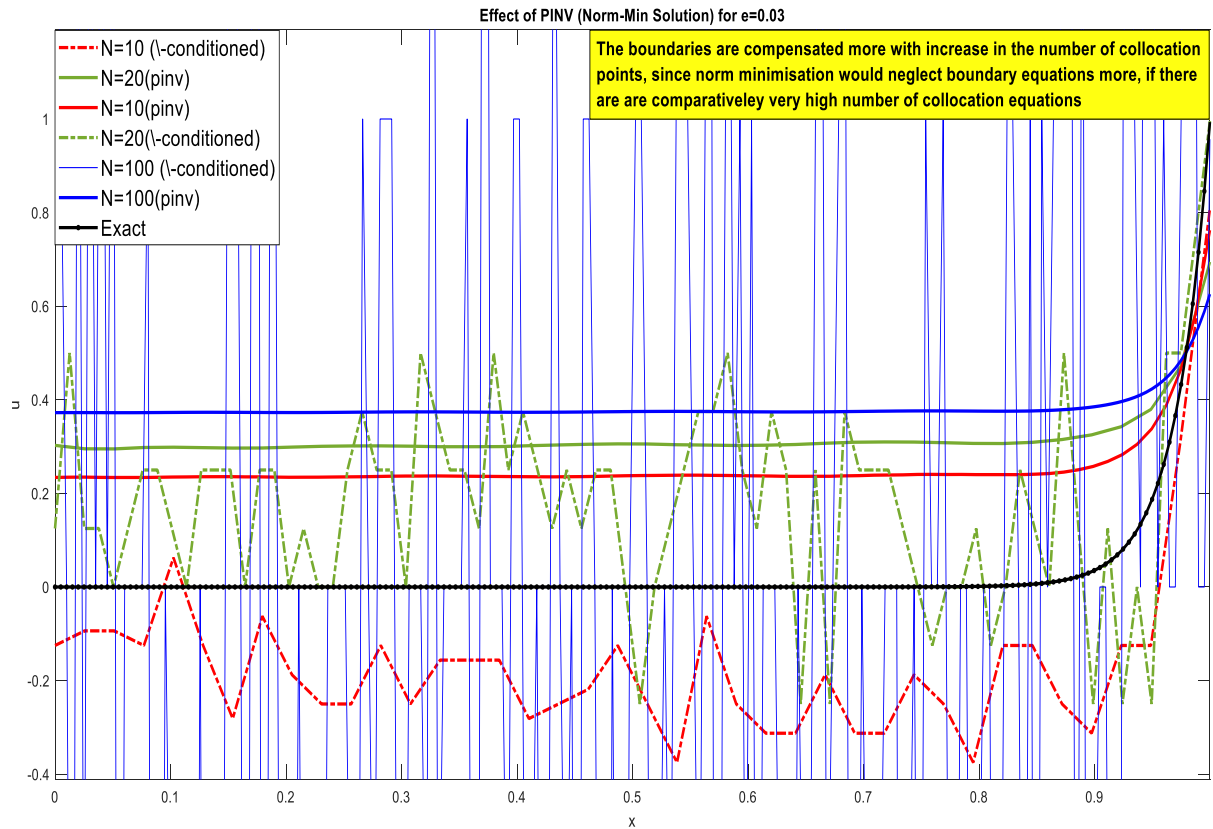

*Figure 6.5: ELM-PINN predictions with and without pseudoinverse*

### 6.3.1. ELM IN DISTRIBUTED PINN

The architecture remains the same but numerous networks are used, one over each panel.

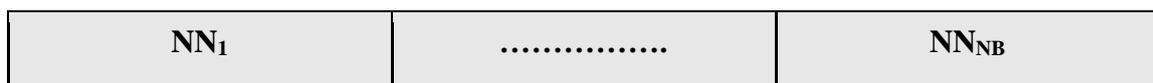

*Figure 6.6: Split of domain for ELM-DPINN*

The formulation of the network remains the same.

$$NN_i(x) = \Sigma c_i \tanh(m_i x + b_i)$$

Equations at collocation points:

$$NN'(x) - \varepsilon NN''(x) = 0$$

$$\Sigma c_i (m_i - \varepsilon m_i^2) \tanh(m_i x + b_i) = 0$$

Equations for boundary conditions:

$$NN(x_b) = \Sigma c_i \tanh(m_i x_b + b_i) = Y_b$$

Continuity equations at interfaces:

$$NN_L(x_{int,L}) - NN_R(x_{int,R}) = 0$$

$$\Sigma c_{L,i} \tanh(m_{L,i} x_{int,L} + b_{L,i}) - \Sigma c_{R,i} \tanh(m_{R,i} x_{int,R} + b_{R,i}) = 0$$

Differentiability equations at interfaces:

$$NN'_L(x_{int,L}) - NN'_R(x_{int,R}) = 0$$

$$\Sigma c_{L,i} m_{L,i} \tanh(m_{L,i} x_{int,L} + b_{L,i}) - \Sigma c_{R,i} m_{R,i} \tanh(m_{R,i} x_{int,R} + b_{R,i}) = 0$$

The number of boundary equations will be **2**. Let the number of blocks be **NB** and the number of collocation equations per block be **N**. So, the number of continuity equations at interfaces will be **NB − 1** and the number of differentiability equation at interfaces = **NB − 1** For exact solution, we need **NB(N+2)** unknowns. So, **N+2 neurons** can be taken **in each network** for exact solution.

### 6.3.2. OBSERVATIONS

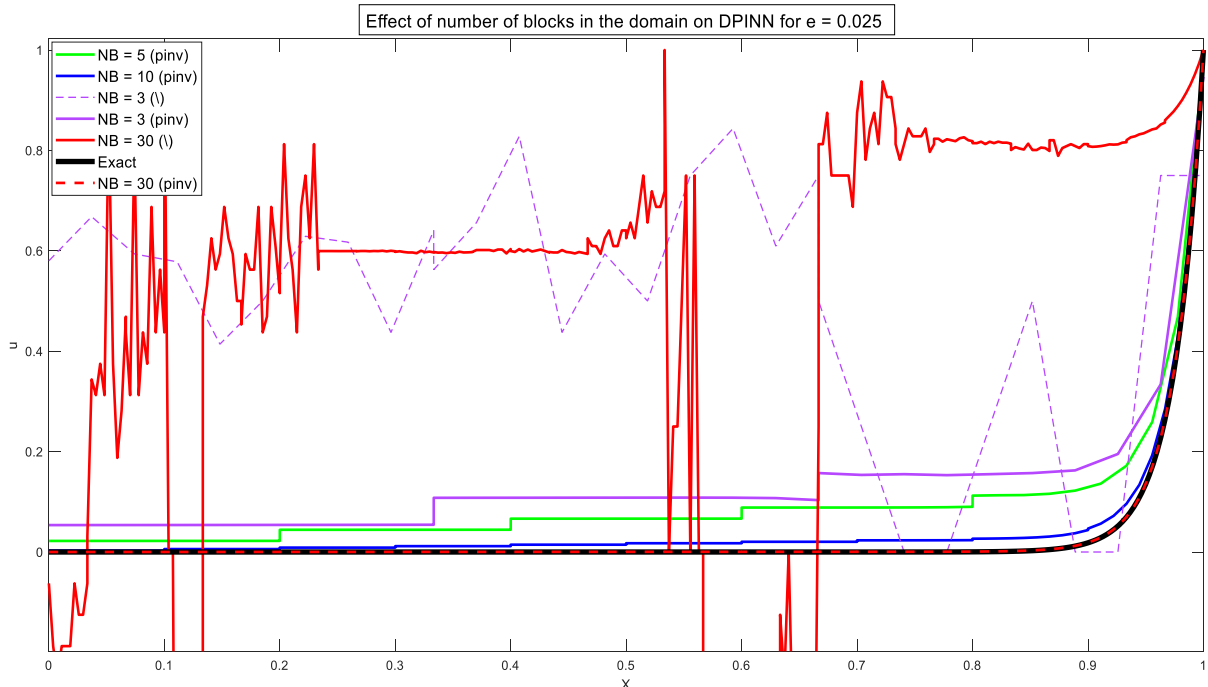

*Figure 6.7: Predictions of ELM-DPINN with different number of blocks*

The effect of number of blocks for a case where ε = 0.025 can be seen in the figure above. With increase in the number of blocks the prediction with pseudoinverse seem to match with the exact solution well. The pseudo inverse seems to stabilise the random predictions like in case ELM based PINN. The continuous red line is the conditioned inverse while dashed red is the pseudoinverse prediction for the same fixed weights. The discontinuities too reduce with increasing number of blocks.

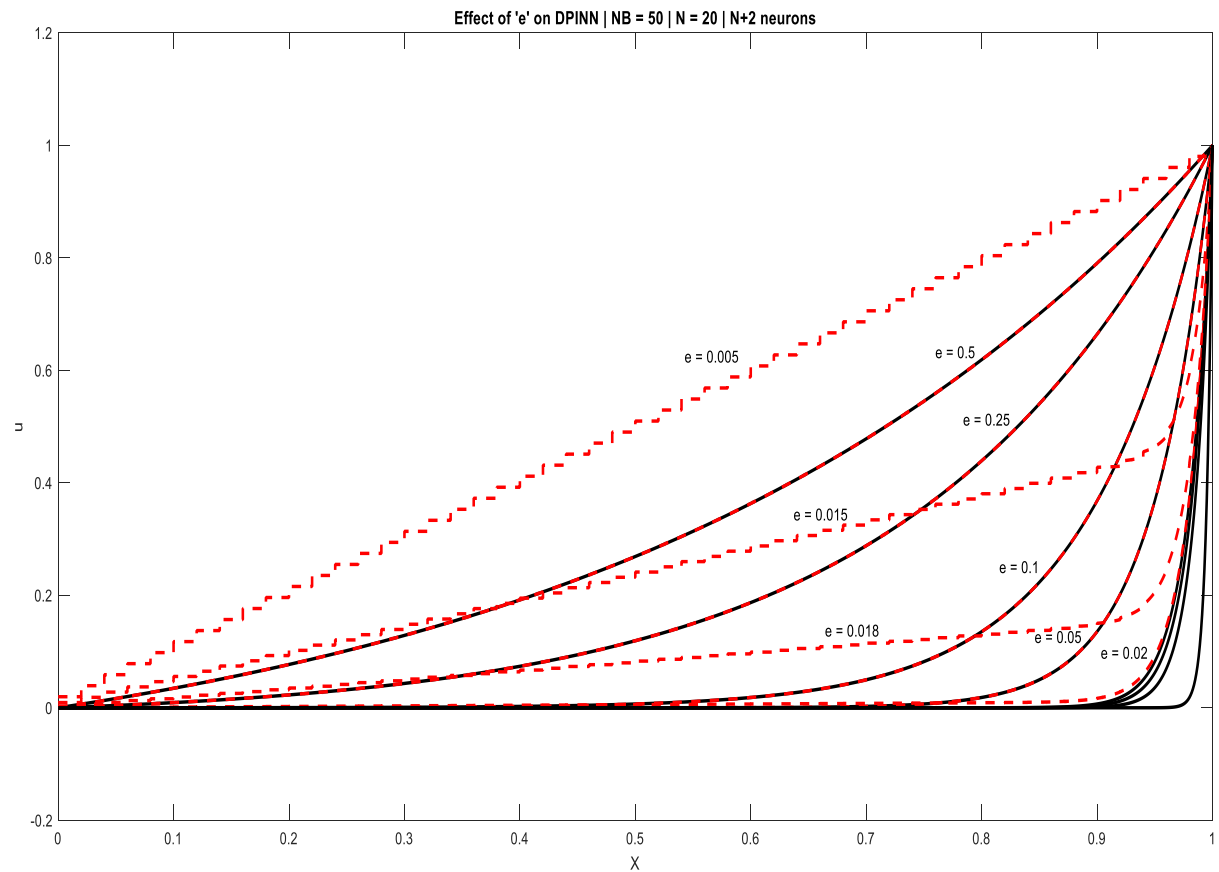

*Figure 6.8: Predictions of ELM-DPINN for different ε*

The effect of decreasing ε is captured in the figure above. With 50 blocks with N+2 neurons per network where N is the number of collocation points for exactness solution, very accurate solution could be obtained for ε > 0.02 (ELM-PINN could solve for ε > 0.06). It should be noted that since with conditioned matrix, there are large oscillations and they are dependent on random fixed weights as well pseudoinverse is used as it gives the norm-minimised solution. As the ε is decreased more the same deviations in continuity and constraints appear like an optimization problem, since the least norm solution is kind of least square optimised solution.

The figure below shows the effect of number of collocation points (N) per block with exact solution criteria. When N=1, it clearly showed the inability to solve the problem. At least 2 neurons are required to be able to approximate and with increase in the number of neurons the approximation improves till a number after which the solution become difficult probably because the weightage of boundaries and interface criterions are washed out for a given number of blocks and for a particular value of ε. Moreover, the collocation points have to be at the edges at least, else it fails epically.

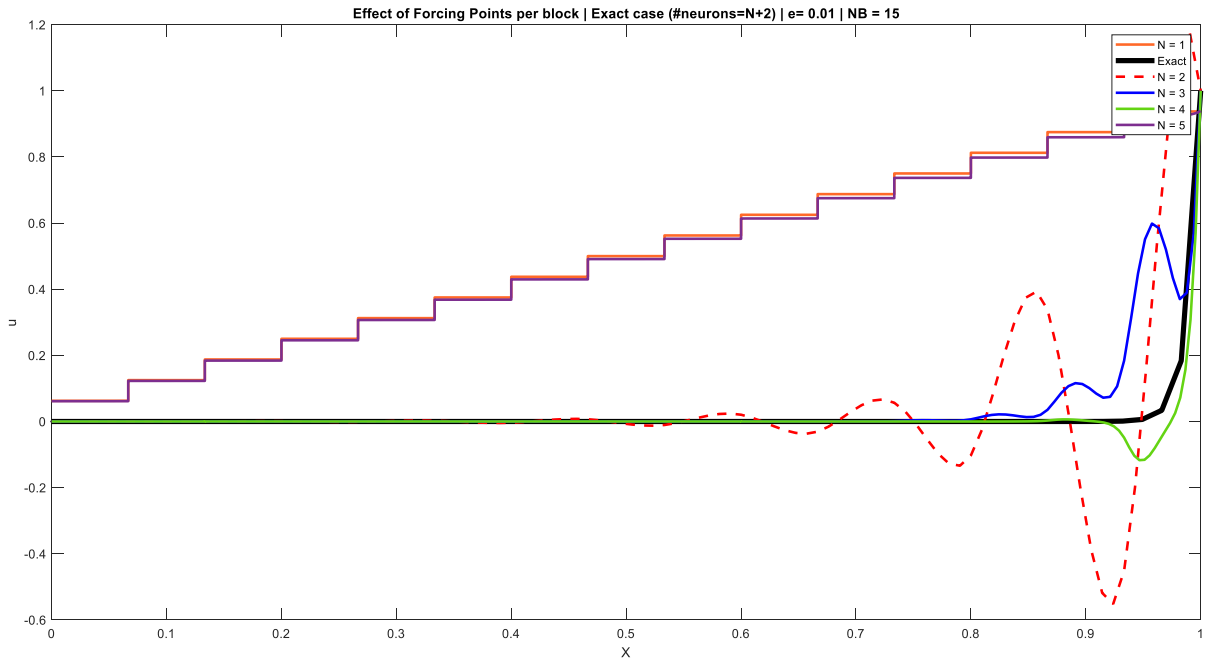

*Figure 6.9: Predictions of ELM-DPINN for different number of blocks*

The figure below shows the effect of number of neurons per network. If we have M data points and we need to fit a regression curve, it needs to have at least M unknown to tune them and pass through all of them. This case is valid here as well. It can be observed that for number of neurons less than $N + 2$, the network is unable to approximate to the solution.

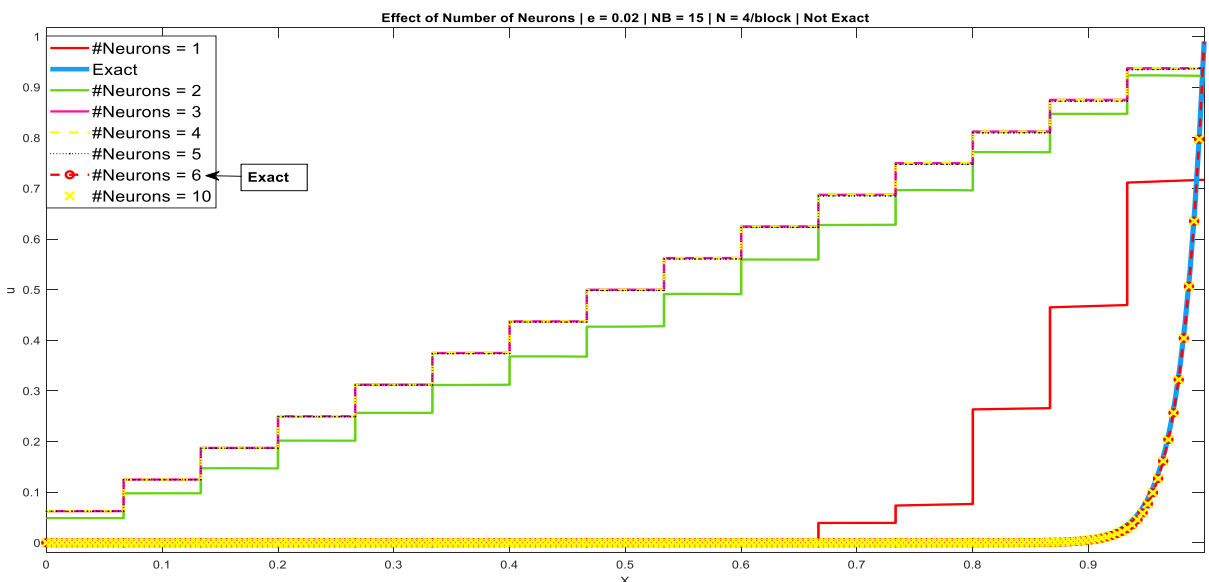

*Figure 6.10: Predictions of ELM-DPINN for different number of neurons*

## 6.4. NORMALIZATION OF ELM BASED DPINN

On scaling each block like the case with DPINN, ELM based DPINN also works much better. The transformed differential equation becomes,

$$\frac{N_B \cdot \varepsilon}{x_R - x_L} \frac{\partial^2 \phi}{\partial \xi_i^2} - \frac{\partial \phi}{\partial \xi_i} = 0$$

Where $N_B$ is the number of block and $\xi_i$ is the local variable in the i$^{th}$ block. The effect of number of number of blocks can be noticed in the figure below. For $\varepsilon = 0.005$, if $N_B$ is is 3 or 5 then steps are seen which indicates the optimization failure. With increasing the number of blocks $N_B$ optimization clearly happens smoothly. The solution from central differencing (CDS) and normalised ELM-DPINN have same colour while CDS model has dashed type lines. For same number of collocation points, CDS clearly approximates more accurately but with high $N_B$ values the CDS and normalised ELM-DPINN have very close solution, nearly equal.

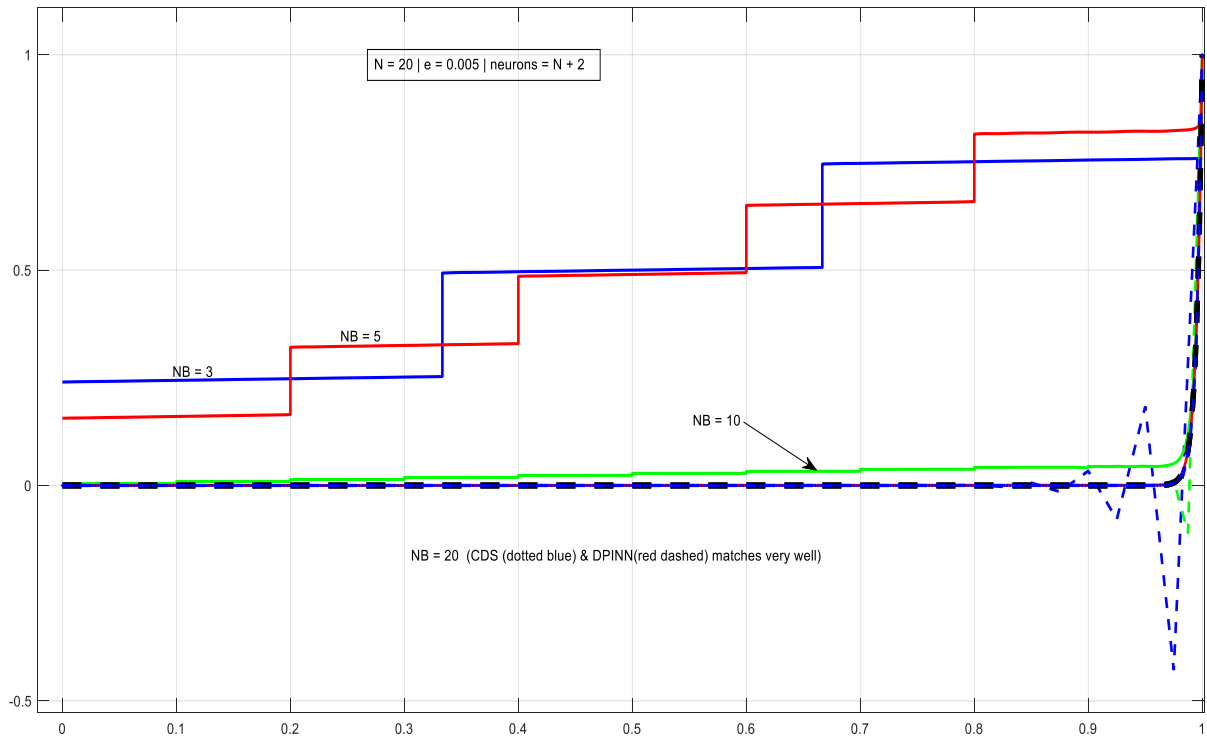

*Figure 6.11: Predictions of ELM-DPINN and CDS for different number of blocks*

The figure below shows the effect of number of collocation points per block (N). With increase in N, approximation is very accurate. The solutions from central differencing (CDS) and normalised ELM-DPIN have same colour while normalised ELM-DPIN model has dashed type lines. CDS and normalised ELM-DPIN have same number of collocation points. Clearly, CDS performs better but with high values of N solutions are very close and accurate.

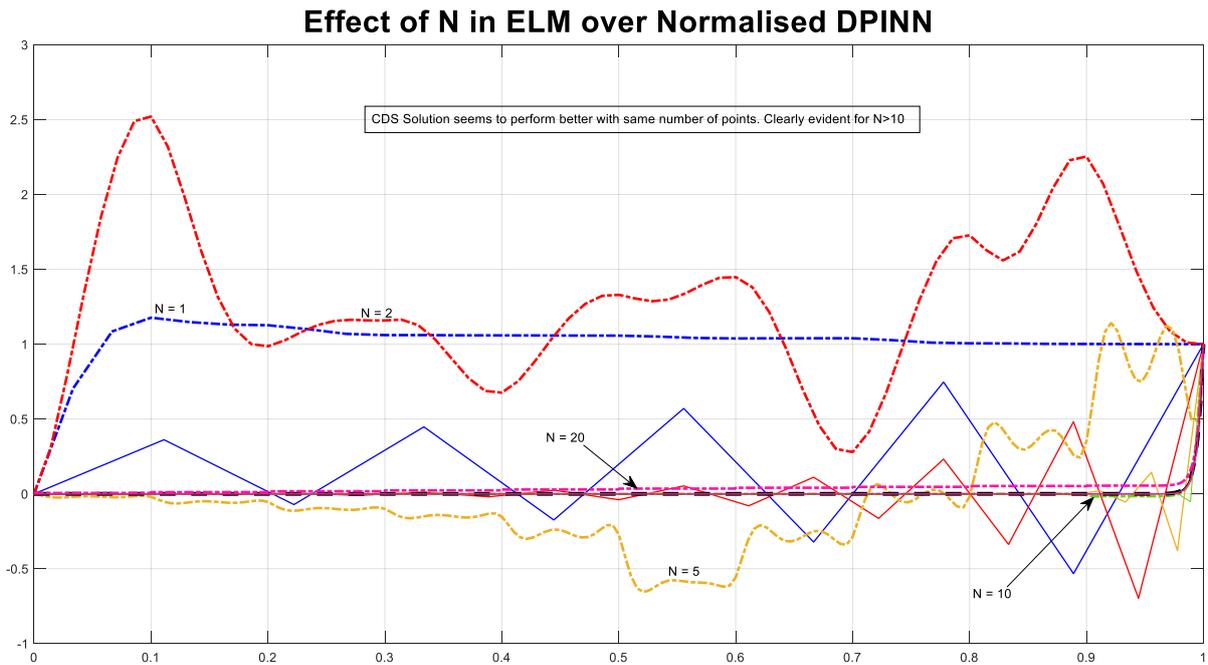

*Figure 6.12: Predictions of ELM-DPINN vs CDS for different number of collocation points*

The figure below shows the solution for different values of $\varepsilon$ with 30 blocks and 20 collocation points per block. Predictions are accurate up to $\varepsilon > 0.0025$. By increasing the number of blocks cases with even lower $\varepsilon$ could be solved but the matrix size becomes very large and the runtime increases heavily.

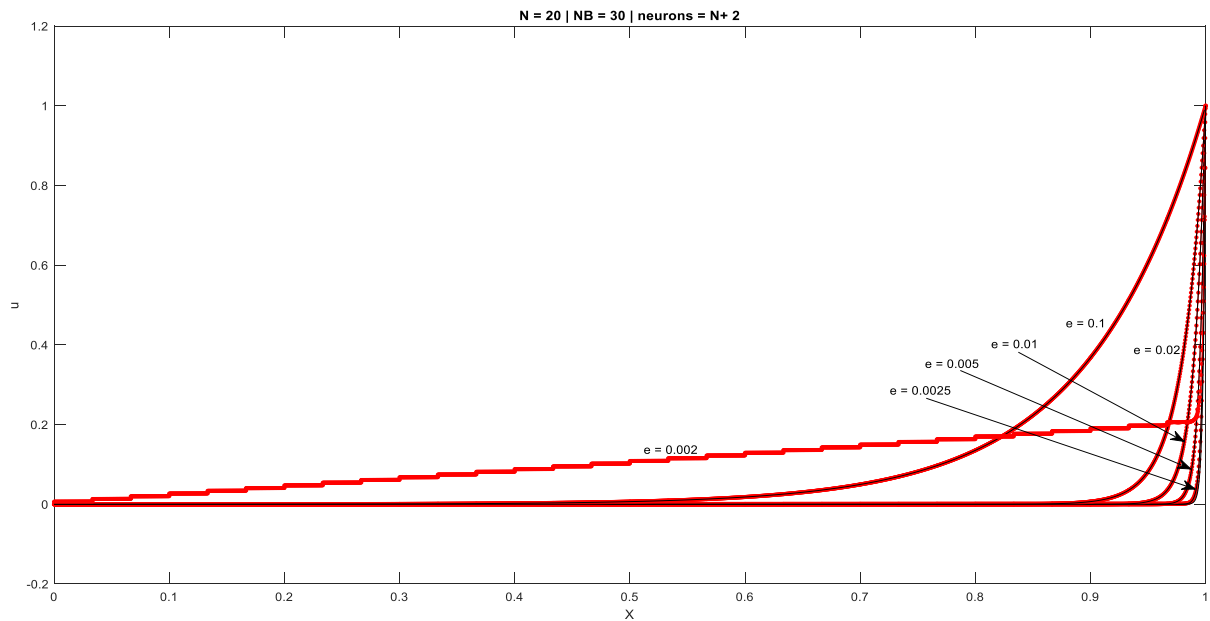

*Figure 6.12: Predictions of ELM-DPINN for different $\varepsilon$*

# CHAPTER 7

# PARAMETRIC STUDY: 1D UNSTEADY ADVECTION USING DPINN

The case of 1D unsteady advection of square pulse is taken into account the have an extensive parametric study. It is necessary to know how does the model behave by variation of different parameters. By doing parametric study, it becomes easier to understand how an optimized model would look like. It gives an understanding to how to tune the network smoothly. Moreover, it also gives us an idea about the abilities and limitations of the model. The cases were run by fixing the random seed. The equation for one-dimensional unsteady advection is,

$$\frac{du}{dt} + C\frac{du}{dx} = 0$$

Unlike the advection diffusion problem, this is time-variant. The two independent variables are x and t. So, the network takes two input and the domain is broken into 2D sub-domains.

## 7.1. EFFECT OF NUMBER OF BLOCKS

The tables below show the values of losses at convergence for different combinations of number of blocks in domain 'x' (NBX) and number of blocks in domain 't' (NBT). The network used here has a single hidden layer with only two neurons. It has 100 collocation points per block each in 'x' and 't'. The losses decrease by putting increasing the number of blocks, both in 'x' as well as 't' domain. It can also be noted that with increase in the value of 'C' the convergence loss increases. Losses in the order of 5.0 x $10^{-4}$ is a descent result in this case. Basically, for higher values of 'C' there is a need for a greater number of blocks, given all other parameters and conditions are unaltered.

| Models (NBT = 5) | NBX = 1 | NBX = 2 | NBX = 3 | NBX = 4 | NBX = 5 |
|---|---|---|---|---|---|
| C = 0.1 | 3.98262e-03 | 4.19683e-03 | 6.16152e-04 | 1.44674e-04 | 1.82159e-05 |
| C = 0.5 | 4.50196e-03 | 1.88064e-03 | 1.88064e-03 | 2.32029e-04 | 6.69062e-04 |
| C = 1 | 6.54351e-03 | 2.31804e-03 | 1.41816e-03 | 1.47843e-03 | 1.70877e-03 |
| C = 2.5 | 5.12187e-03 | 6.20275e-03 | 3.03736e-03 | 2.53500e-03 | 2.03333e-03 |

*Table 7.1: Losses at stopping point for various C against various divisions in x domain*

| Models (NBX = 5) | NBT = 1 | NBT = 2 | NBT = 3 | NBT = 4 | NBT = 5 |
|---|---|---|---|---|---|
| C = 0.1 | 5.25275e-05 | 5.45994e-06 | 2.08225e-05 | 4.19113e-06 | 1.82159e-05 |
| C = 0.5 | 1.06130e-05 | 8.93782e-03 | 9.78340e-05 | 1.05437e-03 | 6.69062e-04 |
| C = 1 | 9.73916e-03 | 7.28228e-03 | 4.01104e-03 | 2.16407e-03 | 1.70877e-03 |
| C = 2.5 | 7.76721e-03 | 7.58001e-03 | 7.45773e-03 | 1.15753e-03 | 2.03333e-03 |

*Table 7.2: Losses at stopping point for various C against various divisions in time domain*

## 7.2. EFFECT OF COLLOCATION POINTS PER BLOCK

The tables below show the values of losses at convergence for different combinations of number of collocation points per block in domain 'x' (nbx) and number of collocation points per block in domain 't' (nbt). The network used here has a single hidden layer with only two neurons. It has 3 blocks each in 'x' and 't' domains. Unlike the choice of minimum number of blocks, here sufficient number of collocation points need to be taken otherwise overfitting can be observed. A descent choice of nbx and nbt plays a very crucial role in determining the best model. The ratio of nbx and nbt seems to play some hidden role as well, but it is very difficult to figure that exactly. However, the effect of nbx seems to have greater influence than nbt. The rough trend is that loss at convergence reduces with the increase in the number of collocation points. Anyway, the final statement would be that, the nbx and nbt must be chosen carefully by checking different combinations.

| Models (nbt = 25) | nbx = 5 | nbx = 10 | nbx = 25 | nbx = 50 | nbx = 75 |
|---|---|---|---|---|---|
| C = 0.1 | 2.76068e-05 (Over Fit) | 1.50466e-05 (Over Fit) | 1.44878e-04 | 2.68922e-04 (better) | 3.57637e-03 (Under Fit) |
| C = 0.5 | 5.78176e-04 | 1.60295e-03 | 8.27145e-03 | 2.46775e-03 | 1.36292e-03 |
| C = 1 | 2.90163e-03 | 3.20112e-03 | 5.24739e-03 | 1.00236e-02 | 1.71046e-03 |
| C = 2.5 | 9.00347e-03 | 1.11462e-02 | 1.03648e-02 | 1.13559e-02 | 5.88193e-03 |

*Table 7.3: Losses at stopping point for various C against various number of collocation points in x domain*

| Models (nbx = 25) | nbt = 5 | nbt = 10 | nbt = 25 | nbt = 50 | nbt = 75 |
|---|---|---|---|---|---|
| C = 0.1 | 3.99792e-04 | 3.19508e-04 | 2.97383e-04 | 1.43109e-04 | 1.98758e-04 |
| C = 0.5 | 9.12109e-04 | 2.55994e-03 | 2.39506e-03 | 2.12984e-03 | 2.16387e-03 |
| C = 1 | 1.70891e-03 | 2.60400e-03 | 5.67608e-03 | 2.36264e-03 | 1.56567e-04 |
| C = 2.5 | 5.59766e-03 | 1.03164e-02 | 4.81908e-03 | 4.62594e-03 | 5.08784e-03 |

*Table 7.4: Losses at stopping point for various C against various number of collocation points in time domain*

## 7.3. EFFECT OF LEARNING RATE

The table below shows the loss at convergence for typical gradient descent (GD), Adagrad and Adam optimizers. The behaviour for different learning rate could be understood from the loss values. One thing that should be noted is the fact that all the optimizers have their own range of learning rates over which they function well but solving over a range of learning rates would give an idea of the capability in optimizing the problem. The Adam's optimizer performs way better than normal gradient descent optimizer and Adagrad optimizer. While comparison between the other two is not straight forward, but Adagrad can sometime perform remarkably better than gradient descent.

| LOSS | Learning Rate | | | |
|---|---|---|---|---|
| **Optimizer** | **0.0001** | **0.001** | **0.01** | **0.1** |
| GD | 0.0483 | 0.0210 | 0.0193 | 0.0174 |
| Adagrad | 0.3107 | 0.0467 | 0.0189 | 0.0070 |
| Adam | 0.0021 | $1.6162 \times 10^{06}$ | $1.0814 \times 10^{-06}$ | 0.0026 |

*Table 7.5: Losses at stopping point for various learning rate against various optimizers*

The three optimizers discussed here are specialized variants of gradient descent. Apart from these there are some other second order optimizers that do not have learning rates so, they have to be kept out of comparison. Practically, they perform much better than these optimizers. They are discussed in upcoming segments.

The figures below show the trend of log(loss) for different learning rates for each optimizer.

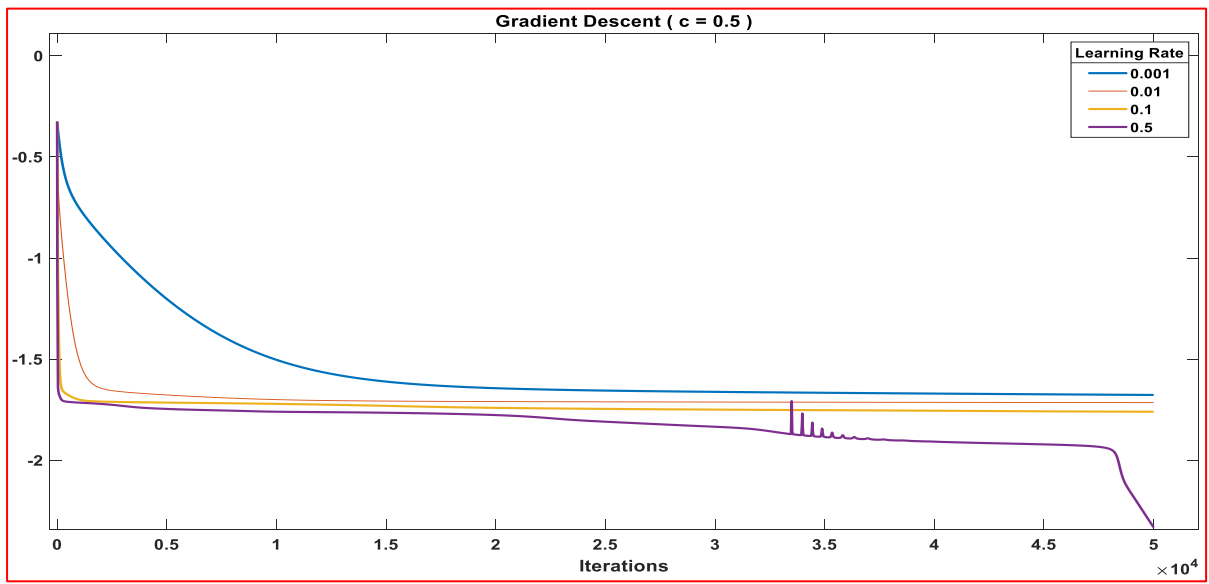

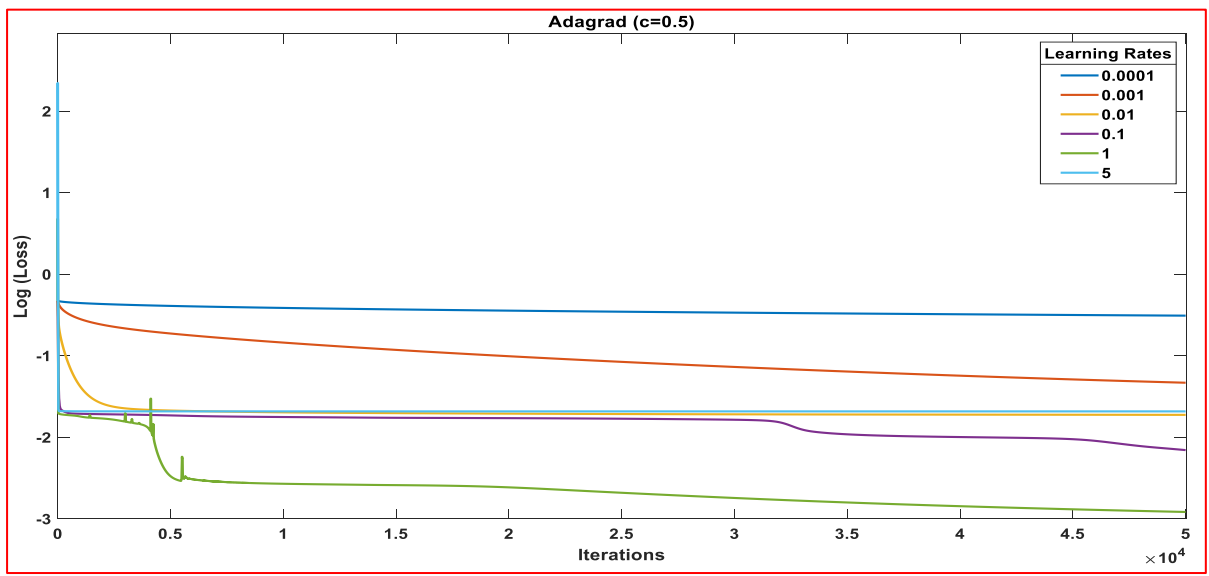

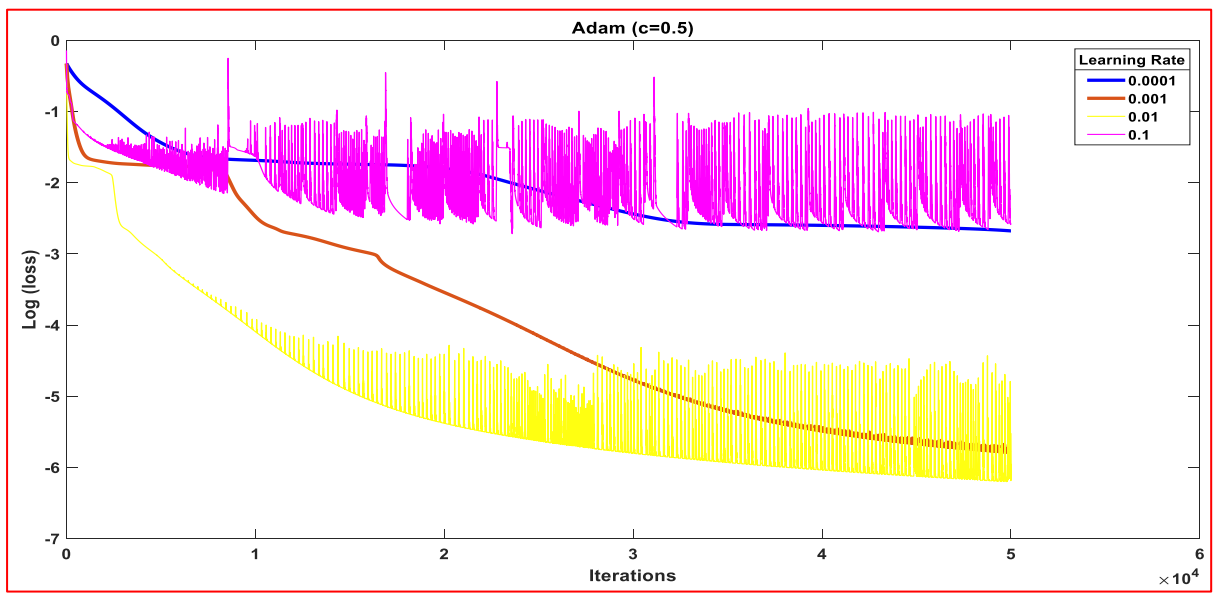

*Figure 7.1(a), (b), (c): log(loss) vs iterations for different learning rates with GD(a), Adagrad(b) and Adam(c)*

## 7.4. EFFECT OF OPTIMIZER

The optimizers that work really well are Adam, L-BFGS-B, conjugate gradient (CG) and SLSQP. The L-BFGS-B seems to be the best optimizer from the bucket. Adam would be at the second place and then SLSQP and CG would follow. The table below shows the loss value at the time of stopping for a case of unsteady advection.

| Models LR | GD | Adam | L-BFGS-B | CG | COBYLA | SLSQP | TNC | Nelder-Mead |
|---|---|---|---|---|---|---|---|---|
| 0.001 | 0.0210 | $1.6162 \times 10^{-06}$ | $1.4204 \times 10^{-07}$ | $4.5154 \times 10^{-04}$ | 0.0203 | $5.6131 \times 10^{-04}$ | 0.0080 | 0.0208 |
| 0.01 | 0.0193 | $1.0814 \times 10^{-06}$ | | | | | | |
| 0.1 | 0.0174 | 0.0026 | | | | | | |

*Table 7.6: Losses at stopping point for various optimizers*

The figure below shows the trend of log(loss) for different optimizers.

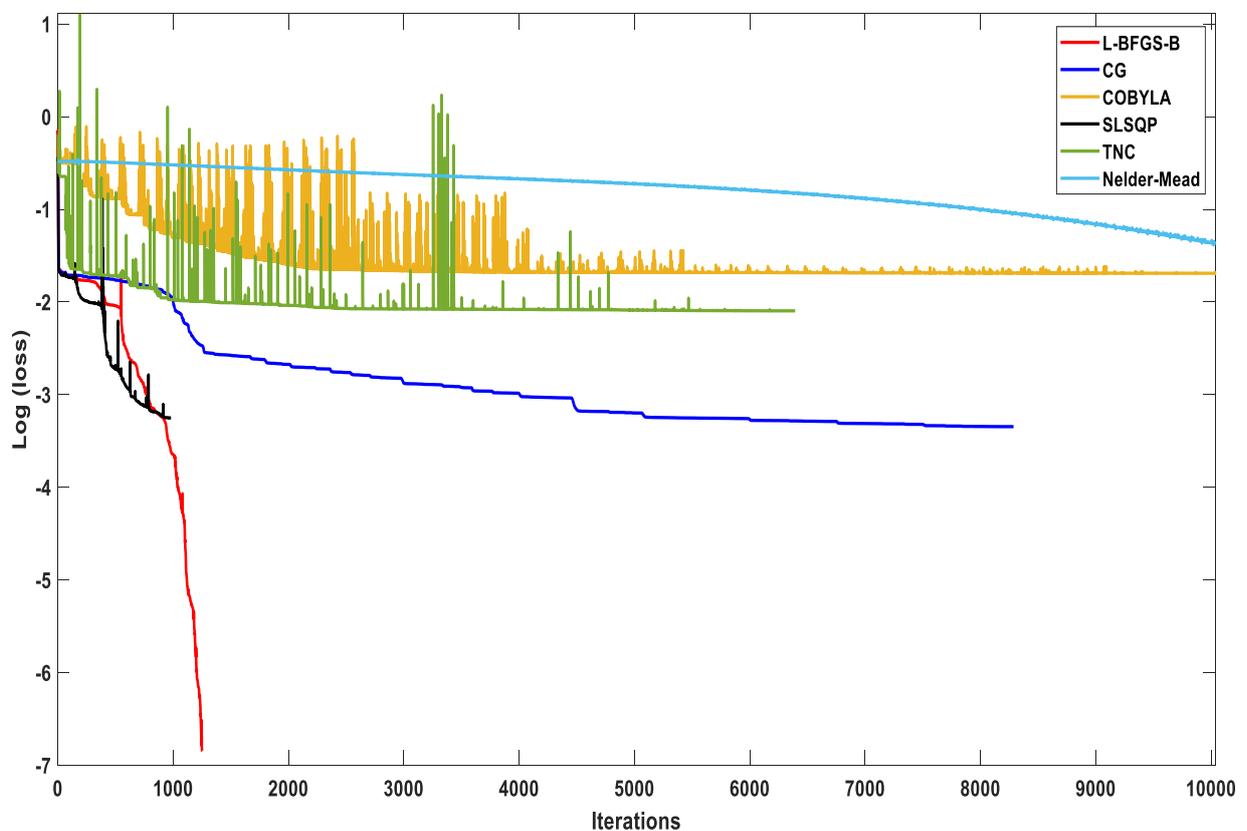

*Figure 7.2: log(loss) vs iterations for different optimizers*

## 7.5. EFFECT OF NON-LINEARITY

Non-linearity (activation function) plays an important role in architecture. The differential equation was solved for different models. In all the cases it was clear that tanh(x) leads the game. Arctan(x) could sometime be very close to tanh(x) but tanh is the fastest among the others in group. Then sigmoid and log(sigmoid) works with bit worse solution.

| Models | | | Loss | | | | |
|---|---|---|---|---|---|---|---|
| C | Neurons | Optimizer | Sigmoid | Tanh | SoftPlus | Log Sigmoid | ArcTan |
| 0.5 | 2 | Adam | 7.75615e-04 | 1.47605e-05 | 4.77046e-03 | 7.31412e-03 | 1.11537e-04 |
| 1 | 2 | Adam | 1.55154e-03 | 2.19862e-04 | 4.46882e-03 | 1.46074e-02 | 4.01917e-04 |
| 0.5 | 2 | L-BFGS-B | 1.28998e-03 | 1.27516e-06 | 1.49661e-02 | 1.79618e-02 | 6.72869e-04 |
| 1 | 2 | L-BFGS-B | 1.70795e-02 | 1.91109e-04 | 1.93503e-02 | 1.93258e-02 | 1.08187e-03 |
| 0.5 | 4 | L-BFGS-B | 2.55921e-06 | 1.59368e-05 | 2.43783e-03 | 2.39689e-03 | 9.42070e-05 |

| 1 | 4 | 2.81592e-06 | 3.33567e-04 | 4.54190e-03 | 2.79268e-03 | 4.18945e-04 |
|---|---|---|---|---|---|---|
| 0.5 | 8 | 9.95310e-05 | 5.13360e-06 | 1.79042e-02 | 2.69044e-03 | 2.04379e-05 |
| 1 | 8 | 3.38175e-04 | 1.34256e-05 | 1.61106e-02 | 9.27748e-03 | 1.62906e-04 |
| 1.5 | 8 | 1.92506e-03 | 7.63881e-05 | 1.70217e-02 | 8.23745e-03 | 7.14311e-04 |
| 2 | 8 | 3.10389e-04 | 2.35750e-04 | 1.77785e-02 | 7.77693e-03 | 1.16140e-04 |

*Table 7.7: Losses at stopping point for different activation functions*

## 7.6. EFFECT OF NUMBER OF NEURONS IN EACH LAYER

The next table shows the values of losses at convergence for different number of neurons per layer. The x-t domain is divided into 3 blocks each in domain 'x' (NBX) and domain 't' (NBT). The number of collocation points per block in domain 'x' (nbx) and number of collocation points per block in domain 't' (nbt) are kept at 20 each. The network used here has a single hidden layer with only two neurons. The prediction loss reduces with increasing number of neurons per layer. It could be observed quite consistently for one or two layers but with lot of neurons there is no much advantage, in fact the complexity of optimisation increases. So, 5 neurons per layers seems to be a very optimal choice.

| Models | Neurons | | | | |
|---|---|---|---|---|---|
| C | Layers | 1 | 2 | 5 | 10 |
| 0.5 | 1 | 1.12387e-02 | 3.08604e-03 | 1.17149e-04 | 2.81565e-04 |
| 1 | 1 | 1.16298e-02 | 9.13148e-03 | 2.98847e-04 | 4.91855e-04 |
| 2 | 1 | 1.19093e-02 | 1.18731e-02 | 1.64297e-03 | 1.62882e-03 |
| 0.5 | 2 | 1.13211e-02 | 1.56274e-03 | 5.96564e-06 | 1.73199e-05 |
| 1 | 2 | 1.39621e-02 | 1.07156e-02 | 1.18483e-04 | 2.97387e-05 |
| 2 | 2 | 1.19235e-02 | 1.18783e-02 | 3.99803e-03 | 3.61938e-04 |
| 0.5 | 3 | 1.14864e-02 | 1.02176e-05 | 1.74242e-05 | 8.67388e-03 |
| 1 | 3 | 1.86805e-02 | 1.39241e-02 | 3.86329e-03 | 1.87314e-02 |
| 2 | 3 | 1.18398e-02 | 7.29442e-03 | 1.13935e-04 | 1.87302e-02 |

*Table 7.8: Losses at stopping point for different number of neurons per layer*

## 7.7. EFFECT OF NUMBER OF LAYERS

The table below gives the loss values for different combinations of number of neurons per layer and the number of layers in an architecture. Increasing the number of layers does not seem to be helping at all. There is no requirement for having deeper networks. Having two or three layers looks like a descent choice. From the table below, it can be observed that the combination of few neurons and few layers gives the least loss value at the time of stopping.

| Models | Layers | | | | | |
|---|---|---|---|---|---|---|
| C | Neurons | 1 | 2 | 3 | 4 | 5 |
| 0.1 | 2 | 2.01146e-03 | 8.21989e-04 | 7.35717e-03 | 7.66397e-03 | 1.95225e-02 |
| 0.1 | 4 | 8.98288e-04 | 1.66258e-04 | 1.00143e-05 | 3.82805e-04 | 1.95221e-02 |
| 0.1 | 8 | 8.83118e-04 | 6.90191e-05 | 2.80913e-04 | 1.95226e-02 | 1.95226e-02 |
| 0.5 | 2 | 2.82546e-03 | 2.85063e-04 | 3.17684e-04 | 1.70925e-05 | 1.03411e-02 |
| 0.5 | 4 | 4.38039e-04 | 1.47312e-05 | 1.14612e-05 | 6.56383e-03 | 1.95252e-02 |
| 0.5 | 8 | 6.62320e-04 | 1.80579e-05 | 1.12166e-02 | 1.95628e-02 | 1.95225e-02 |
| 1 | 2 | 2.11059e-03 | 5.31171e-03 | 1.03038e-02 | 6.06567e-06 | 1.09843e-02 |
| 1 | 4 | 1.05989e-03 | 6.90537e-05 | 4.08750e-05 | 1.42897e-02 | 1.95230e-02 |
| 1 | 8 | 1.04268e-03 | 4.24757e-05 | 1.95199e-02 | 1.95228e-02 | 1.95226e-02 |

*Table 7.9: Losses at stopping point for different number of layers*

# CHAPTER 8

# CONCLUSIONS

To summarize, the proposed methods is a good approximation method to solve wide class of CFD problems. It works much better than the prevalent methods. The older method by I. E. Lagaris et.al. could solve an advection dominant problem if $\epsilon > 0.5$ while the recent version by Maziar Raissi et. al. could solve a problem if $\epsilon > 0.15$. The proposed method Distributed Physics Informed Neural Network (DPINN) could easily solve the same problem for $\epsilon > 0.002$. The next level advancement, ELM based DPINN, could easily solve for $\epsilon > 0.003$ while it struggles a bit for lower $\epsilon$ due to large memory involved in the process. In future these methods can grow up to replace the conventional methods. Otherwise ELM is faster than all the above model discussed above and gives solution in single shot.

# RECOMMENDATIONS FOR THE FUTURE WORK

The problems at current level is the competition with prevalent conventional methods. The machine learning models generally take time to train themselves so, these models are slower compared to conventional models. Since, it has been discussed how these models can overpower conventional methods, there is no doubt on the ability. So, lot of research need to be done to figure out the solutions.

One method can be to go for adaptive splitting of domains like adaptive meshing in conventional methods. That would certainly reduce the computational burden.

Another thing that could be researched on is the optimization issue in solving extreme problems. There should be a method to improve the optimization. One way would be to pose the various loss objectives together in a better way like assigning proper weights, similar to the logic behind Lagrange's multiplier method for multi-objective method.

The solutions for negative $\epsilon$ are somehow, way better. The reason behind this could be researched.

The ELM based DPINN fails for larger matrices due to the issue of large memory. A better method can be proposed to tackle this issue.